\documentclass[a4paper,twocolumn,10pt,3p]{elsarticle}  % Comment this line out if you need a4paper

\usepackage{graphics} % for pdf, bitmapped graphics files
\usepackage{epsfig} % for postscript graphics files
\usepackage{times} % assumes new font selection scheme installed
\usepackage{amsmath} % assumes amsmath package installed
\usepackage{amssymb,mathrsfs}  % assumes amsmath package installed
\usepackage{algorithm}
\usepackage[dvipsnames,usenames]{color}
\usepackage{array}
\usepackage{url}
\graphicspath{ {./fig/} {./pic/}{./bio/}}

\newtheorem{defn}{Definition}

\begin{document}

%%%%%%%%%%%%%%%%%%%%%%%%%%%%%%%%%%%%%%%%%%%%%%%%%%%%%%
\begin{frontmatter}
\title{A Novel Geometry-based Algorithm for Robust Grasping in Extreme Clutter Environment }

\author[ok]{Olyvia Kundu}
\ead{olyvia.kundu@tcs.com}

\author[ok]{Swagat Kumar\corref{corspauth}}
\ead{swagat.kumar@tcs.com}

\address[ok]{TATA Consultancy Services, Bangalore, India 560066}
\cortext[corspauth]{Corresponding author}

\begin{abstract}
  This paper looks into the problem of grasping unknown objects in a
  cluttered environment using 3D point cloud data obtained from a
  range or an RGBD sensor. The objective is to identify graspable regions
  and detect suitable grasp poses from a single view, possibly,
  partial 3D point cloud without any apriori knowledge of the
  object geometry. The problem is solved in two steps:  (1) 
  identifying and segmenting various object surfaces and, (2) searching
  for suitable grasping handles on these surfaces by applying
  geometric constraints of the physical gripper. The first step is
  solved by using a modified version of region growing algorithm that
  uses a pair of thresholds for  smoothness constraint on local
  surface normals to find natural boundaries of object surfaces. In
  this process, a novel concept of \emph{edge point} is introduced
  that allows us to segment between different surfaces of the same
  object. The second step is solved by converting a 6D pose detection
  problem into a 1D linear search problem by projecting 3D cloud
  points onto the principal axes of the object surface. The graspable
  handles are then localized by applying physical constraints of the
  gripper. The resulting method allows us to grasp all kinds of objects
  including rectangular or box-type objects with flat surfaces which
  have been difficult so far to deal with in the grasping
  literature. The proposed method is simple and can be implemented in
  real-time and does not require any off-line training phase for finding these
  affordances.  The improvements achieved is demonstrated through
  comparison with another state-of-the-art grasping algorithm on
  various publicly-available and self-created datasets.
\end{abstract}


\end{frontmatter}

%\linenumbers
%%%%%%%%%%%%%%%%%%%%%%%%%%%%%%%%%%%%%%%%%%%%%%%%%%%%%%%%%%%%%%%%%%%%5

\section{INTRODUCTION} \label{sec:intro}

A robot that can manipulate its environment is much more useful than
one that can only perceive. Such robots can act as active agents which
will someday replace humans from all types of dull, dangerous and
dirty works completely, thereby, freeing them for more creative
pursuits. Grasping is an important capability necessary for realizing
this end. Solving the grasping problem involves two steps.
This first step uses perception module to estimate the pose (position
and orientation) of the object and hence, the gripper pose needed for
picking it. Then, the second step uses a motion planner to generate
necessary robot and gripper movement to make contact with the object.  
The usual approach, currently employed in industries, uses the
available knowledge of accurate geometry of objects as well as the
environment to solve the grasping problem off-line for all objects
and then, apply it to pick objects based on their recognition
during online implementation. Many of the recent approaches now solve
the grasping problem independent of the object identity, thanks to the
availability of 3D or RGBD point cloud available from low cost depth
and range sensors.  In general, grasping unknown objects in a
cluttered environment still remains an open problem and has attracted
considerable amount of interest in the recent past.
      
%This paper focusses on solving the perception part of the problem
%which is more commonly known as grasp pose detection (GPD)
%\cite{plattgrasppose2016} or simply, robotic grasp detection
%\cite{JainICRA16}. A large part of related work in this field deals
%with object recognition and object categorization based on their
%shape and then apply know grasps for these objects. The availability
%of object models simplifies the grasping problem to a greater extent.
%However, it still suffers from two major limitations - first, one has
%to maintain a database of all known objects and hence dealing with
%unknown objects will be difficult and secondly, it is still difficult
%to find a match for a partial point cloud in a cluttered environment.
%More recently, researchers are focussing on localizing graspable
%regions irrespective of the object identity directly in a RGBD or 3D
%range point cloud. 

There are several approaches to solve the grasping problem which are
reviewed briefly in the next section. In this paper, we are primarily
interested in solving the first part of the problem, namely, finding
graspable regions and suitable grasp poses (together known as
graspable affordances) for a two-finger parallel-jaw gripper. This
is more formally termed as robotic grasp detection \cite{JainICRA16}
or simply, grasp pose detection (GPD) \cite{plattgrasppose2016}. The
graspable affordances are to be detected in a RGBD or a 3D point cloud
obtained from a single view of the range or depth sensor, without
requiring any apriori knowledge of the object geometry. Our work is
inspired by the approach presented in \cite{Pas2013LocalizingGA}
\cite{ten2016localizing} which uses surface curvature to localize
graspable regions in the point cloud. The advantage of this approach
lies in its simplicity which allows real-time implementation and does
not require any time-consuming and data-intensive training phase
common in most of the learning-based methods \cite{saxena2008}
\cite{lenz2015deepgrasp} \cite{johns2016deep}. However, this approach
suffers from several limitations, which forms the basis for the work
presented in this paper. For instance, it can not be used for grasping
objects with flat surfaces, such as boxes, books etc. This is
partially remedied in \cite{pas2015using} where authors use Histogram
of Gradients (HoG) features \cite{dalal2005histograms} to generate
multiple hand hypotheses and then, train a SVM network to detect
the valid grasps among these hypotheses.  Secondly, their algorithm
necessitates creating several spherical regions of fixed user-defined
radius to search for these graspable regions. Hence, it can not be
used for localizing varying sizes of handles as it requires adjusting
the radii of these spherical regions. Because of these limitations,
the above algorithm performs poorly in extreme clutter environment
where there could be multiple objects adjacent to each other. 

In this paper, we propose a new approach based on surface normals to
overcome these limitations. It primarily involves two steps. The first
step uses surface continuity of surface normals to identify the
natural boundaries of objects \cite{vosselman2004recognising}
\cite{rabbani2006segmentation}. A modified version of the region-growing
algorithm is proposed that can distinguish between difference surfaces of
the same object having large variation in the direction of their
surface normals. This is done by defining \emph{edge points} and
introducing a pair of thresholds on the similarity criterion used by
the region growing algorithm.  This is a novel concept which
dramatically improves the performance of the region growing algorithm
in removing spurious edges and thereby, identifying natural boundaries
of each object even in a clutter. One of the clear advantage of this
approach is that it allows us to find graspable affordances for box
type objects with flat surfaces, which is considered to be a difficult
problem in the vision-based grasping literature.  Also unlike
\cite{Pas2013LocalizingGA} \cite{ten2016localizing}, the proposed
algorithm requires far too less number of user-defined parameters and
does not require initialization using spherical regions at multiple
locations. 

%In contrast, we rely
%on locally computed surface normals and use surface discontinuity to
%identify natural boundaries of objects \cite{vosselman2004recognising}
%\cite{rabbani2006segmentation}, which are then used for grasping. This
%is achieved by using a modified version of region growing algorithm
%that uses a dual boundary condition on similarity criterion in order
%to decide the label for a neighboring point in the cloud. This is
%empirically shown to remove spurious edges giving rise to more
%continuous surfaces which improves the performance of the grasping
%algorithm. This also allows one to distinguish between different
%surfaces of a box type object. In addition, use of region growing to
%create continuous surfaces based on smoothness constraint eliminates
%the need of many user-defined parameters, thereby, providing better
%generalization over a large variety of objects.  This is an important
%improvement over the method in \cite{ten2016localizing} where multiple
%spheres of fixed radius are selected to initialize the search for
%graspable regions.

Once the surfaces for different objects are segmented, the second step
uses the gripper geometry to localize the graspable regions. The six
dimensional grasp pose detection problem is simplified by making
practical assumption of the gripper approaching the object in a
direction opposite to the surface normal of the centroid of the
segment with its gripper closing plane coplanar with the minor axis of
the segment. The principal axes for each segment (major, minor and
normal axes) are computed using Principal Component Analysis (PCA)
\cite{wold1987principal}. Essentially, the valid grasping regions are
localized by carrying out a one-dimensional search along the principal axes of the
segment and imposing the geometrical constraints of the gripper. This is made
possible by projecting the original 3D Cartesian points of the surface
onto the principal axes of the surface segment.  This approach is
mathematically much simpler compared to other methods
\cite{vezzani2017grasping} \cite{tableTopGrasping} which require more
complex  processes to make such decisions.  The proposed grasping
method is found to be more robust in localizing graspable affordances in
extreme clutter scenario as compared to the state-of-the-art
algorithms. This improvement is achieved without compromising on the
real-time performance of the algorithm.

In short, the major contributions made in this paper are as follows:
(1) a new method is proposed that exploits segmentation based on
smoothness of surface normals
\cite{rabbani2006segmentation} to find graspable affordances in a 3D
point cloud. This allows us to grasp box-type objects with flat
surfaces, which is considered to be a difficult problem in the
grasping literature.  (2) The introduction of a novel concept of
\emph{edge point} and a modified region growing algorithm that uses a
pair of thresholds on smoothness constraint allows us to distinguish
between different surfaces of the same object and helps in finding the
natural boundaries of objects in a clutter by removing spurious edges.
This, in turn, allows us to grasp box-type objects which is otherwise
considered to be difficult. (3) The problem of 6D pose detection is
simplified by constraining the search space by using the gripper
geometry and converted into a one-dimensional search problem through
scalar projections of 3D points on to the principal axes of the
surface segment.  The resulting method is quite simple and can be
implemented in real-time and does not require any data-intensive and
off-line training phase. (4) In this process, we contribute a new
dataset that can be used for analyzing the performance of various
grasping algorithms. This dataset exhibit various real world scenario
including extreme clutter, constrained view within a bin etc. Finally,
the improvement achieved by the proposed algorithm is demonstrated
through rigorous experiments  on several publicly available dataset
including our own.

The rest of this paper is organized as follows. A brief overview of
related literature is provided in the next section. The symbols and
notations used for explaining the method is provided in Section
\ref{sec:sn}. The proposed method is explained in Section
\ref{sec:meth} and the experimental and simulation results are
provided in Section \ref{sec:res}. The summary and conclusion is made
in Section \ref{sec:conc}.

\section{Related Work} \label{sec:relw}

Grasping is a challenging problem which has attracted considerable
amount of interest over last couple of decades. This section provides
an overview of various related work in this domain. We will
particularly focus on \emph{precision} grasping where the object is
held with the tips of the fingers providing higher sensitivity and
dexterity compared to \emph{power} grasping which involves large area
of contact between the hand and the object \cite{gori2013ranking}. If
the accurate 3D model of the object is available, one can leverage
force-closure and form-closure to find stable grasp configurations for
holding the object but it presupposes the availability of contact
points on the object \cite{bicchi2000robotic}. These contact points
are, however, not easily available in real world scenarios and hence,
forms the subject matter of our interest in this paper. The rapid
advancement in computer vision algorithms has enabled researchers to
use vision sensors to identify graspable regions in objects. These
methods extract 2D image features and combine them with geometric
methods either to compute size or shape of the object \cite{saxena2008}
\cite{kehoe2013cloud} \cite{song2011grasp}. The availability of low
cost range or RGBD sensors have further simplified the grasping
problem by providing 2.5D or 3D point cloud data. In the later
category, a number of approaches have been tried to extract object
shape information. For instance, Fischinger and Vincze
\cite{fischinger2012empty} use a novel Height accumulated feature
(HAF) to detect shape and find grasping regions in a cluttered
environment. Similarly, the authors in \cite{vezzani2017grasping}
\cite{varadarajan2011object} use superquadric functions to model
objects. There are other methods that exploit geometric properties of
gripper and object surfaces to detect suitable grasping regions as in
\cite{tableTopGrasping} \cite{ten2016localizing}.  Some methods try to
fit a primitive shape around the object point cloud and this is used
for deciding the suitable graspable pose for the robotic gripper
\cite{jain2016grasp} \cite{somani2014shape} \cite{behnke2012shape}
\cite{kragic2008boxprimitive}. Many of these methods produce multiple
grasp hypotheses based on certain heuristics and then use machine
learning methods to evaluate them  \cite{geidenstam2009learning}
\cite{plattgrasppose2016} \cite{saxena2008learning}. Quite recently,
deep learning methods are being increasingly used for solving the
grasping problem \cite{lenz2015deepgrasp} \cite{johns2016deep}
\cite{pinto2016supersizing} \cite{Schwarz:7139363}. These methods
require time-consuming data gathering and off-line training processes
which limit their applications to many real world problems. 

We are particularly inspired by the work by Ten pas and Platt
\cite{ten2016localizing} where the grasping regions (or affordances)
are directly computed from a single view 3D point cloud obtained from a
RGBD sensor, without requiring any apriori knowledge of the object
geometry. They use a fixed radius circle to cluster 3D points in the
environment at multiple locations. This radius is decided based on
the geometry of the gripper and clearance required for avoiding collision
with neighboring objects. In a sense, the authors do not make use of
surface attributes to find object boundaries. The success of their
method relies on the careful selection of this radius which can not be
generalized to wide varieties of objects in a cluttered scenario.
Secondly, it is biased towards identifying curved surfaces and hence,
can not deal with objects with flat surfaces. This is partially
remedied in \cite{pas2015using} where such rectangular edges are
identified using HoG features and a trained SVM classifier. They further
improve the accuracy of their method in \cite{plattgrasppose2016} by
training a deep network with multi-view images of the same object.  

In contrast to the above approaches, we use region growing algorithm
\cite{vosselman2004recognising} to cluster 3D points based on the
orientation of surface normals \cite{mitra2003estimating}. Such
methods are popular for segmenting point clouds based on smooth
constraint \cite{rabbani2006segmentation}
\cite{vosselman2004recognising}. We modify the existing region growing
algorithm introducing a concept of \emph{edge points} and
incorporating two user-defined thresholds which remarkably improves
the detection of discrete boundaries even on the same object. In other
words, we exploit the discontinuity of surface normals to identify the
natural object boundaries unlike Platt's approach
\cite{ten2016localizing} which merely uses pre-defined spheres to
define object boundaries. This forms the first main contribution of
this paper which allows us to detect graspable regions for a wide
variety of objects including flat surfaces in a heavily cluttered
scene as will be demonstrated later in this paper. 

\begin{figure*}[!t]
  \centering
  \begin{tabular}{ccc}
    \scalebox{0.8}{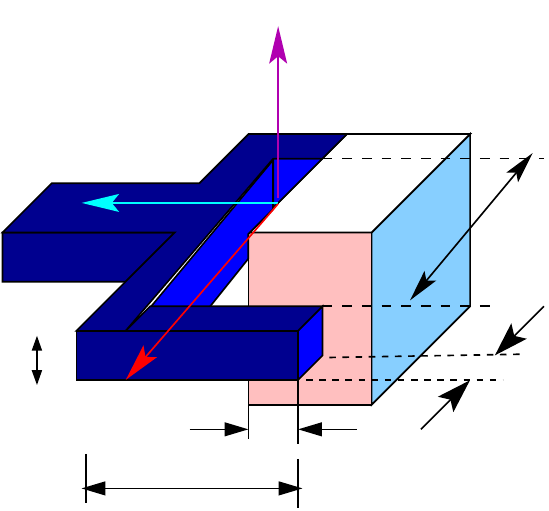} &
    \scalebox{0.8}{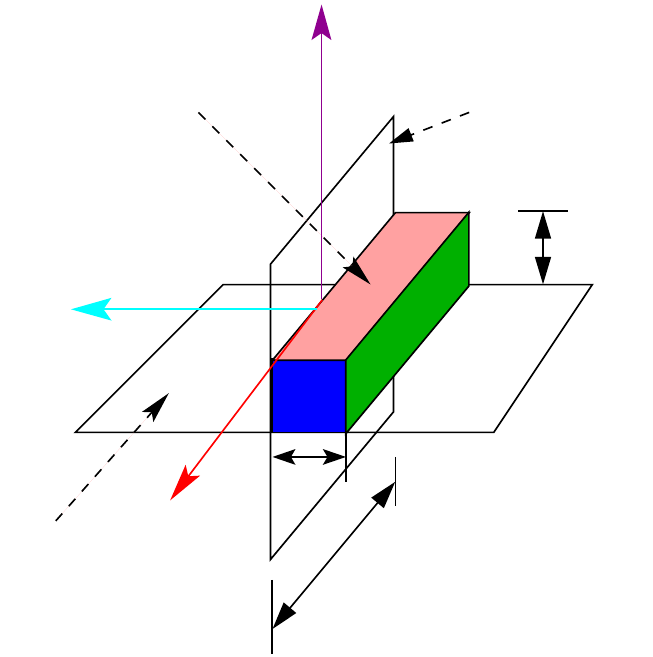} &  \hspace{-1cm}
    \scalebox{0.8}{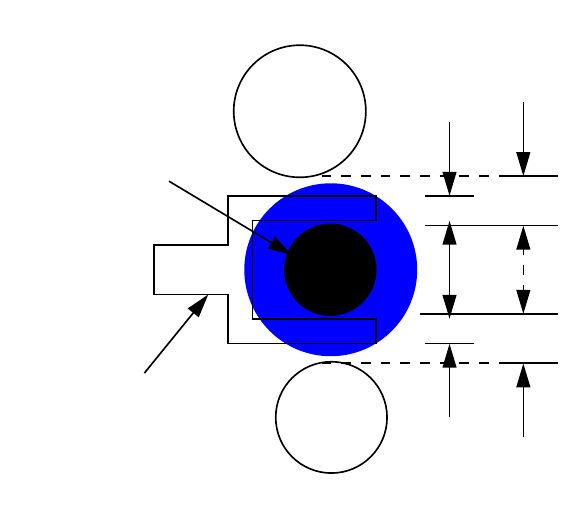}   \\
    \scriptsize{(a) Gripper Geometry} & \scriptsize{(b) Grasping planes} & \scriptsize{(c) Clearance between objects}
  \end{tabular}
  \caption{Grasp Configuration for a two finger parallel jaw gripper.
    For a successful grasp, the clearance between objects must be
    greater than the width of each finger ($g > w$). The gripper
    approaches the object along a direction opposite to the surface
    normal with its gripper closing plane coplanar with the minor axis $\hat{f}$
  of surface segment as shown in (b). The objects 1 and 2 in (c) are
obstacles which are to be avoided by the gripper while grasping the
target object 3. The 6D pose detection problem becomes a 1-D linear
search for a volume of $l\times b \times e$ along the major axis
$\hat{a}$. }
  \label{fig:gripper_object}
\end{figure*}

Once these object boundaries are identified, the geometrical
attributes (size, pose etc.) of these objects are computed using
Principal Component Analysis \cite{wold1987principal}. The information
obtained from this step is further used to localize the graspable
regions on a given object surface. The grasping problem involves
search in a 6D space which is a computationally intensive task. The
dimensionality of this search space is reduced by imposing constraints
through primitive shape fitting \cite{jain2016grasp} or by using
superquadrics \cite{vezzani2017grasping} \cite{Decomposition:Grasp} or
another functions to model object shape.  In contrast to these
methods, we convert the 6D search problem to a one-dimensional search
problem by exploiting the geometrical attributes of the surface as
well as the gripper. This simplification is achieved without
necessitating any complex mathematical process and can be generalized
for any kind of objects without necessitating any apriori knowledge.
The problem formulation and the solution details are described next in
this paper.

\section{Problem Definition} \label{sec:sn}

In this paper, we look into the problem of finding graspable
affordances for a two finger parallel-jaw gripper in a 3D point cloud
obtained from a single view of a range or RGBD sensor. The affordances for objects are
to be computed in an extreme clutter scenario where many objects could
be partially occluded. The problem is solved by taking a geometric
approach where the geometry of the robot gripper is utilized to
simplify the problem. 

Various geometrical parameters corresponding
to the gripper and the object to be grasped is shown in Figure
\ref{fig:gripper_object}(a). The maximum hand aperture is the maximum
diameter that can be grasped by the robot hand and is denoted by $d$.
It should be greater than the diameter $b$ of a cylinder encircling
the object.  It is further assumed that each finger of the gripper has
a width $w$, thickness $e$ and total length $h$. The minimum amount of
length needed for grasping an object successfully is assumed to be
$l$. There has to sufficient clearance between objects so that a
gripper can make contact with the target object without colliding with
its neighbors. Let this minimum clearance needed between two objects
be $g$ and it should be more than the width of each finger, i.e.,
$g>w$ to avoid collision with non-target objects while making a
grasping manoeuvre.  This clearance is shown as blue ring in Figure
\ref{fig:gripper_object}(c). Each object surface is associated with
three principal axes, namely, $\hat{n}$ normal to the surface and two
principal axes - major axis $\hat{a}$ orthogonal to the plane of
finger motion (gripper closing plane) and minor axis - $\hat{f}$ which
is orthogonal to other two axes as shown in Figure
\ref{fig:gripper_object}(b). Readers can refer to \cite{pas2015using}
to understand some of the terms which have been used here without
being defined to avoid repetitions. 

The proposed grasp pose detection algorithm takes a 3D point cloud
$\mathscr{C} \in \mathscr{R}^3$ and a geometric model of the robot
hand as input and produces a six-dimensional grasp pose handle $H
\subseteq SE(3)$. The six-dimensional grasp pose is represented by the vector
$\mathbf{p} = [x, y, z, \theta_x, \theta_y,\theta_z]$, where $(x,y,z)$
is the point where a closing plane of the gripper and object surface
seen by the robot camera intersect; and, $(\theta_x, \theta_y,
\theta_z)$ is the orientation of the gripper handle with respect to a
global coordinate frame. Searching for a suitable 6 DOF grasp pose is
a computationally intensive task and hence, a practical approach is taken
where the search space is reduced by applying several constraints. For
instance, it is assumed that the gripper approaches the object along a
plane which is orthogonal to the object surface seen by the robot
camera. In other words, the closing plane of the gripper is normal to
the object surface as shown in Figure \ref{fig:gripper_object}(b).
Since the mean depth of the object surface is known, the pose
detection problem becomes a search for three-dimensional $(l\times b\times e)$
bands along the major axis $\hat{a}$ where $l$ is the minimum depth
necessary for holding the object. Hence, the grasp pose detection
becomes a one-dimensional search problem once an object surface is
identified.   

Hence, the problem of computing graspable affordances or grasp pose
detection boils down to two steps: (1) creating surfaces in three
point clouds and, (2) applying geometric constrains of a two finger
parallel jaw gripper to reduce the search space for finding suitable
gripper hand pose. The details of the proposed method to solve these two problems is
described in the next section.

\section{Proposed Method} \label{sec:meth}

As explained in the previous section, the proposed method for finding
graspable affordances involves two steps: (1) Creating continuous
surfaces in the 3D point cloud and then, (2) applying geometrical
constraints to search for suitable gripper poses on these surfaces.
This is described next in the following subsections.  

%\textcolor{blue}{The proposed method is flexible enough
%  to be applied on unorganized point cloud data so that it can be a
%  part of a system where input data is transmitted over a network.
%  This is an advantage as the structured data is usually bulky in size
%  and can give rise to high latency when transmitted over the network
%  [??]. Many of the methods, such as, \cite{lenz2015deepgrasp}
%  \cite{PintoG15} \cite{fischinger2012empty} use structured point
%  clouds and suffer from this limitation. On the other hand, there are
%  methods, such as, \cite{vezzani2017grasping}
%  \cite{Pas2013LocalizingGA} that use organized point clouds don't
%  suffer from this limitation. Our method falls into the later
%category.} (This could be removed \dots) 

%structured
%%input. Some methods can work on unorganized data but they either fit
%%shape primitive or rely on prior models \cite{superquadric2017}. The
%%work described in \cite{Pas2013LocalizingGA} can work on unorganized
%%data without any prior information, hence has maximum similarity with
%%our method. 
%
%Unlike methods \cite{PintoG15}\cite{fischinger2012empty} that
%explicitly require availability of structured point cloud, our
%proposed method can directly work with unorganized point clouds which
%are much lighter compared to structured data and can be easily
%transmitted over network. 

\subsection{Creating Continuous Surfaces in 3D point cloud}

\begin{figure}[!t]
  \centering
  \scalebox{0.4}{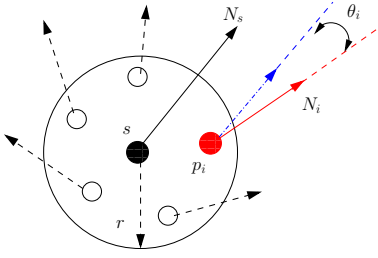}
  \caption{Defining an edge point. It is a point on the surface around
  which the surface normals are widely scattered in different
directions.}
  \label{fig:edgept}
\end{figure}

The method involves creating several surface patches in the 3D point
cloud using region growing algorithm \cite{vosselman2004recognising}
\cite{rabbani2006segmentation}. The angle between surface normals is
taken as the smoothness condition and is denoted by symbol $\theta$.  The
process starts from one seed point and the points in its
neighbourhood are added to the current region (or label) if the angle
between the surface normals of new point and that of seed point is
less than a user-defined threshold. Now the procedure is repeated with
these neighboring points as the new seed points. This process
continues until all points have been labeled to one region or the
other. The quality of segmentation heavily depends on the choice of
this threshold value.  A very low value may lead to over segmentation
and a very high value may lead to under segmentation. The presence of
sensor noise further exacerbates this problem leading to spurious
edges when only one threshold is used. This limitation of the standard
region growing algorithm is overcome by introducing a concept called
\emph{edge points} and using a pair of thresholds instead of one. The
use of two thresholds is inspired by a similar technique used in Canny
edge filter \cite{chen2008double} \cite{jie2012improved} and is
demonstrated to provide robustness against spurious edges. This
modified version of the region growing algorithm is described next in
the following section.

\begin{figure*}[!t]
  \centering
    \scalebox{0.4}{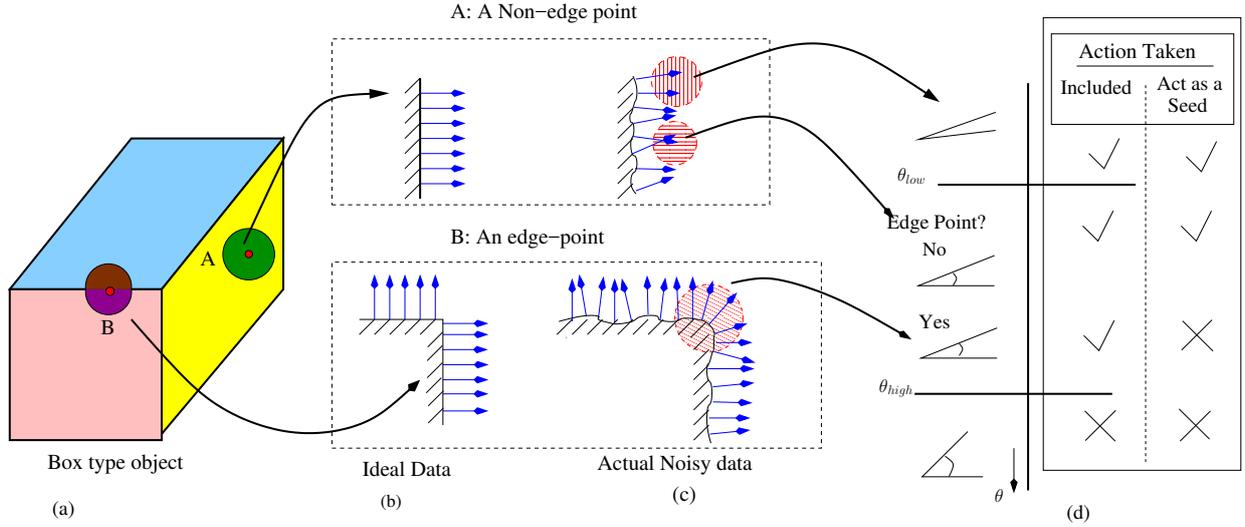} 
  \caption{Update criteria of proposed region growing is illustrated
  here a cuboid. Two cases are shown here,- one on a face (A) and
other on the boundary (B). Noisy data leads to error in normal
directions as shown in (c) compared to ideal input data (b) resulting
either spurious boundary or undetected boundary. Our method combines
the boundary condition with two thresholds (d) in order to achieve
better performance. }
  \label{fig:region_grow}
\end{figure*}

To begin, we first describe the concept of edge points and then,
explain how a pair of two thresholds on smoothness condition can
improve the performance of the standard region growing algorithm. Some
of the notations which will be used for describing the proposed method
are as follows. Also refer to Figure \ref{fig:edgept} for a better
understanding of these notations.  Let us consider a seed point $s \in \mathscr{C}$
with its own spherical neighborhood $\mathscr{N}(s)$ shown as a circle
in Figure \ref{fig:edgept}. It is further assumed that this
neighborhood consists of $m$ points ($p_i, i=1,2,\dots,m$) in the 3D
point cloud.  Mathematically, this neighborhood may be written as
follows: \begin{equation} \mathscr{N}(s) = \{p_i \in \mathscr{C}\;
  \Big | \; \lVert s-p_i \rVert \le r \};\ i = 1,2, \dots,m
  \label{eq:neighb} \end{equation} where $r$ is an user-defined radius
of the spherical neighborhood. Each neighboring point $p_i$ has an
associated surface normal $N_i$ which makes an angle of $\theta_i$
with the normal associated with the seed $N_s$.  As stated earlier,
$\theta_i$ is the smoothness condition for the region growing
algorithm. In this context, we define two thresholds $\theta_{low}$
and $\theta_{high}$ which are used for defining the region label for
the neighboring point and creating new seed for further propagation.
Let $Q_s$ be the set of new seeds which will be used in the next
iteration of the region growing algorithm. Before describing the
modification to the standard region growing algorithm, it is necessary to
introduce the concept of \emph{edge points} which is defined as
follows: \begin{defn}[Edge Point] Let $R(s)$ be a set of those
  neighbors $p_i$ of seed point $s$ for which $\theta_i >
  \theta_{high}$. In other words, \begin{equation} R(s) = \{p_i \in
      \mathscr{N}(s)\ | \ \theta_i > \theta_{high}, \ i = 1,2, \dots,
    m \} \label{eq:rs} \end{equation} Let $C_R$ be the cardinality of
  the set $R(s)$,i.e., $C_R = |R(s)|$.  Then, a seed point will be
  called as an \emph{edge point} if the following condition is
  satisfied: \begin{equation} \frac{C_R}{m} > k; \quad 0 < k < 1.0
    \label{eq:cond1} \end{equation} The set of all edge points for a
  given point cloud $\mathscr{C}$ be denoted by the symbol
  $\mathscr{E}$ and $\mathscr{E} \subset \mathscr{C}$.  \raggedright
  \hfill $\blacksquare$ \label{def:ep} \end{defn} The value of $k =
0.4$ is found to be empirically effective in providing better
segmentation of surfaces as will be shown later in this section.
Essentially, an edge point is a point on the edge of a surface where a
majority of its neighbors will have surface normals scattered in all
directions and for such a seed point, the neighboring points will have
angles $\theta_i>\theta_{high}$ as mentioned above.  One such edge
point is shown in Figure \ref{fig:region_grow} as point B. An edge
point is different from a non-edge point in the sense that the later
lies away from an edge and its neighbors have surface normals more or
less in the same direction. One such non-edge point is shown as point
A in Figure \ref{fig:region_grow}. Even with sensor noise, the
neighboring points around such a seed point will have surface normals
with smaller values of angles with respect to the surface normal of
the seed point, i.e., $\theta_i < \theta_{high}$.

Now, in the region growing algorithm starting with the seed point $s$,
the label $L\{p_i\}$ for a neighboring point $p_i \in \mathscr{N}(s)$
is defined as follows:
\begin{equation}
  \begin{array}{ccc}
    \textrm{if}\;  \theta_i < \theta_{low};  & \textrm{then,} &   L\{p_i\} = L\{s\} \; \wedge \; p_i \to Q_s  \\
    \textrm{if}\;  \theta_i > \theta_{high};  & \textrm{then,} & L\{p_i\} \ne L\{s\} \; \wedge \; p_i \not\to Q_s 
  \end{array}
  \label{eq:rg}
\end{equation}
where the notation $p_i\to Q_s$ indicates that the point $p_i$ is
added to the list of seed points which will be used by the region
growing algorithm in the next iteration.  However, if the angle between normals lies between the above two thresholds,
i.e., $\theta_{low} < \theta_i < \theta_{high}$, the label to the neighboring point is
assigned as follows: \begin{eqnarray}
  \textrm{if}\; s \notin \mathscr{E} & \textrm{then}, &  L\{p_i\} = L\{s\} \wedge  p_i \to Q_s \nonumber\\ 
  \textrm{if}\; s \in \mathscr{E} & \textrm{then} &  L\{p_i\} = L\{s\} \wedge p_i \not\to Q_s  
  \end{eqnarray}

\begin{figure}[!t]
  \centering
  \begin{tabular}{ccc}
\includegraphics[width=0.3\linewidth,height=0.4\linewidth]{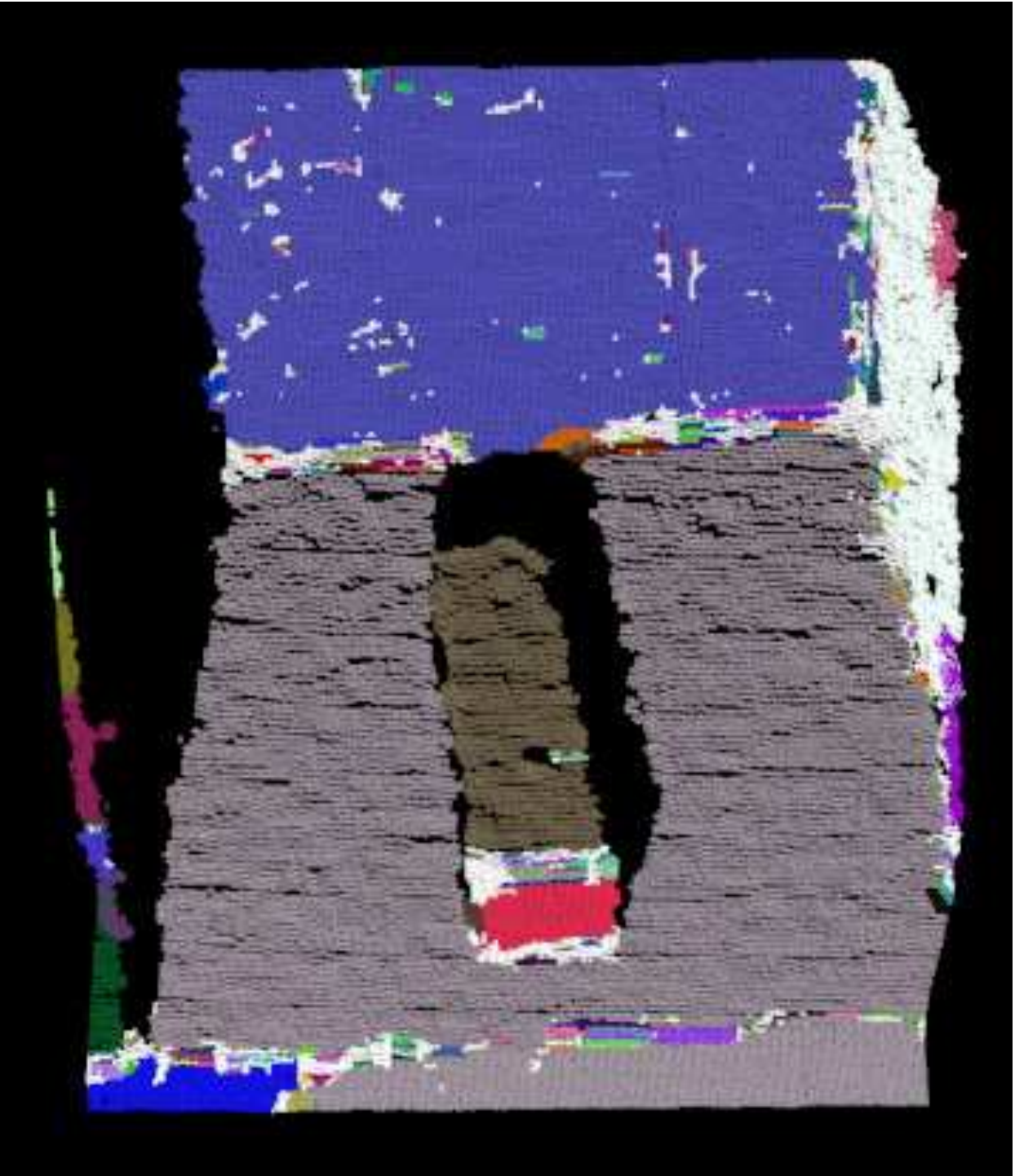} & 
\includegraphics[width=0.3\linewidth,height=0.4\linewidth]{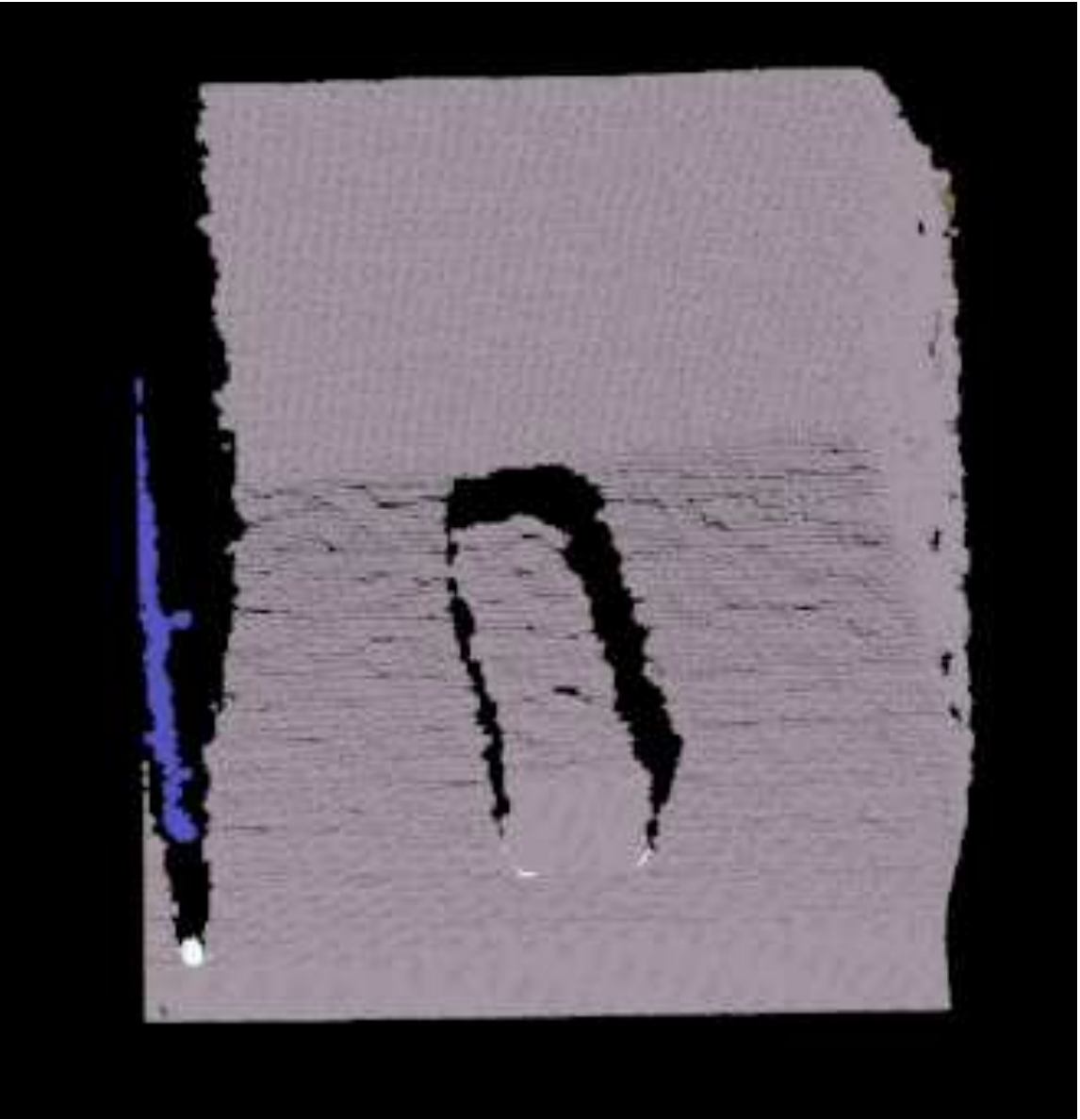} &
\includegraphics[width=0.3\linewidth,height=0.4\linewidth]{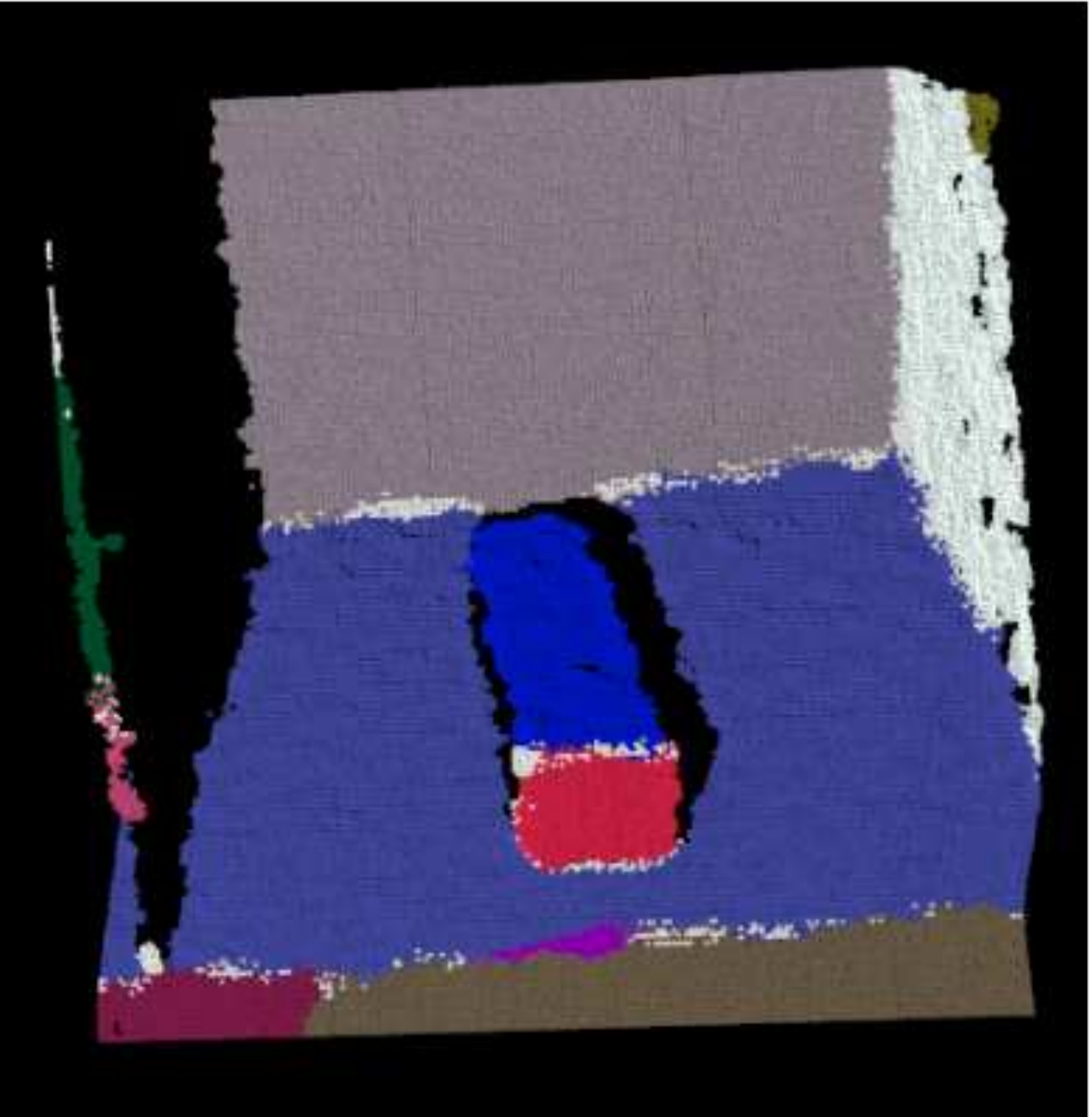} \\
\scriptsize{(a)} $\theta_{low}$ & \scriptsize{(b)} $\theta_{high}$ & \scriptsize{(c)} $\{\theta_{low},\theta_{high}\}$ \\
\includegraphics[width=0.3\linewidth,height=0.4\linewidth]{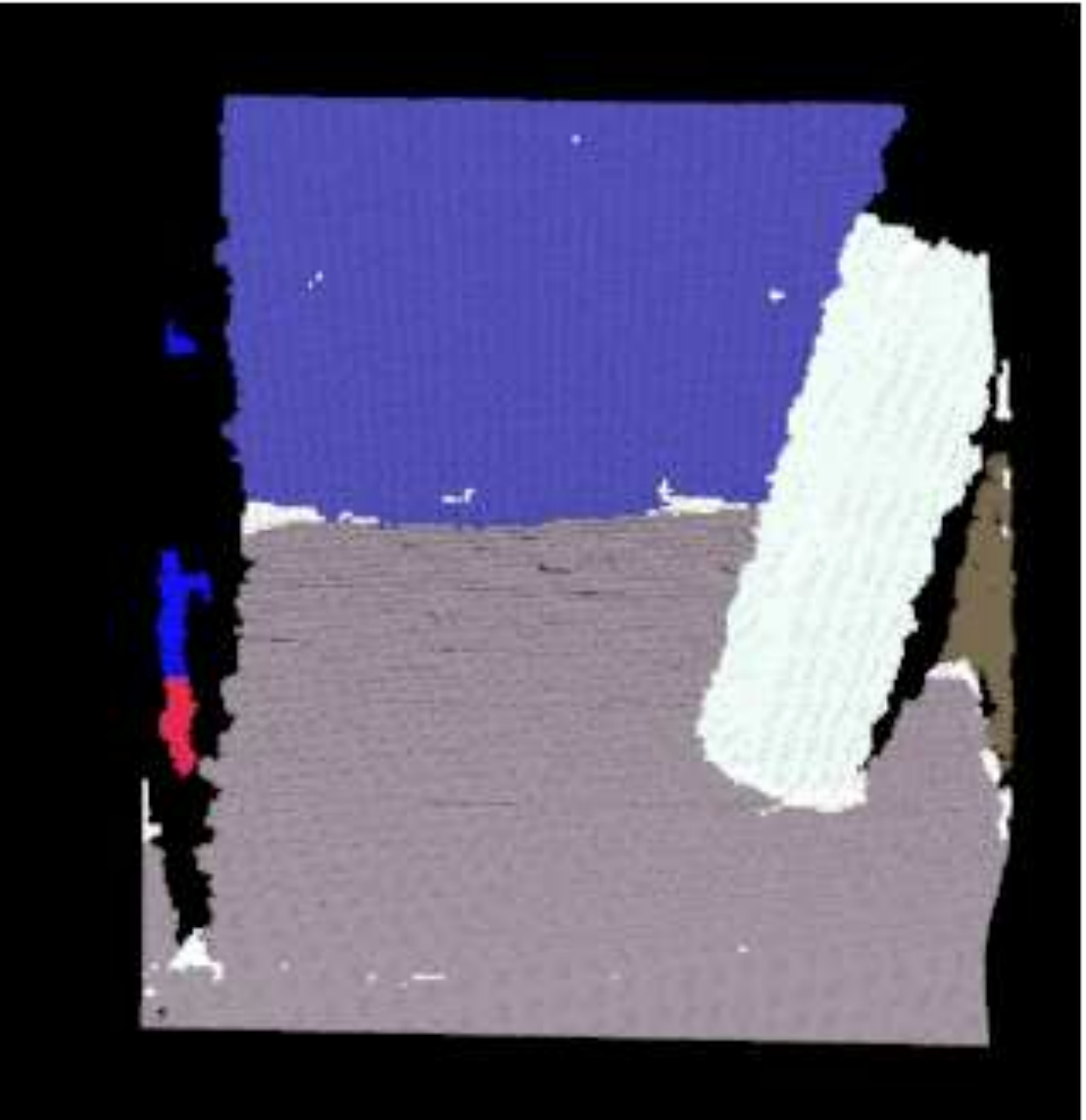} & 
\includegraphics[width=0.3\linewidth,height=0.4\linewidth]{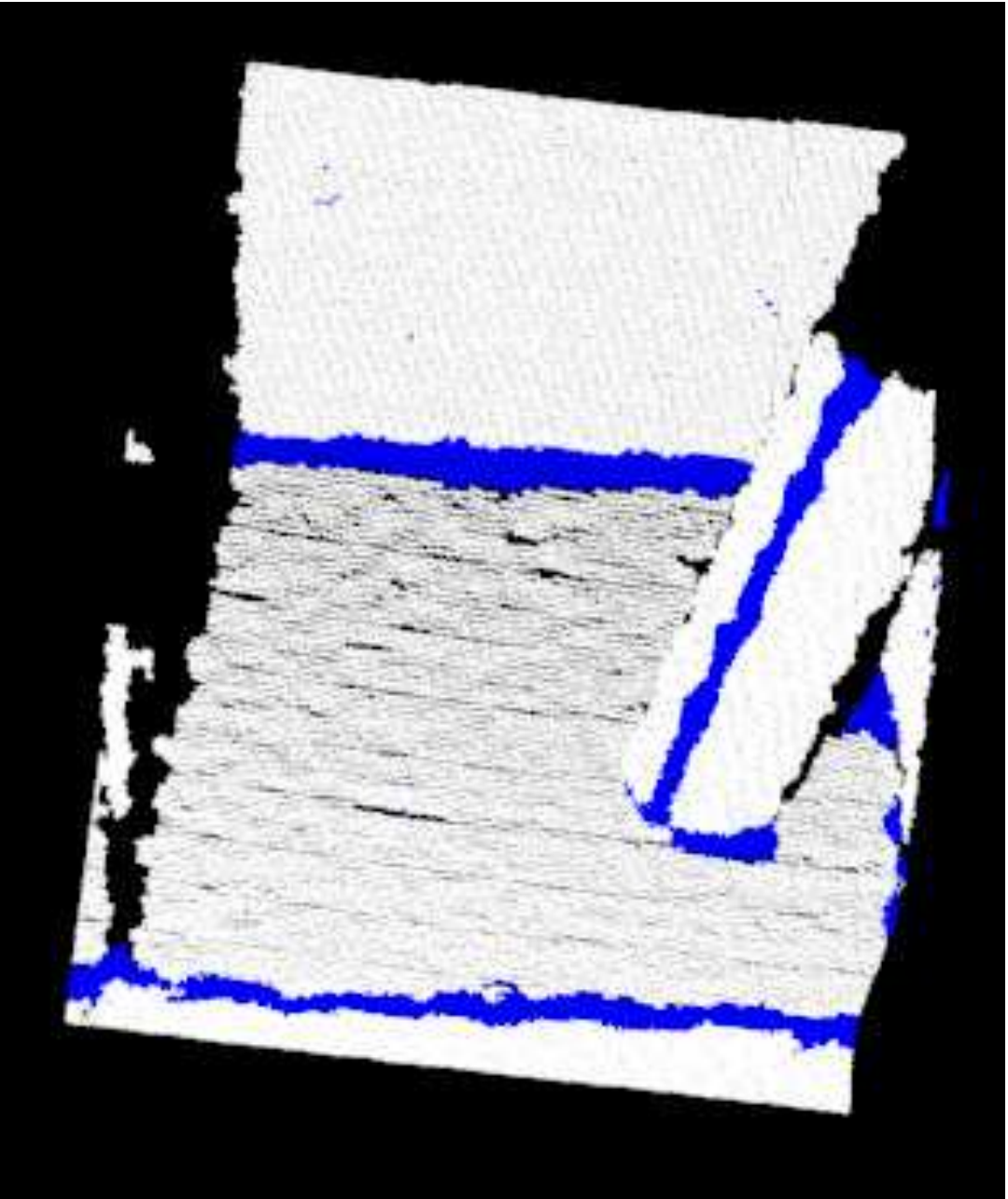} & 
\includegraphics[width=0.3\linewidth,height=0.4\linewidth]{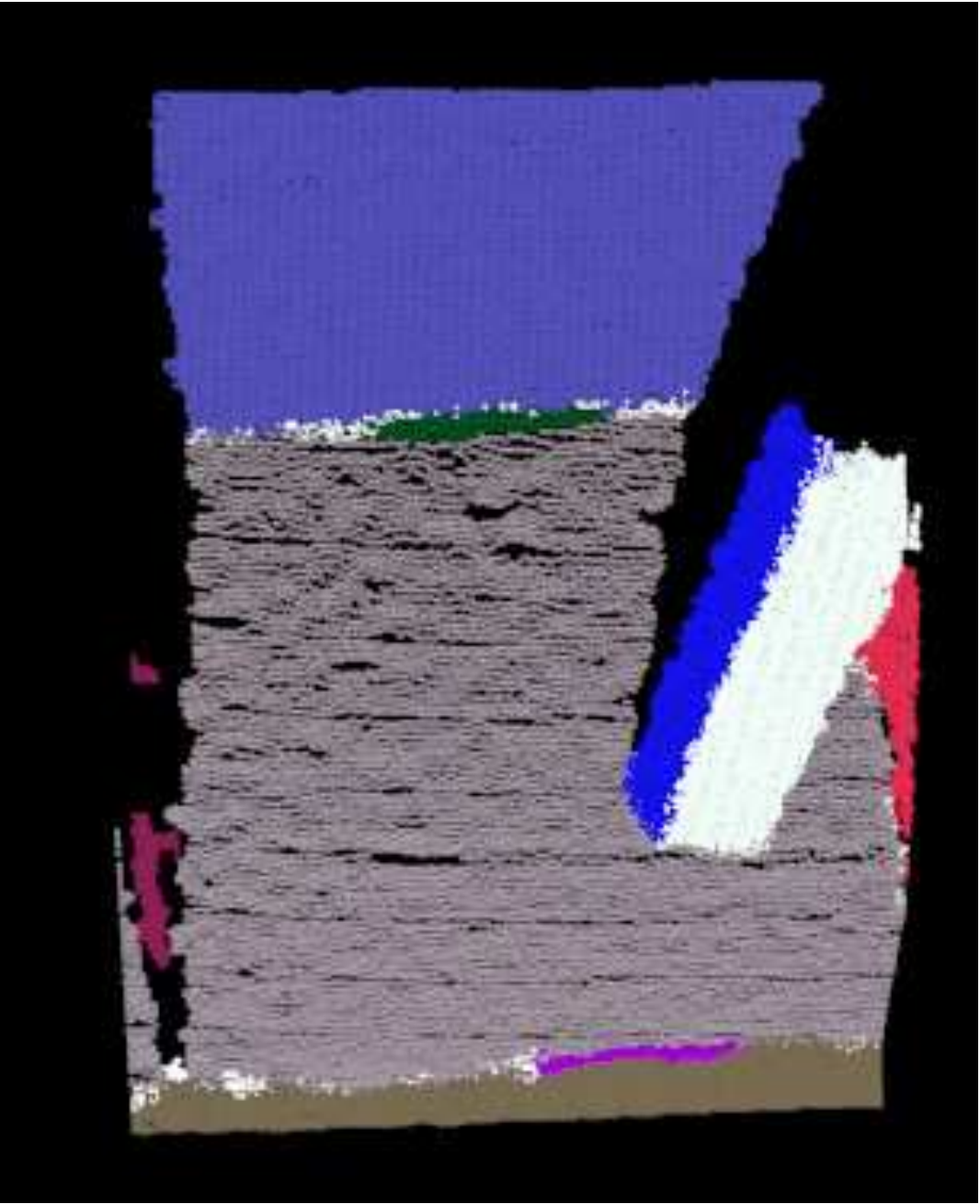} \\
\scriptsize{(d)} & \scriptsize{(e)} & \scriptsize{(f)} 
\end{tabular}
\caption{ Effect of double thresholds on smooth condition in region
  growing algorithm. (a) Using single threshold: $\theta_{low}$ leads
  to under-segmentation - multiple and discontinuous patches on the
  same surface (b) Using single threshold: $\theta_{high}$ leads to
  over-segmentation where different surfaces (having different
  normals) are merge together into a single surface. (c) using double
  boundary thresholds $ \{\theta_{low},\ \theta_{high}\}$ provides
  better surface segmentation compared to the case when only one
  threshold is used. (d) Shows the case when only one threshold is
  used. Two orthogonal surfaces of the object gets merged into one
continuous surface. (e) Shows the edge points in blue color (f) Shows
that the use of double thresholds lead to creation of two surfaces for
the rectangular object.}

\label{fig:rg_effect}
\end{figure}

 The above equation only states that while the neighboring point $p_i$
 is assigned the same label as that of the seed point $s$, it is not
 considered as a new seed point if the current seed is an edge point.
 This allows the region growing algorithm to terminate at the edges of
 each surface where there is a sudden and large change in the direction
 of surface normals thereby obtaining the natural boundaries of the
 objects.  The above process for deciding labels for neighboring
 points is demonstrated pictorially in Figure
 \ref{fig:region_grow}. It is also shown how a pair of thresholds
 are effective in dealing with sensor noise, thereby eliminating
 spurious edges. The effect of this modified version of region growing
 algorithm on the object segmentation can be seen clearly in Figure
 \ref{fig:rg_effect}. Figures \ref{fig:rg_effect}(a) and (b) shows the
 case of segmentation obtained with only one threshold. In the first
 case, a lower threshold cut-off value $\theta_{low}$ is used while in
 the later, upper cut-off threshold $\theta_{high}$ is used. As
 discussed earlier, lower value of threshold leads to
 under-segmentation and may generate multiple patches even on the same
 surface. On the other hand, higher value of thresholds leads to
 over-segmentation where different surfaces of a rectangular box may
 be identified as a single surface patch. In contrast to these two
 cases, the use of two thresholds provide better segmentation leading
 to creation of two separate surfaces one for each face of the
 rectangular box. 
 
This modified version of region growing algorithm allows us to find
graspable affordances for rectangular box-type objects which were
hitherto difficult. For instance, authors in
\cite{Pas2013LocalizingGA} \cite{ten2016localizing} find graspable
affordances only for objects with cylindrical or spherical shapes as
they relied on curve fitting methods. In \cite{pas2015using}
\cite{plattgrasppose2016}, authors use a trained SVM to identify
rectangular edges using HoG features and pre-defined hand poses were
used for grasping objects at these detected regions. Compared to these
approaches, the above proposed method is much simpler which does not
require any training phase and can be implemented in real-time. More
details about real-time implementation will be provided in the
experiment section later in this paper. The surfaces identified in
this section are then used to find valid graspable regions on the
object as described in the next section.

\section{Finding Graspable Affordances }
Once the surface segments are created, the grasping algorithm needs to
find suitable handles which could be used by the  gripper for picking
objects. This is otherwise known as the problem of grasp pose
detection \cite{plattgrasppose2016}  which essentially aims at finding
six dimensional pose for the gripper necessary for making a stable
grasping contact with the object. However, this is a computationally
intensive task as one has to search in a 6-dimensional pose space.
The searching procedure is broadly handled in two ways. In one
approach, the object to be picked is matched with its CAD model. Once
a match is found, then the geometric parameters of the object model is
used to compute the 6 DOF gripper pose directly.  As CAD models may
not always be available, the objects are generally approximated with
some basic shape primitives \cite{jain2016grasp}
\cite{somani2014shape} \cite{behnke2012shape} or superquadric
\cite{vezzani2017grasping} models. While these methods take 3D point
cloud as input, other methods can work with RGBD data.  They generally
take color and depth information as image and apply a sliding window
based search with different scale to find valid grasping regions
\cite{lenz2015deepgrasp} \cite{johns2016deep}. We simplify this search
problem at first by grouping similar type of points based on the
boundaries obtained in the previous step and, by making some practical
assumptions about the grasping task. As described earlier in section
\ref{sec:sn}, the gripper is assumed to approach the object in a
direction opposite to surface normal of the object. It is also assumed
that the gripper closing plane coincides with the minor axis of the
surface segment under consideration as shown in Figure
\ref{fig:gripper_object}(b). In this way, the 6D pose problem is
solved in a single step and can be implemented in real-time. However,
it is still necessary to identify suitable regions on the surface
segments that can fit within the fingers of the gripper while ensuring
that the gripper does not collide with neighbouring objects.  In other
words, one still needs to search for a three-dimensional cube of
dimension $l\times b \times e$ around the centroid of the object
segment as shown in Figure \ref{fig:gripper_object}(b).  This requires
carrying out a linear search along the three principal axes of the
surface to find regions that meet this bounding box constraint.  These
regions are the graspable affordances for the object to be picked by
the gripper. The details of the search process is described next in
this section.  

Let us assume that the region growing algorithm, described in the
previous section, leads to the creation of $S$ segments in the 3D
point cloud $C \in \mathscr{R}^3$. As a first step, we extract the
following parameters for each of these segments $s = 1,2,\dots,S$:                            

\begin{itemize}
  \item The centroid of the segment: $\boldsymbol{\mu}_s = [\mu_x^s, \mu_y^s,\mu_z^s]$.
  \item The associated surface normal vector:  $\hat{n}_s \in \mathscr{R}^3$.
  \item First two dominant directions obtained from Principle
    Component Analysis (PCA)  and their corresponding lengths. These
    two axes correspond to vectors $\hat{a}$ and $\hat{f}$
    respectively in Figure \ref{fig:gripper_object}(b). 
\end{itemize}

\begin{figure}[!t]
\scalebox{0.5}{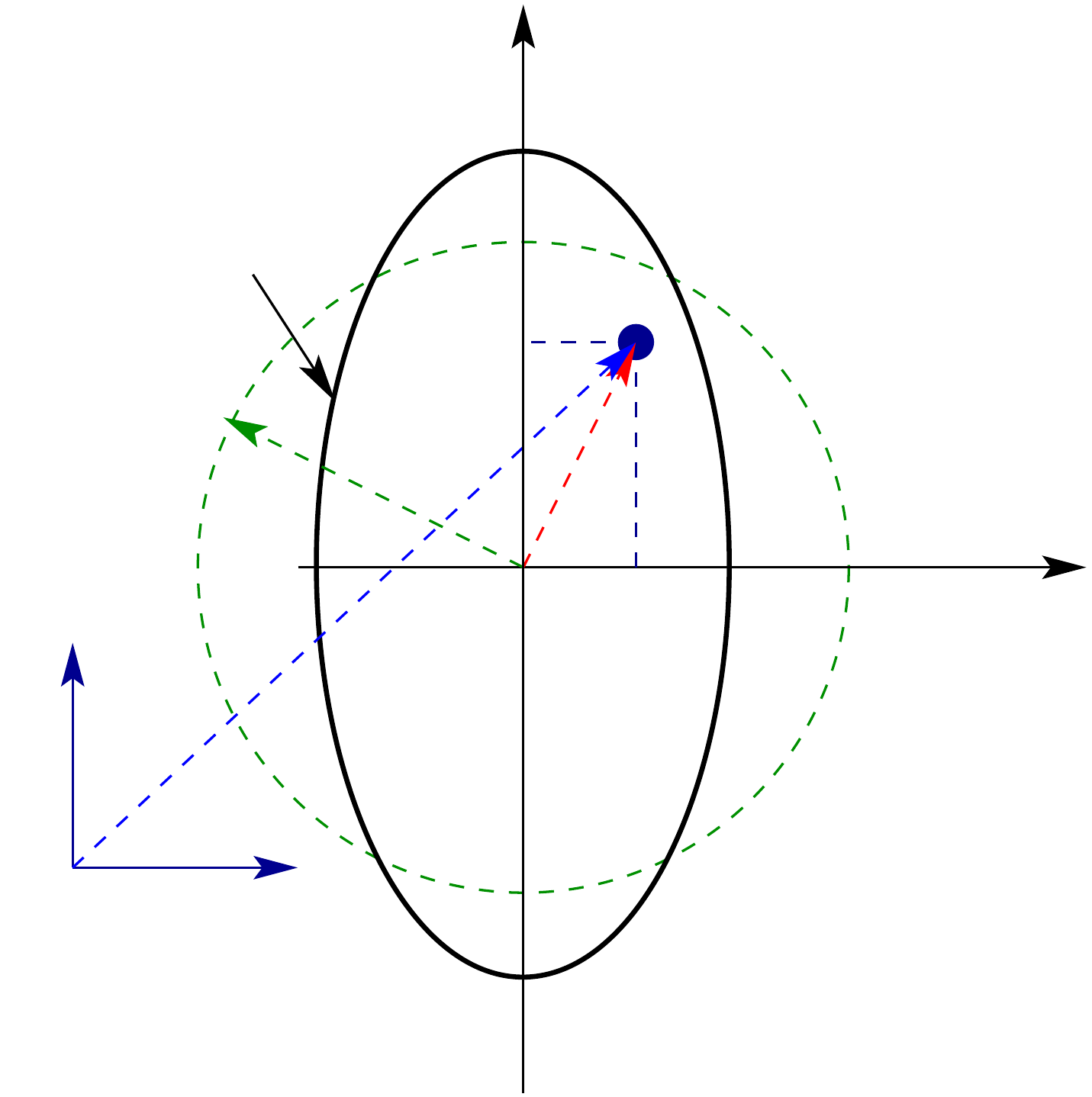}
\caption{Scalar Projection of 3D cloud points to a new coordinate
  frame. The 3D point cloud points on the surface segment $s$
  represented by $\vec{p}_O$ within the sphere of radius $d/2$ is
  projected onto the axes of the new coordinate frame
  $(\hat{f},\hat{a},\hat{n})$ represented by the vector
  $\vec{q}_{O'}$. These projected points are used for finding 
  suitable graspable affordances for the object. The axes $\hat{z}$
  and $\hat{n}$ are perpendicular to the plane of the paper and point
outward.}
\label{fig:scalproj}
\end{figure}

The search for suitable handles starts from the centroid $\mu_s$ of
the surface and proceeds along the three principal axes, i.e., major
axis $\hat{a}$, minor axis $\hat{f}$ and surface normal $\hat{n}$. In
order to do this, the 3d point in the original point cloud
corresponding to the surface segment under consideration $s$ are
projected onto these new axes ($\hat{f},\hat{a},\hat{n}$) as shown in
Figure \ref{fig:scalproj}. So for every point $\vec{p}_O =
(x_1,y_1,z_1)_O$ in the orginal coordinate system
$(\hat{x},\hat{y},\hat{z},O)$ that lies within a sphere of radius
$d/2$ results in a vector $\vec{q}_{O'} = (f_1,a_1,n_1)_{O'}$ in the
new coordinate system ($\hat{f},\hat{a},\hat{n},O'$). The radius of
the sphere is selected to be half of the maximum hand aperture of the
gripper to be used for picking the object. The third axes $\hat{n}$
and $\hat{z}$ are normal to the surface of the paper and hence is not
displayed in the figure. 

\begin{figure}[!t]
\centering
\begin{tabular}{cc}
\includegraphics[width=0.45\linewidth,height=0.4\linewidth]{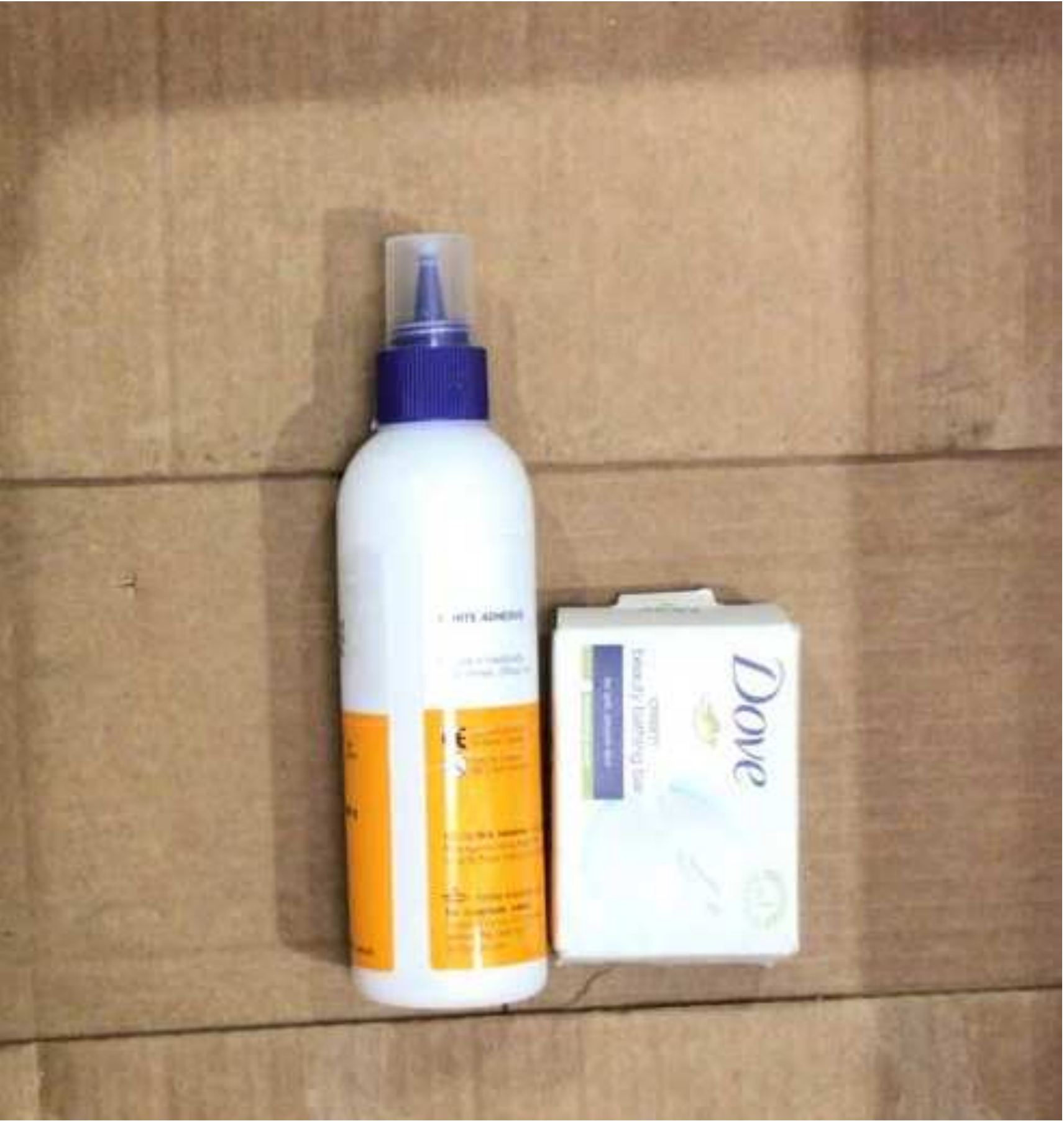} &
\includegraphics[width=0.45\linewidth,height=0.4\linewidth]{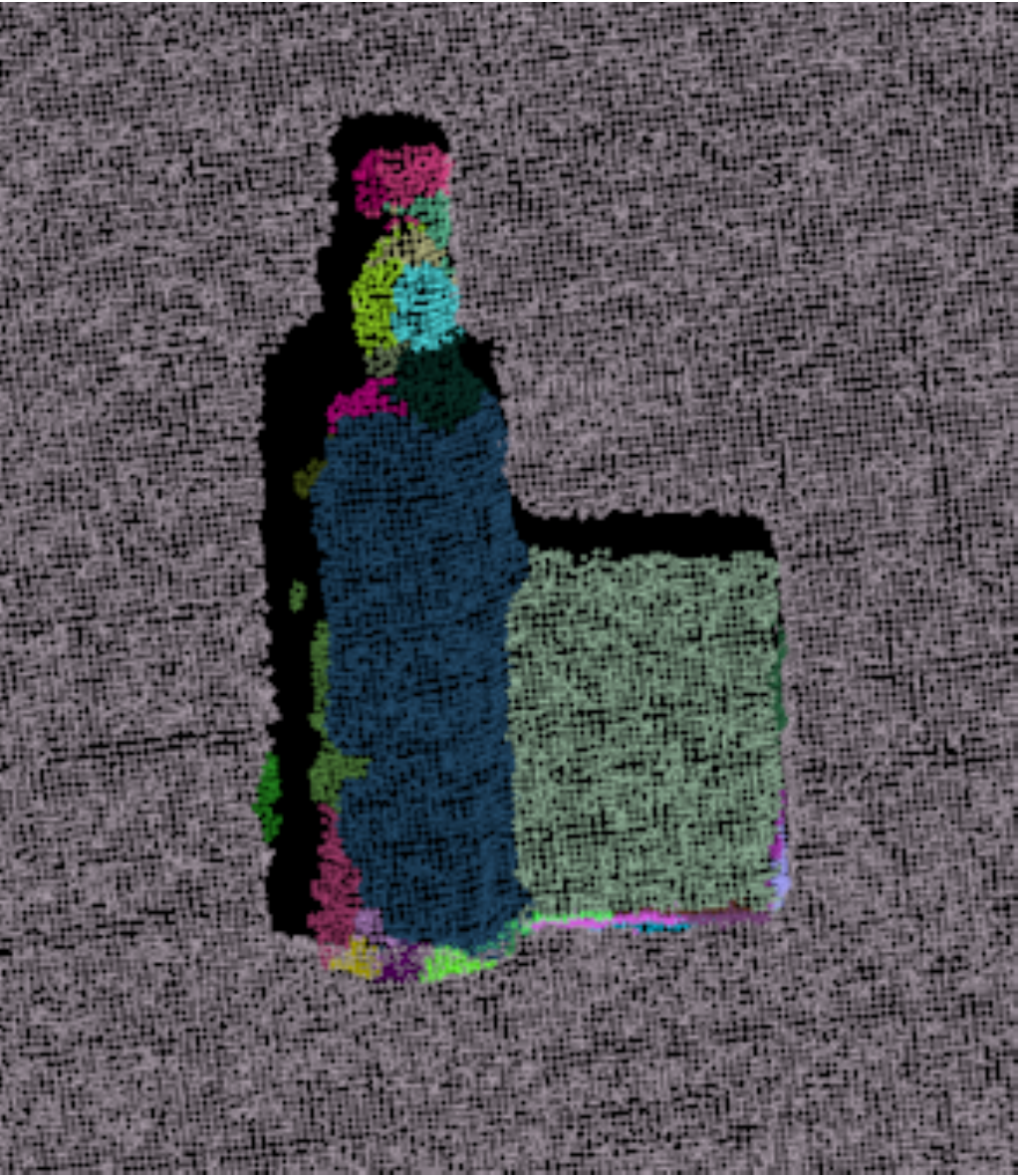} \\
\scriptsize{(a)} & \scriptsize{(b)} \\
\includegraphics[width=0.45\linewidth,height=0.4\linewidth]{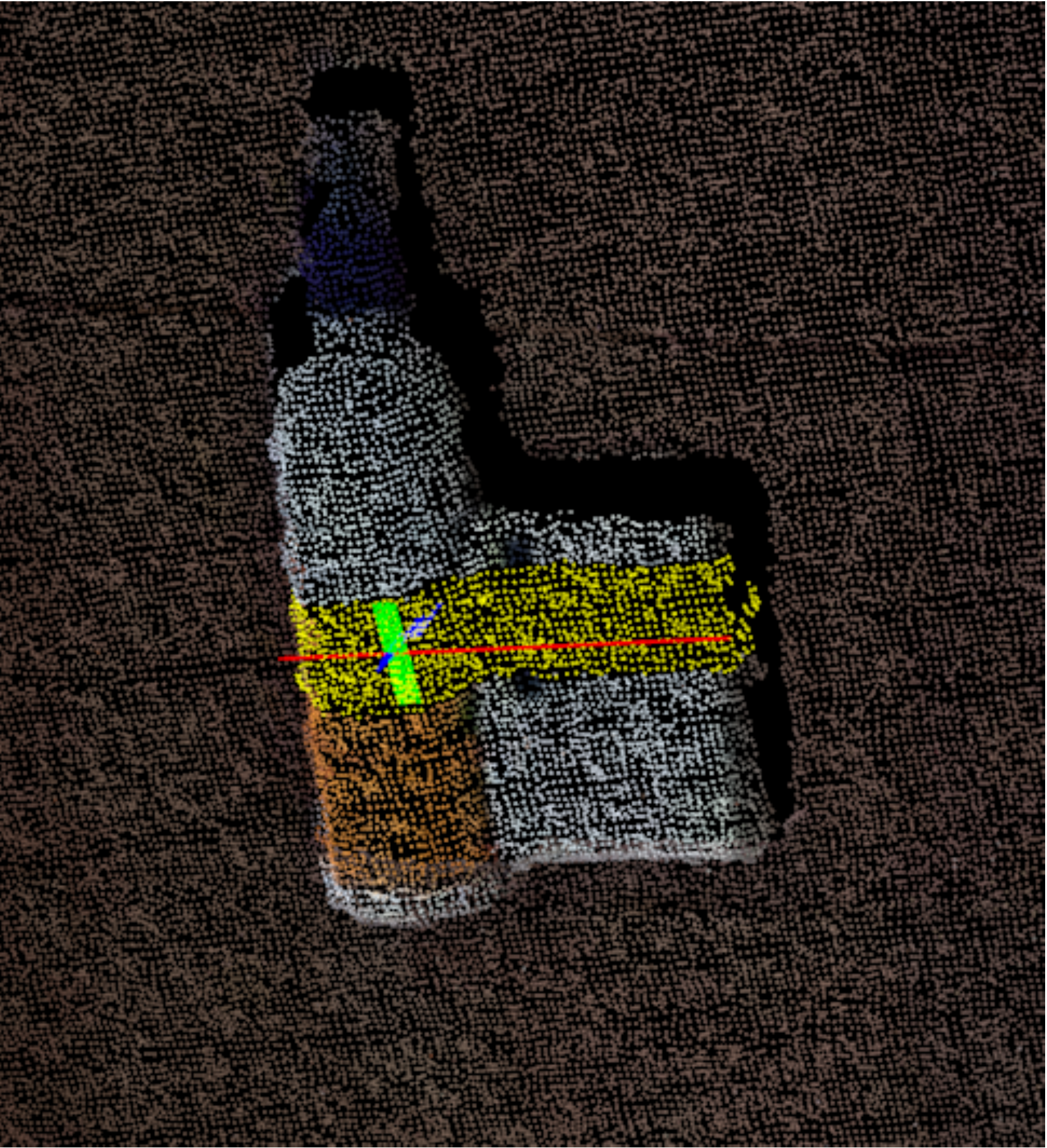} &
\includegraphics[width=0.45\linewidth,height=0.4\linewidth]{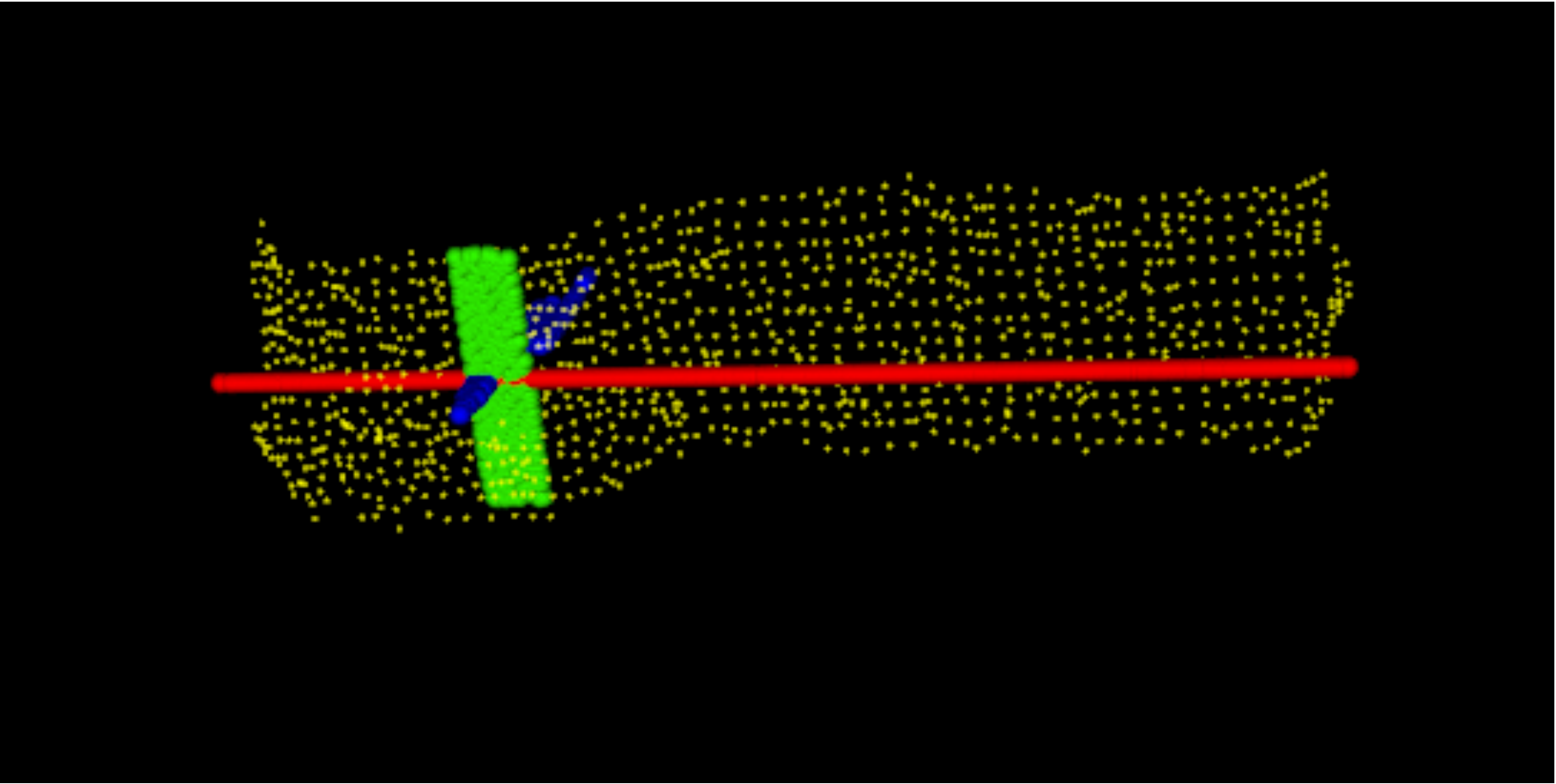} \\
\scriptsize{(c)} & \scriptsize{(d)} \\
\includegraphics[width=0.45\linewidth,height=0.4\linewidth]{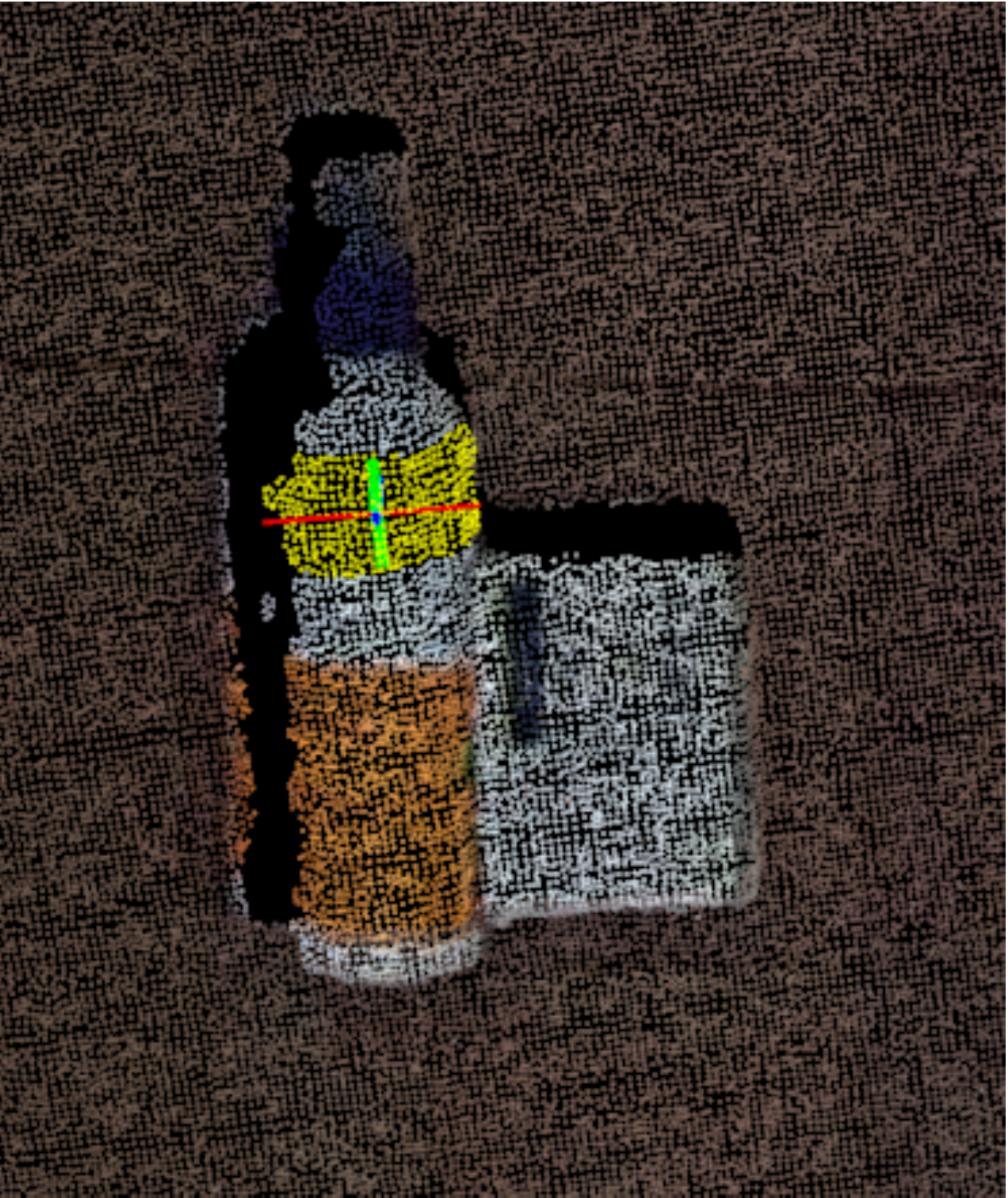} &
\includegraphics[width=0.45\linewidth,height=0.4\linewidth]{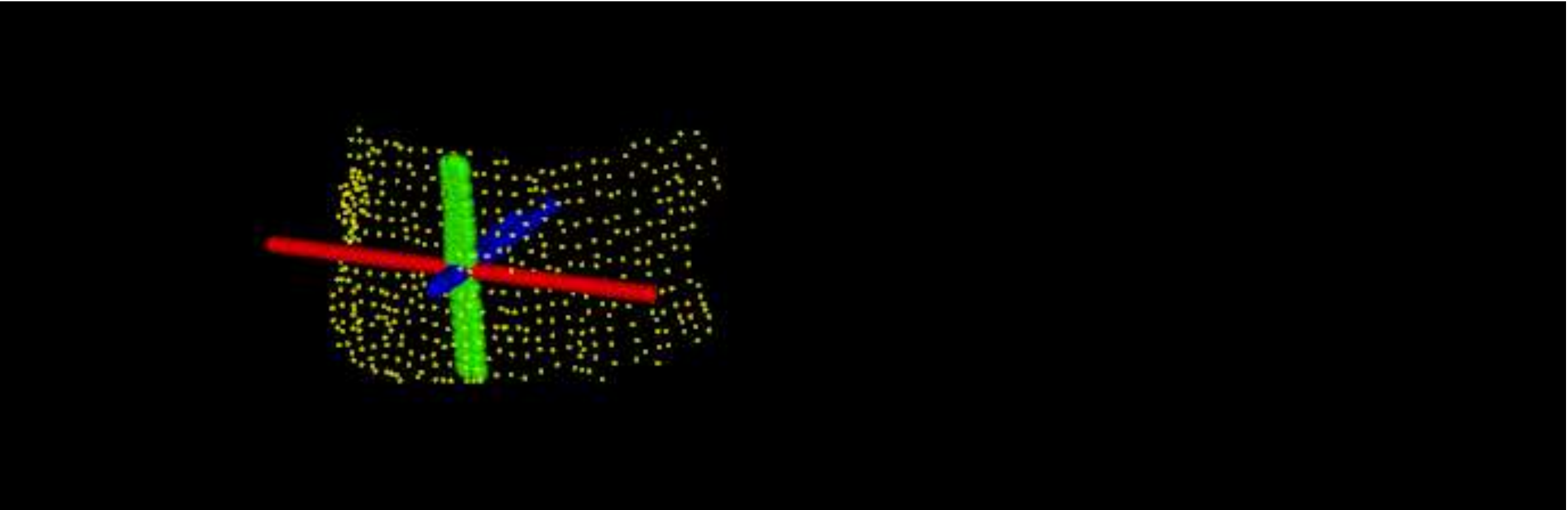} \\
\scriptsize{(e)} & \scriptsize{(f)} \\
\includegraphics[width=0.45\linewidth,height=0.4\linewidth]{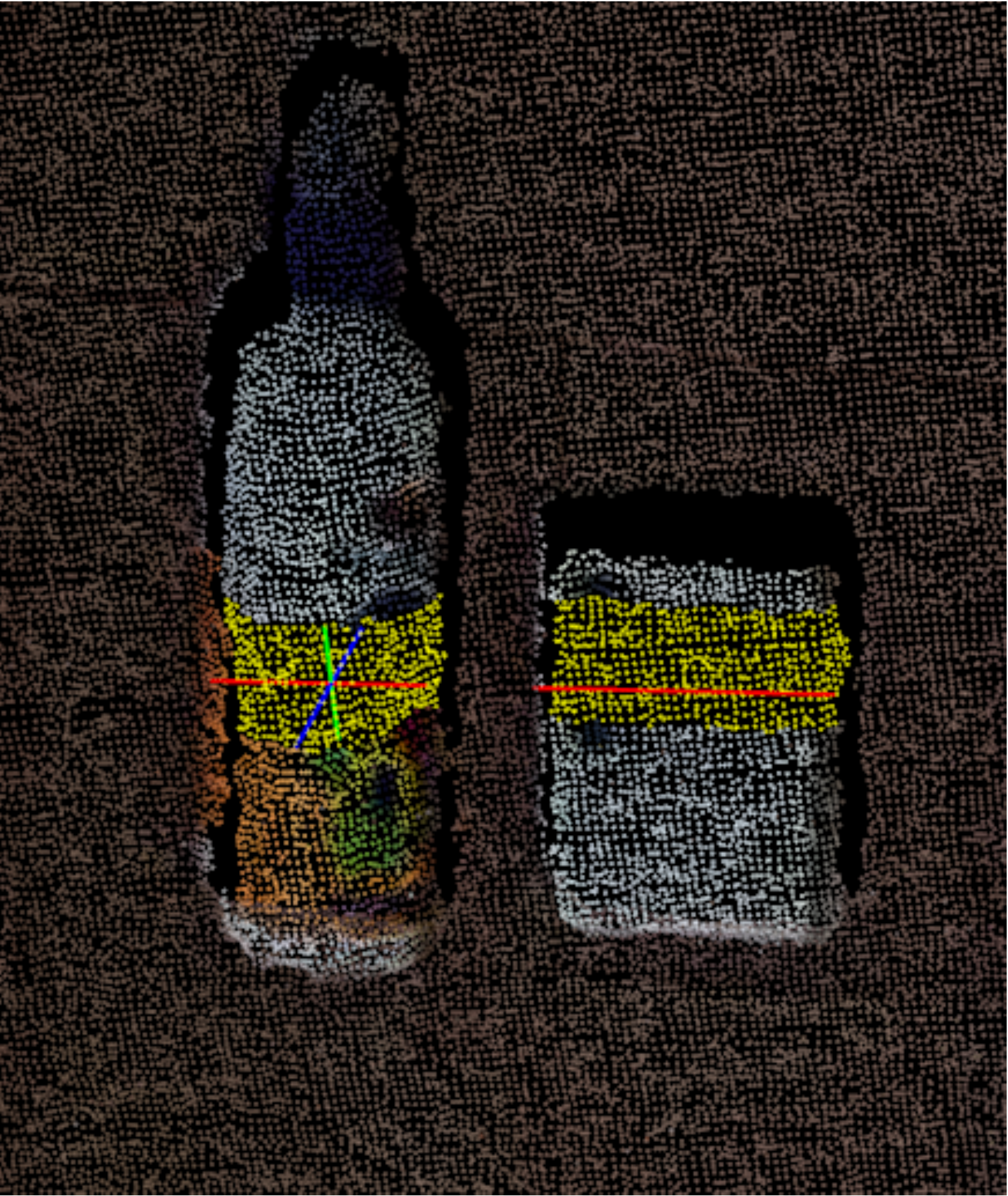} &
\includegraphics[width=0.45\linewidth,height=0.4\linewidth]{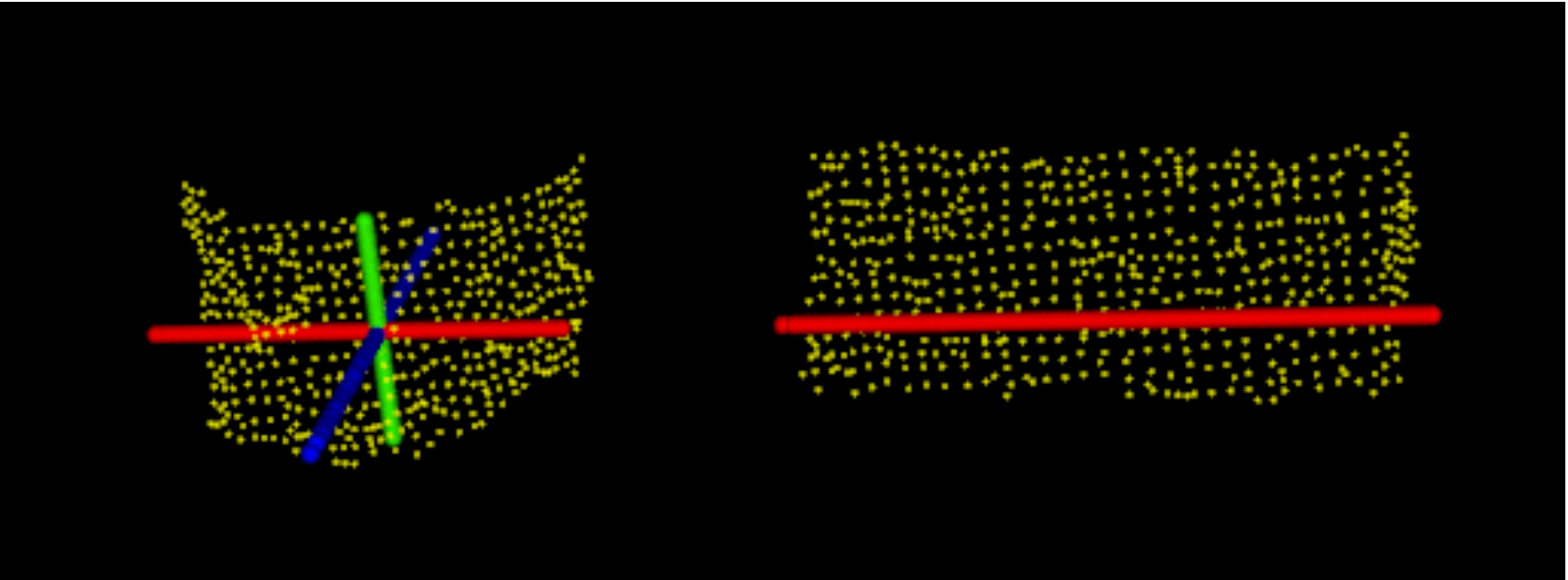}\\
\scriptsize{(g)} & \scriptsize{(h)} 
\end{tabular}
\caption{Searching for suitable grasping handle. (a) Actual picture for a case where objects are stacked very close to each other (b) Segmented point cloud obtained after applying region growing on surface normals; (c) - (d) Suitable handle is not found as discontinuity is detected in the horizontal axis (e) - (f) suitable handle is found for another patch on the same object; (g)-(h) suitable handle found when objects are separate as a discontinuity is detected along the red axis }
\label{fig:linsearch}
\end{figure}

Through this scalar projection, the three dimensional search problem
is converted into three one-dimensional search problems, which is
computationally much simpler compared to the former. The search is
first performed along the direction $\hat{a}$ and $\hat{n}$
respectively. All the points that lie within the radius $e/2$ around
the centroid $\mu_s$ is considered to be a part of the gripper handle.
Similarly, all points of the surface that lie within the radius of $l$
along a direction of $-\hat{n}$ is considered to be part of the
gripper handle. Please note that $e$ and $l$ are the width and the
length of gripper fingers needed for holding the object. Once these
two boundaries are defined, we get a horizontal patch of points
extending along the minor axis $\hat{f}$ as shown in Figure
\ref{fig:linsearch} (c), (e) and (f). So now, we need to find the
boundary along the minor axis to see if it would fit within the
gripper finger gap.  This is done by searching for a gap along the
minor axis $\hat{f}$ which is at least bigger than a given user
defined threshold which itself depends on the thickness of the gripper
finger. The idea is that there should be sufficient gap between two
objects to avoid collision with the neighboring objects. This is
illustrated in Figure \ref{fig:linsearch}. The working of the search
process could be understood by analyzing this figure as explained in
the following paragraph. 

The figure \ref{fig:linsearch}(a) shows two objects which has been
kept adjacent to each other such that their boundaries touch each
other. The figure (b) shows the surface segments obtained using the
proposed region growing algorithm. The objective is to find a suitable
graspable handle for the cylinder object. The figure (c) shows the
horizontal patch obtained using the linear search as explained above.
Since there is no gap along the minor axis (shown in red), the region
belonging to both the objects within the yellow band gets included
into the graspable region. Total horizontal length of this band may
exceed the maximum hand aperture $d$ of the gripper making it an
invalid grasping handle for the object. Now the next band of width $e$
on the top of the last band is taken into consideration. In this case,
a gap is found immediately around the boundary of the cylindrical
surface along the minor axis as shown in Figure
\ref{fig:linsearch}(e). Since this length along the red axis fits
within the gripper handle, it will be considered as a valid handle for
the object. The figures \ref{fig:linsearch}(g)-(h) shows the case when
these two objects have been kept apart. In this case, the gap is found
along the minor axis and hence the handle for the bottle is detected
successfully without any further search.  Hence the search process
involves four steps: 

\begin{enumerate}
\item Project all the points on the surface segment within a spherical radius of $d/2$ onto the axes $\hat{a}$, $\hat{f}$ and $\hat{n}$. 
\item Fix boundary along the major axis $\hat{a}$ at a distance of $e/2$ on either side of the centroid the patch under consideration. 
\item Fix boundary along the normal axis $-\hat{n}$ at a distance of $l$ from the top surface. 
\item Search for gap along the minor axis $\hat{f}$ on either side of the centroid. If this gap is greater than or equal to $g$, then search is stopped. The resulting patch is considered a valid grasping handle for the object if the total length of the patch along the minor axis is less than  maximum hand aperture $d$ of the gripper. 
\end{enumerate}

A new patch along the major axis either side of the centre patch is
analyzed for validity in case the current one fails to satisfy the
gripper constraints. So it is possible to obtain multiple handles on
the same object, which is very useful, as the robot motion planner may
not be able to provide a valid end-effector trajectory for a given
graspable affordance. The gripper approaches the object at the
centroid of the yellow patch shown in Figure \ref{fig:linsearch}(c) or
(e) along the direction of surface normal (shown in blue color in
Figure \ref{fig:linsearch} (d) or (f) respectively, towards the object
with its gripper closing plane coinciding with the minor axis (show in
red color).  As one can appreciate, the pose detection problem is
solved by a simple method that converts a 6D search problem into a
simple 1-D search problem. This is much faster computer other method
such as \cite{vezzani2017grasping} that use complex optimization methods to arrive at
the same conclusion. The proposed provides remarkable improvement over
the state of the art method \cite{Pas2013LocalizingGA} \cite{ten2016localizing} which
provides much inferior performance in a cluttered environment as will
be shown in the next section. 

\section{Experimental Results} \label{sec:res}

\begin{figure}[!t]
  \centering
\includegraphics[width=0.2\linewidth,height=0.2\linewidth]{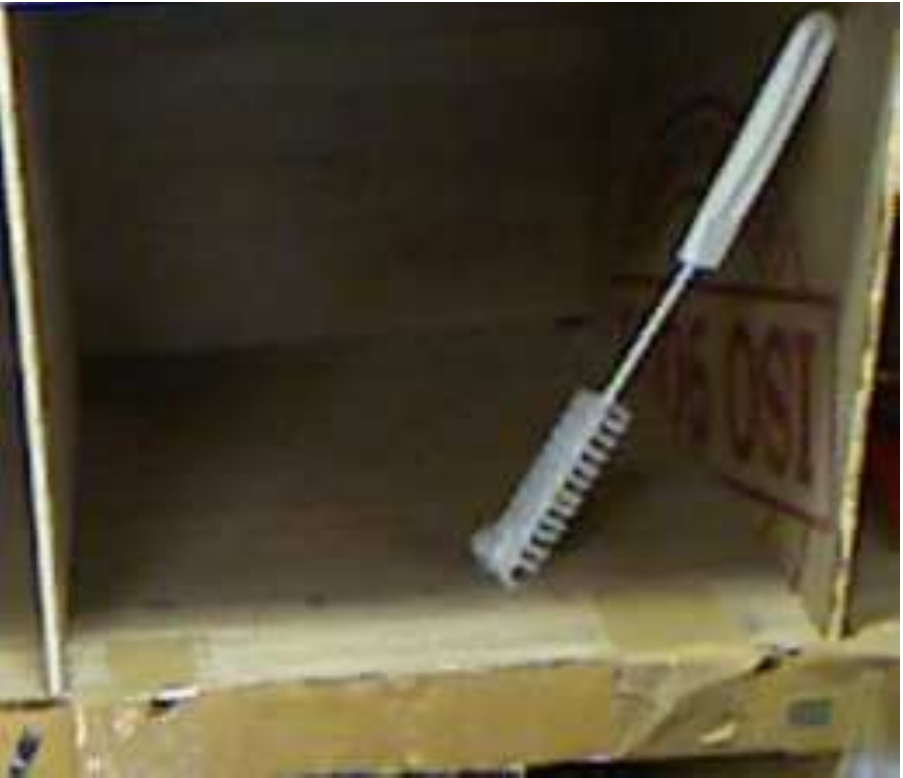}  
\includegraphics[width=0.2\linewidth,height=0.2\linewidth]{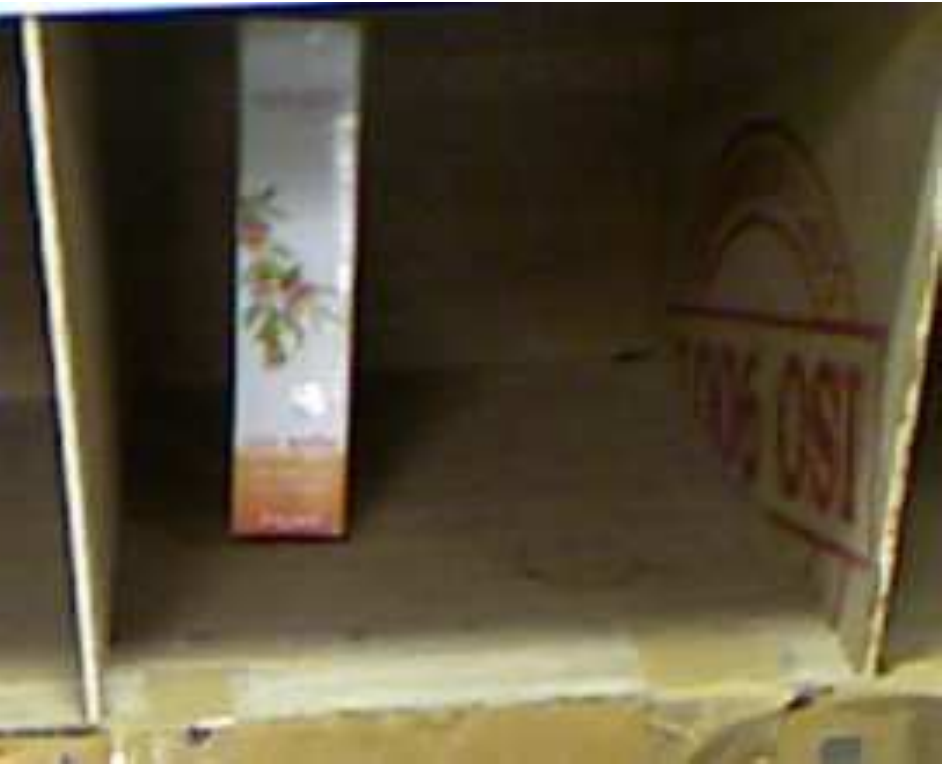} 
\includegraphics[width=0.2\linewidth,height=0.2\linewidth]{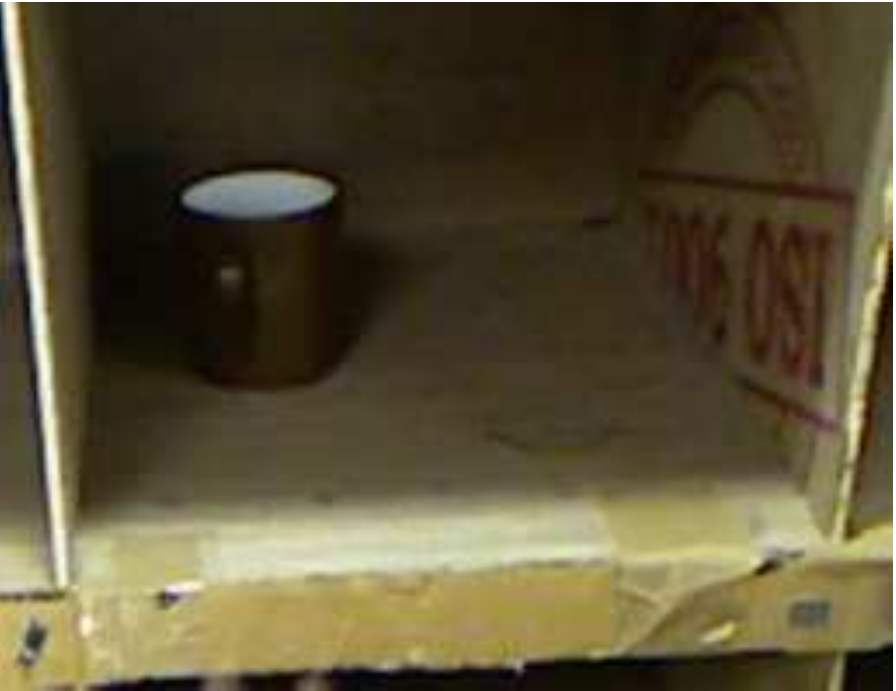} 
\includegraphics[width=0.2\linewidth,height=0.2\linewidth]{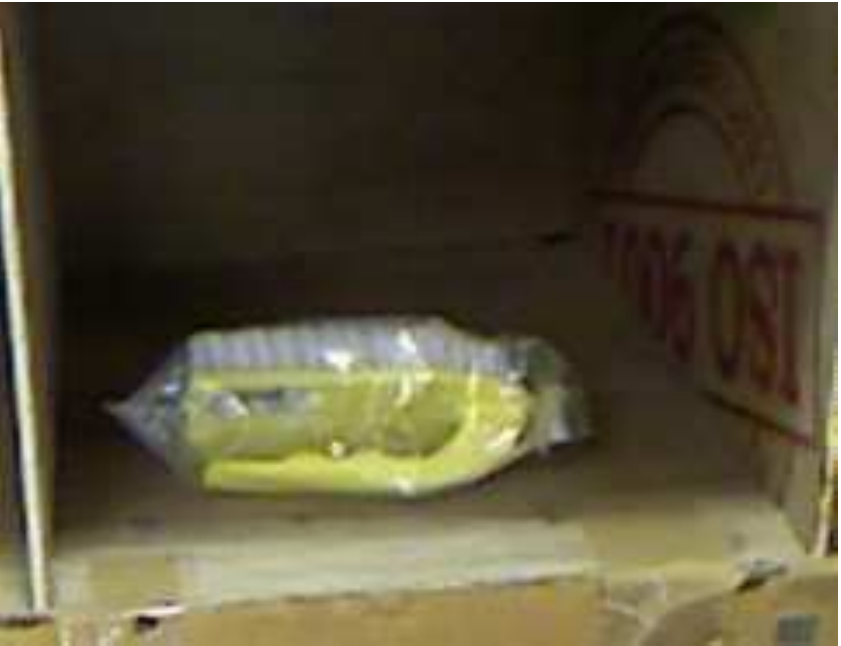} \\
\includegraphics[width=0.2\linewidth,height=0.2\linewidth]{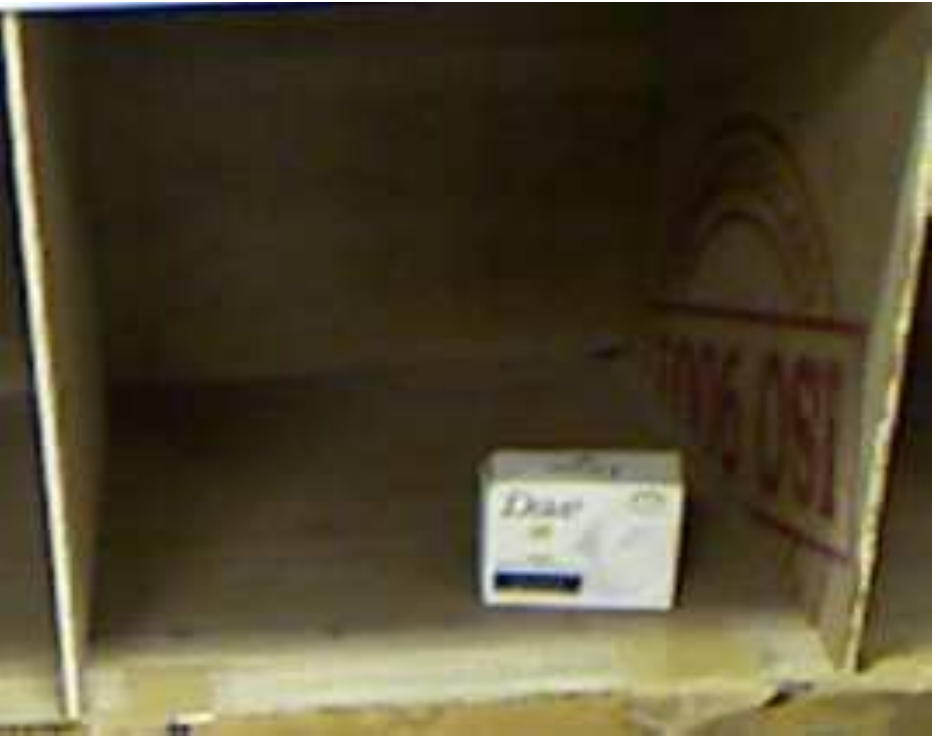}  
\includegraphics[width=0.2\linewidth,height=0.2\linewidth]{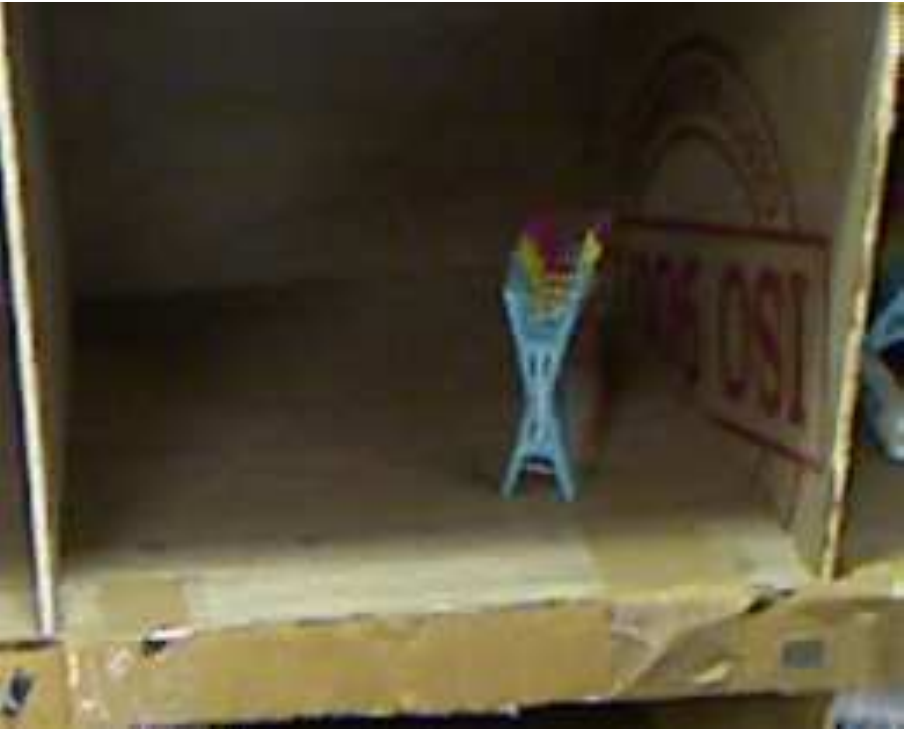} 
\includegraphics[width=0.2\linewidth,height=0.2\linewidth]{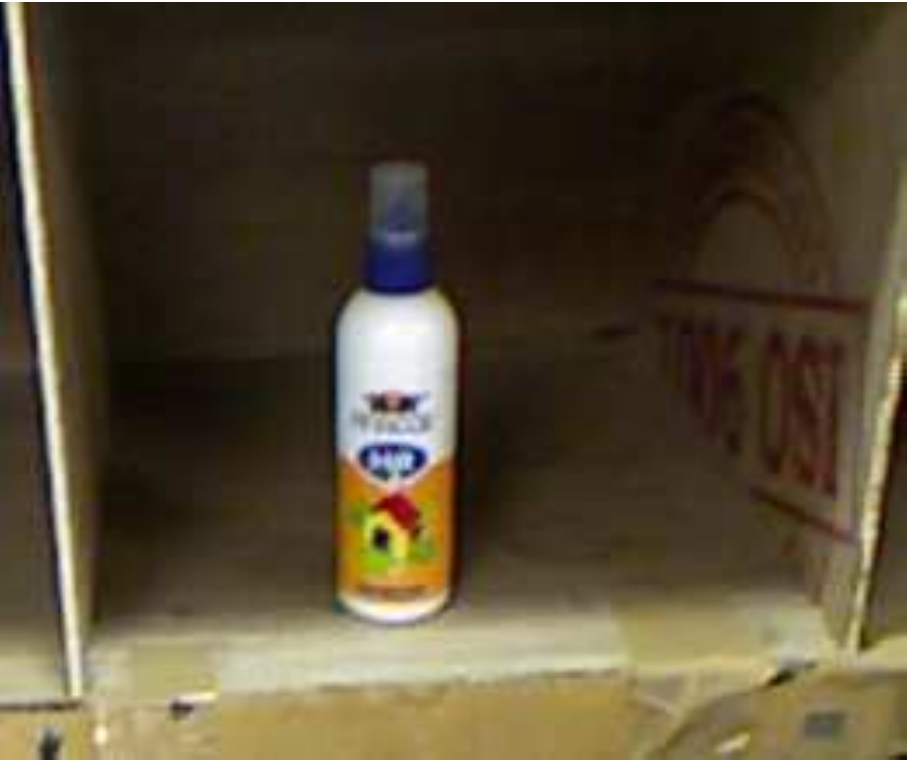} 
\includegraphics[width=0.2\linewidth,height=0.2\linewidth]{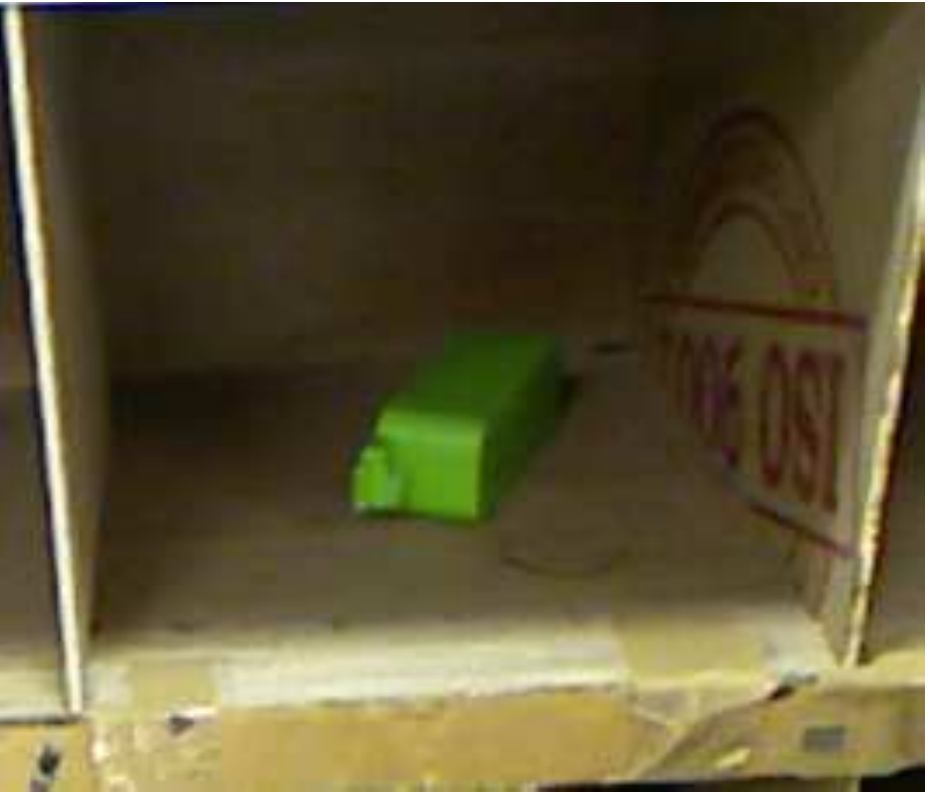} \\
\includegraphics[width=0.2\linewidth,height=0.2\linewidth]{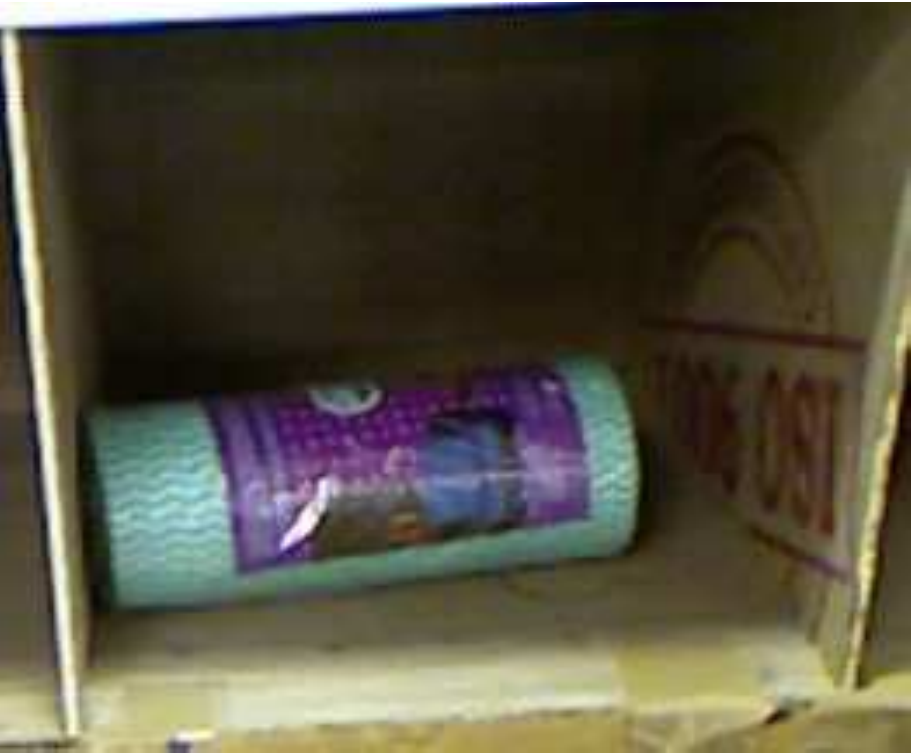}  
\includegraphics[width=0.2\linewidth,height=0.2\linewidth]{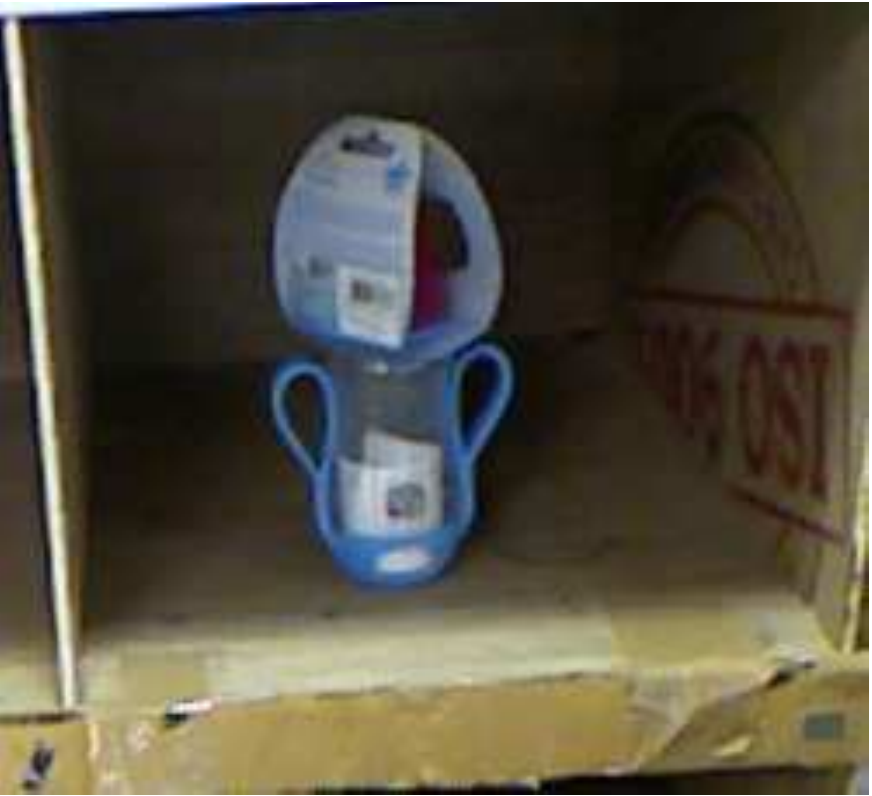} 
\includegraphics[width=0.2\linewidth,height=0.2\linewidth]{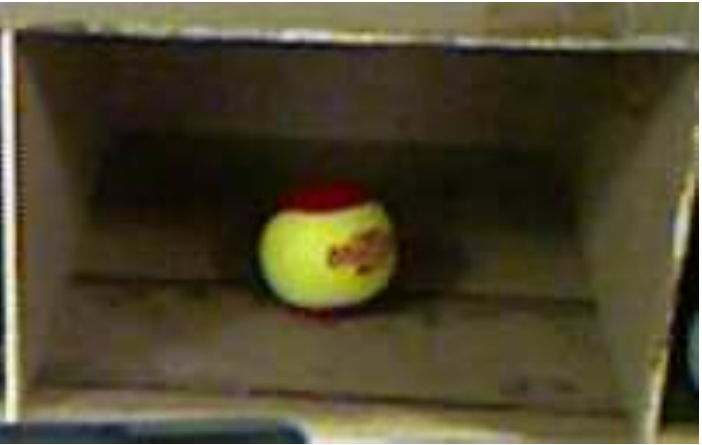} 
\includegraphics[width=0.2\linewidth,height=0.2\linewidth]{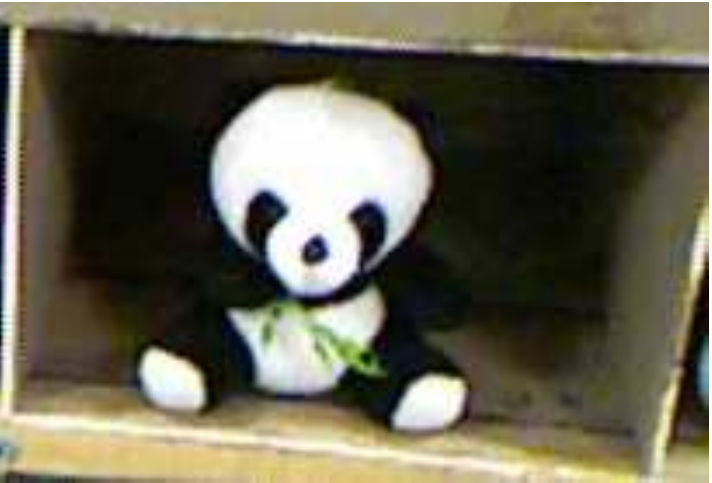} \\
\scriptsize{(a) Snapshot of TCS Grasp Dataset 1} \\   \vspace{0.3cm}
\includegraphics[width=0.3\linewidth,height=0.2\linewidth]{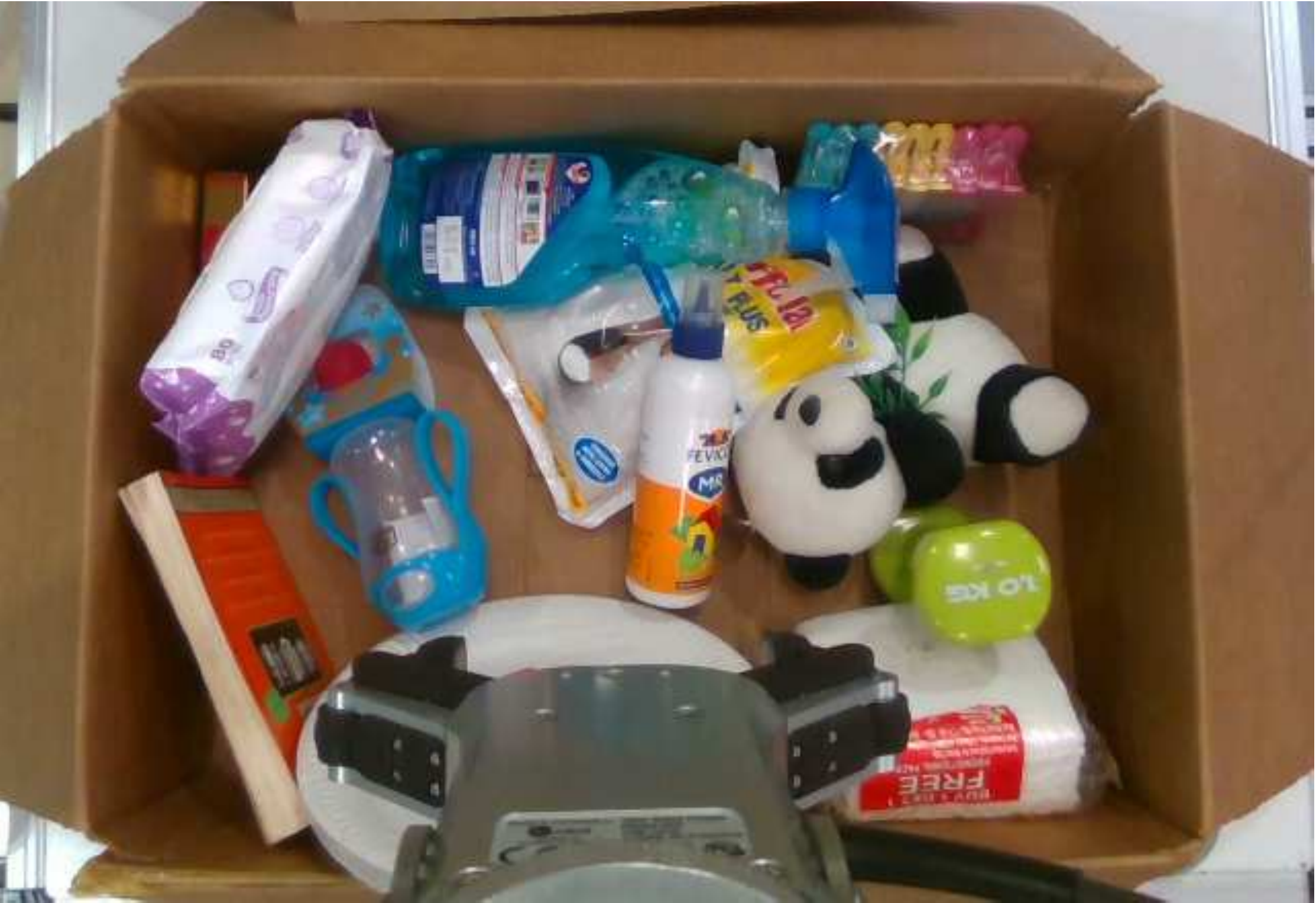} 
\includegraphics[width=0.3\linewidth,height=0.2\linewidth]{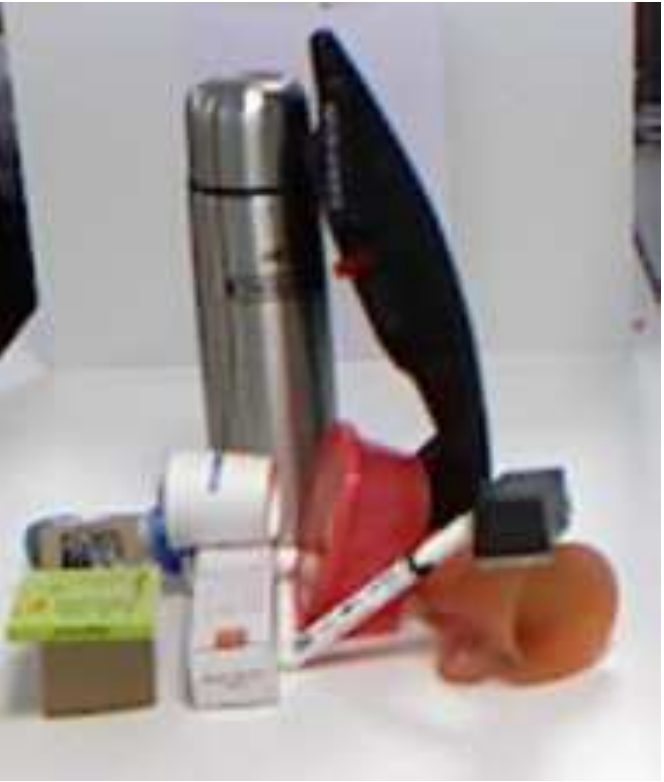} 
\includegraphics[width=0.3\linewidth,height=0.2\linewidth]{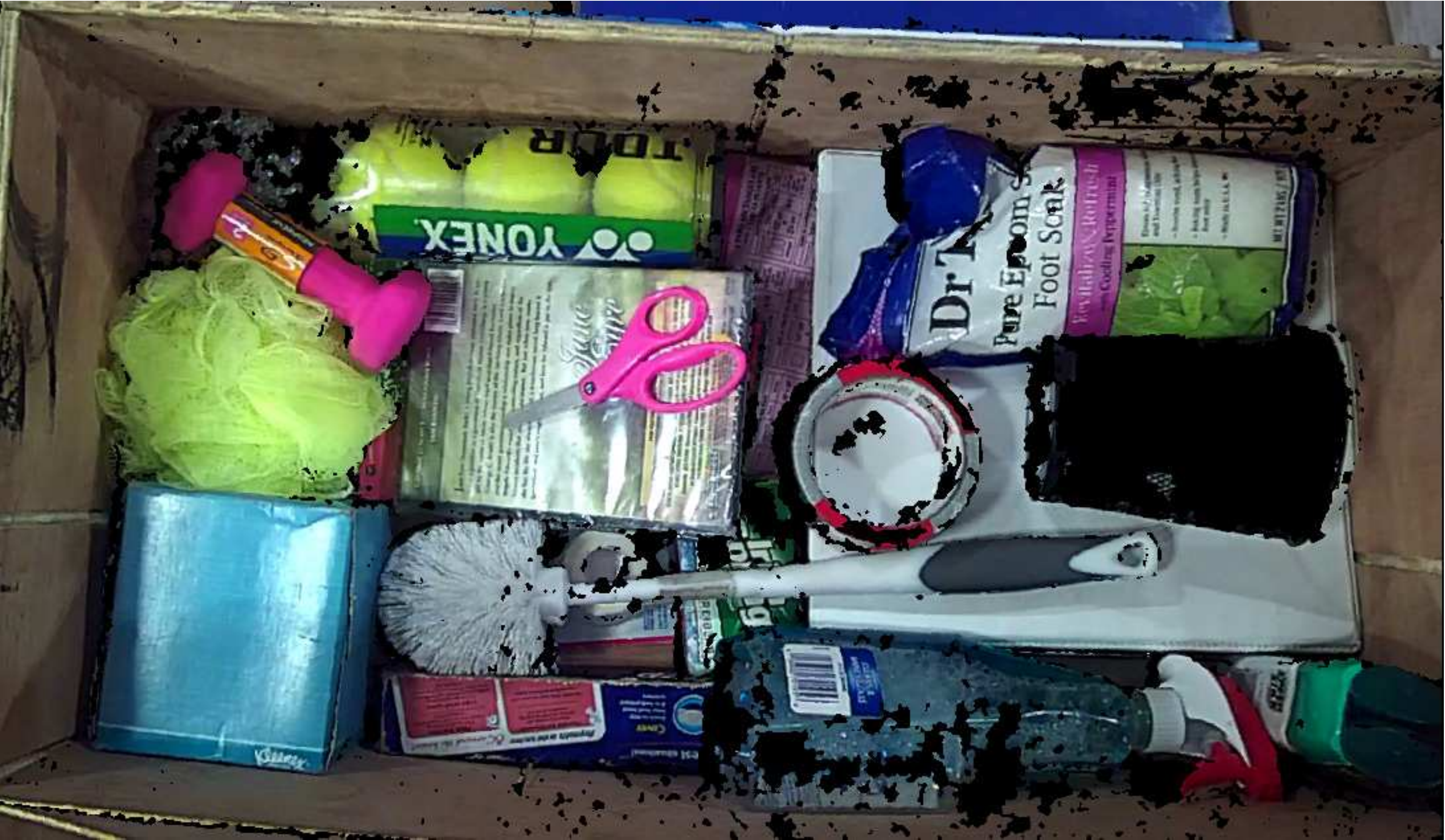} \\
\includegraphics[width=0.3\linewidth,height=0.2\linewidth]{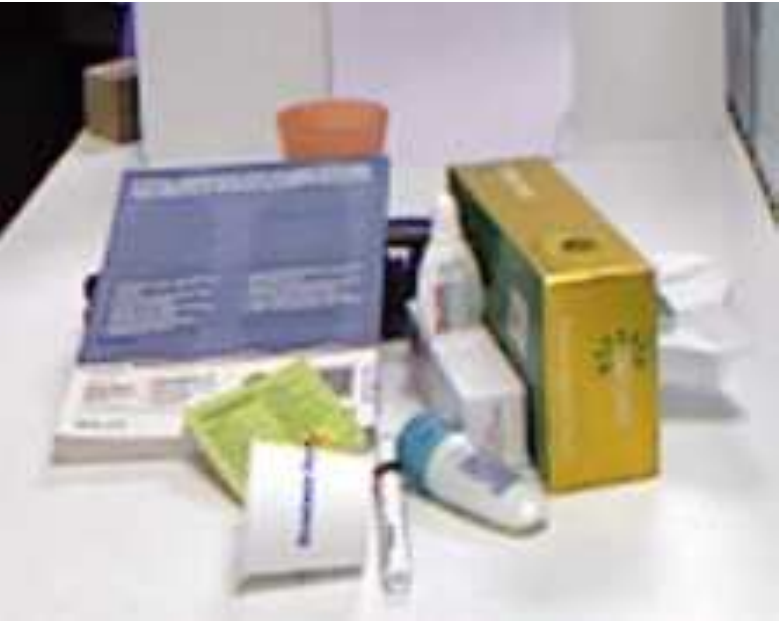} 
\includegraphics[width=0.3\linewidth,height=0.2\linewidth]{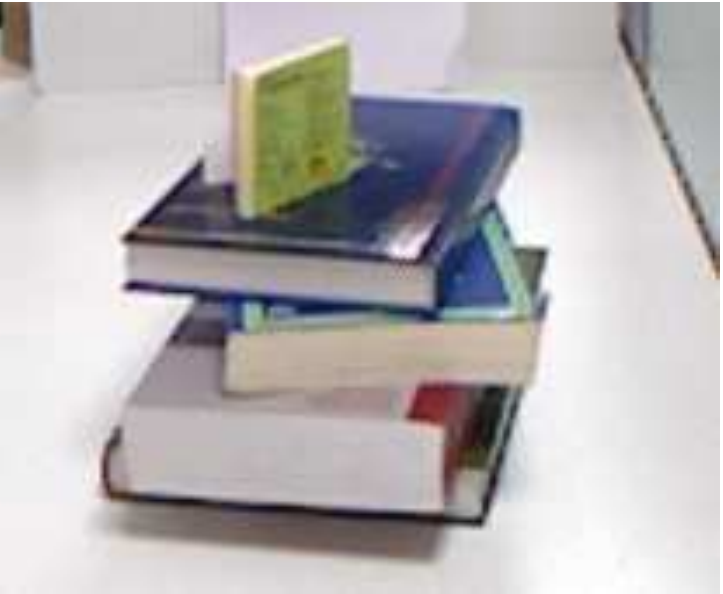} 
\includegraphics[width=0.3\linewidth,height=0.2\linewidth]{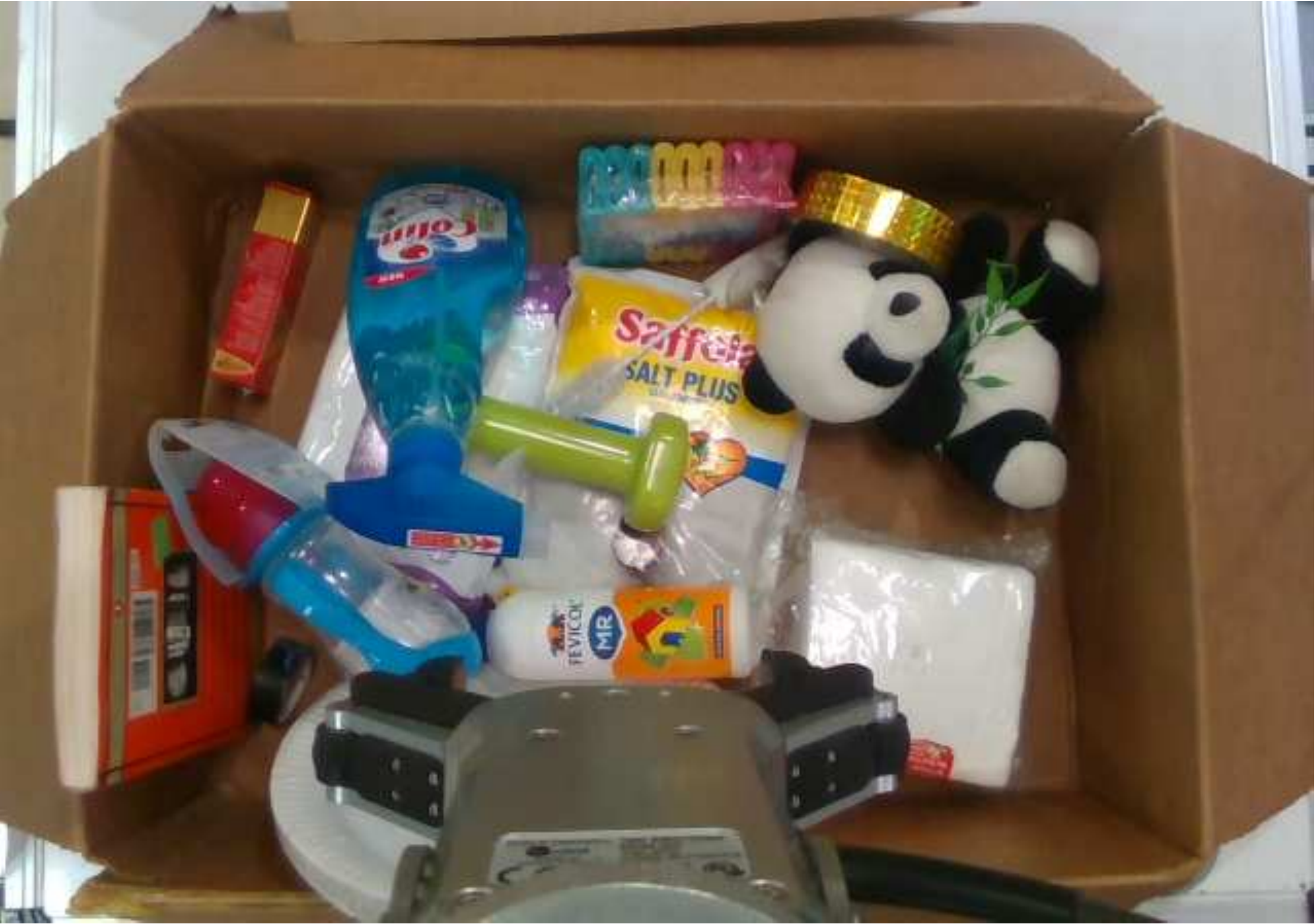}\\
\scriptsize{(b) Snapshot of TCS Grasp Dataset 2}
\caption{Snapshot of frames in TCS Grasping datasets 1 and 2. Each dataset consists of
images and point cloud data files along with annotations in text files.}
\label{fig:snapshot}
\end{figure}

In this section, we provide results of various experiments performed
to establish the usefulness of the proposed algorithm in comparison to
the existing state-of-the-art methods. As explained before, our focus
is to find suitable graspable affordances for various household items.
The input to our algorithm is a 3D point cloud obtained from an RGBD
or a range sensor and, the output is a set of graspable affordances
comprising of graspable regions and gripper pose required to pick the
objects. We have particularly tested our algorithms on datasets
obtained using Kinect \cite{zhang2012microsoft}, realsense
\cite{draelos2015intel}  and Ensenso \cite{ensenso} depth sensors.  An
additional smoothing pre-processing step is applied to the Ensenso
point cloud which are otherwise quite noisy compared to that obtained
using either Kinect or realsense sensors. As we will demonstrate
shortly, we have considered grasping of individual objects in an
extremely cluttered environment.  The performance of the proposed
algorithm is compared with other methods on four different datasets,
namely, (1) Big bird dataset \cite{singh2014bigbird}, (2) Cornell
Grasping dataset \cite{lenz2015deepgrasp}, (3) ECCV dataset
\cite{Aldoma2012}, (4) Kinect Dataset \cite{SegIROS11} (5) Willow
garage dataset \cite{willow}, (5) the TCS Grasping Dataset-1 and (6)
TCS Grasping Dataset-2.  The last two are created by us as a part of
this work and is made available online \cite{tcsgraspdata2018} along
with the program source code for the convenience of
users. A snapshot of images for these two datasets are shown in Figure
\ref{fig:snapshot}. The first TCS dataset contains 382 frames each
having only single object in its view inside the bin of a rack where
the view could be slightly constrained due to poor illumination.
Similarly, the second dataset consists of 40 frames with multiple
objects in extreme clutter environment. Each dataset contains RGB
images, point cloud data (as .pcd files) and annotations in the text
format. These datasets exhibit more difficult real world scenarios
compared to what is available in the existing datasets. The algorithm
is implemented on a Linux laptop with a i7 processor and 16 GB RAM.

\subsection{Performance Measure}
Different authors use different parameters to evaluate the performance
of their algorithm. For instance, authors in \cite{plattgrasppose2016}
use \emph{recall at high precision} as a measure while few others as
in \cite{lenz2015deepgrasp} use \emph{accuracy} as a measure. In some
cases, accuracy may not be a good measure for grasping algorithms
because the number of true negatives in a grasping dataset is usually
much more than the number of true positives. So, the accuracy could be
high even when the number of true positives (actual handles detected)
are less (or the precision is less). There are other researchers as in \cite{pas2015using}
\cite{jain2016grasp} \cite{vezzani2017grasping} who use \emph{success
rate} as a performance measure which is defined as the number of times
a robot is able to successfully pick an object in a physical
experiment. The success rate is usually directly linked to the
precision of the algorithm as the false detections or mistakes could
be detrimental to the robot operation. In other words, a grasping
algorithm with high precision is expected to yield high success rate.
The precision is usually defined as the fraction of total number of
handles detected which are true. However, in a cluttered scenario, the
precision may  not always provide an effective measure to evaluate the
performance of the grasping algorithm.  For instance, it is possible to
detect multiple handles for some objects and no handles at all for
some others, without affecting the total precision score. In other
words, the fact that no handles are detected for a set of objects may
not have any effect on the final score as long as there are other
objects for which more than one handle is detected. 

In our case, the precision is considered to be 100\% as any handle
that does not satisfy the gripper and the environment constraints is
rejected. In order to address the concerns mentioned above, we use
\emph{recall at high precision} as a measure of the performance of our
algorithm which is defined as the fraction of total number of
graspable objects for which \emph{at least one valid handle} is detected.
Mathematically, it can be written as 
\begin{equation} {\scriptsize
    \textrm{recall \%}} =
    \frac{\parbox[t]{6cm}{\scriptsize\textrm{Number of objects for
    which at least one handle is
  detected}}}{\parbox[t]{6cm}{\scriptsize\textrm{Total number of
  graspable objects}}} \times {\scriptsize \textrm{100}}
  \label{eq:puh} \end{equation} 
The total number of graspable objects
includes objects which could be actually picked up by the robot gripper
in a real world experiment. It excludes the objects in the clutter
which can not picked up due to substantial occlusion.  This forms the
ground truth for the experiment.  Note that the above definition is
slightly different from the conventional definition of recall in the
sense that the later may include multiple handles for a given object
which are not considered in our definition.  We analyze and compare
the performance of our algorithm with an existing state-of-the-art
algorithm using this new metric as described in the next section.

\begin{figure}[!t]
  \centering
  \begin{tabular}{cc}
\includegraphics[width=0.43\linewidth,height=0.35\linewidth]{pic/handle/single/toothepaste_org.eps} &
\includegraphics[width=0.43\linewidth,height=0.35\linewidth]{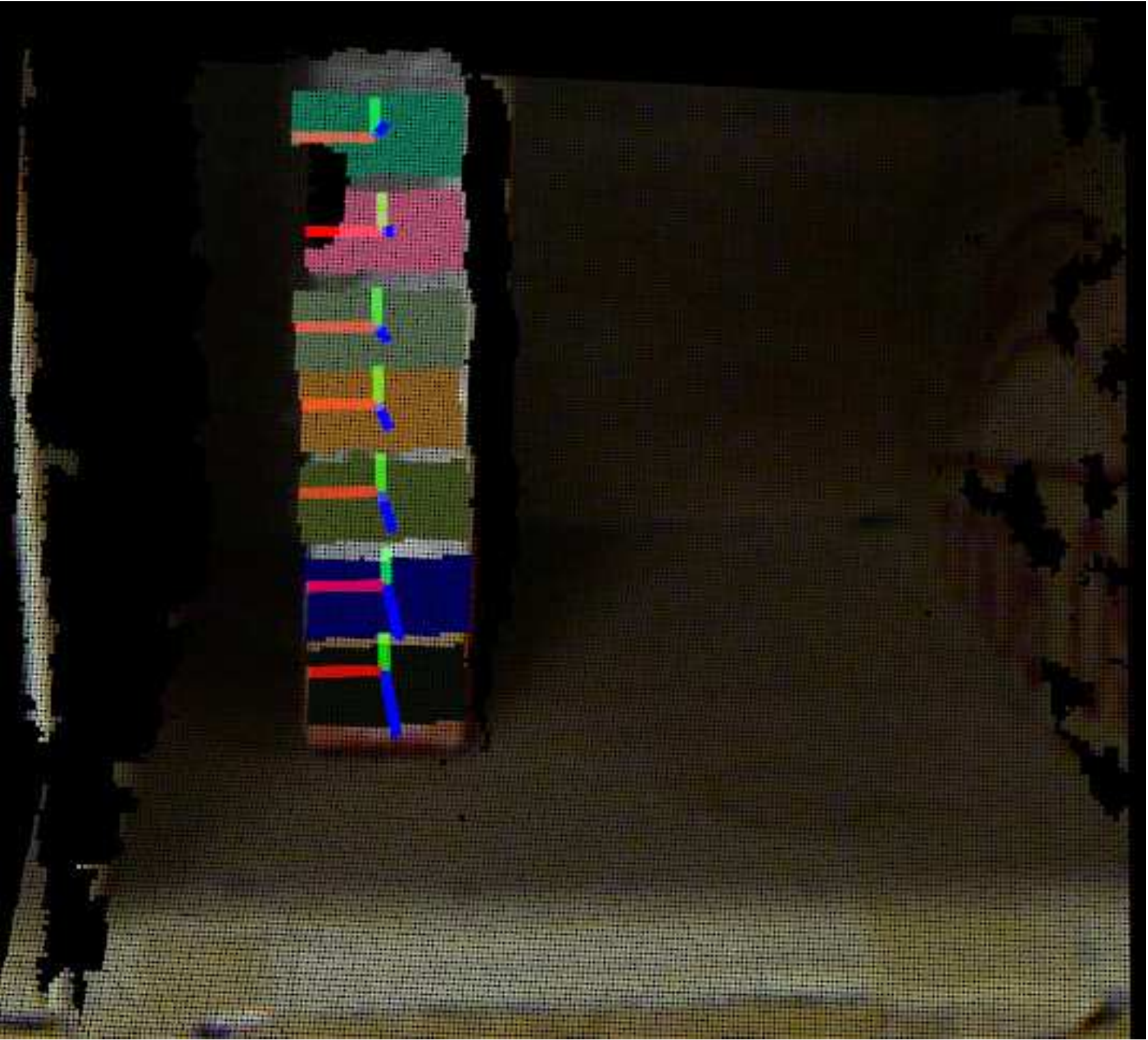} \\
\scriptsize{(a) Toothpaste} &     \\
\includegraphics[width=0.43\linewidth,height=0.35\linewidth]{pic/handle/single/fevicol_org.eps} &
\includegraphics[width=0.43\linewidth,height=0.35\linewidth]{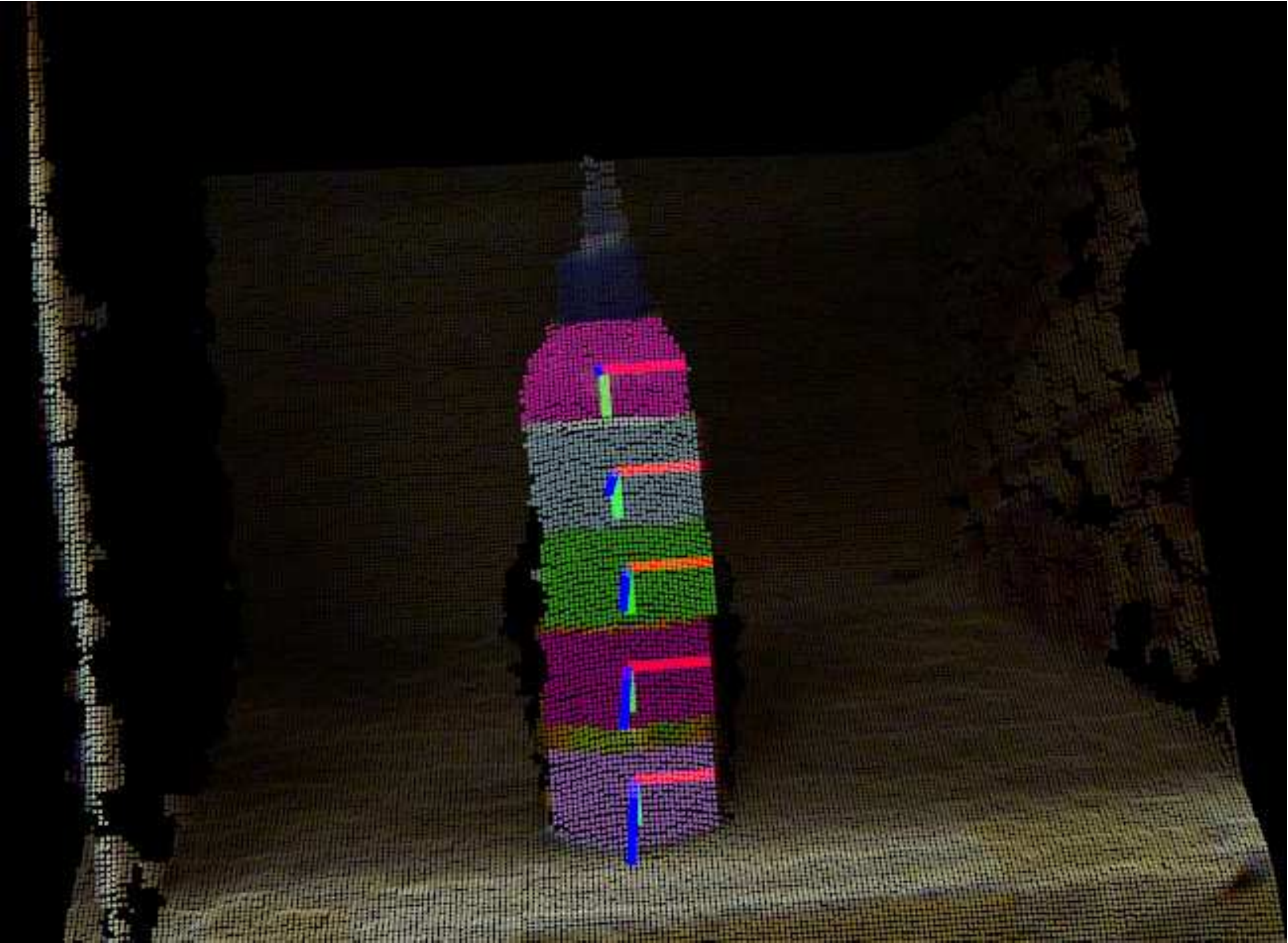} \\
\scriptsize{(b) Fevicol} &     \\
\includegraphics[width=0.43\linewidth,height=0.35\linewidth]{pic/handle/single/battery_org.eps} &
\includegraphics[width=0.43\linewidth,height=0.35\linewidth]{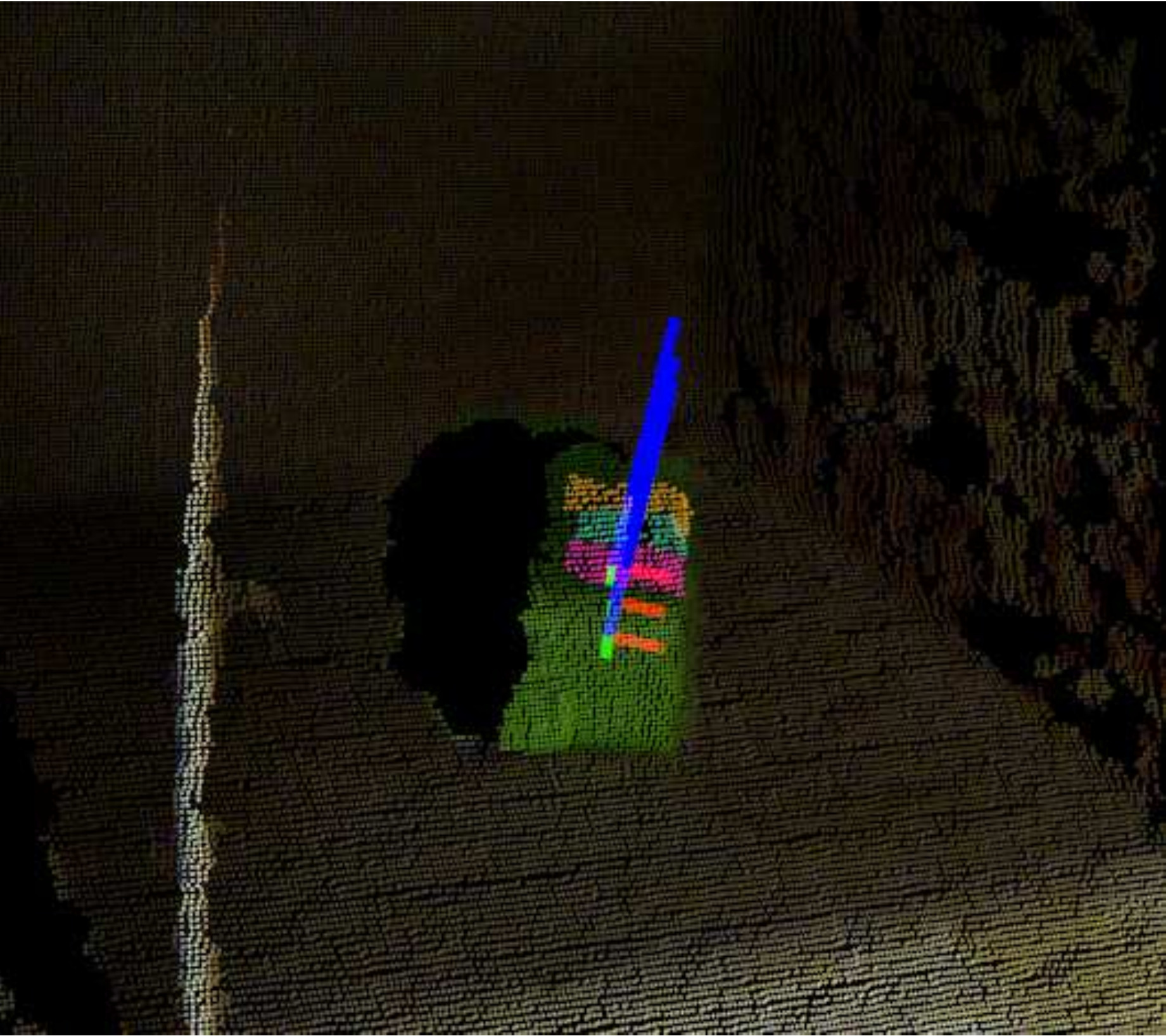} \\
\scriptsize{(c) Battery} &     \\
\includegraphics[width=0.43\linewidth,height=0.35\linewidth]{pic/handle/single/spout_brush.eps} & 
\includegraphics[width=0.43\linewidth,height=0.35\linewidth]{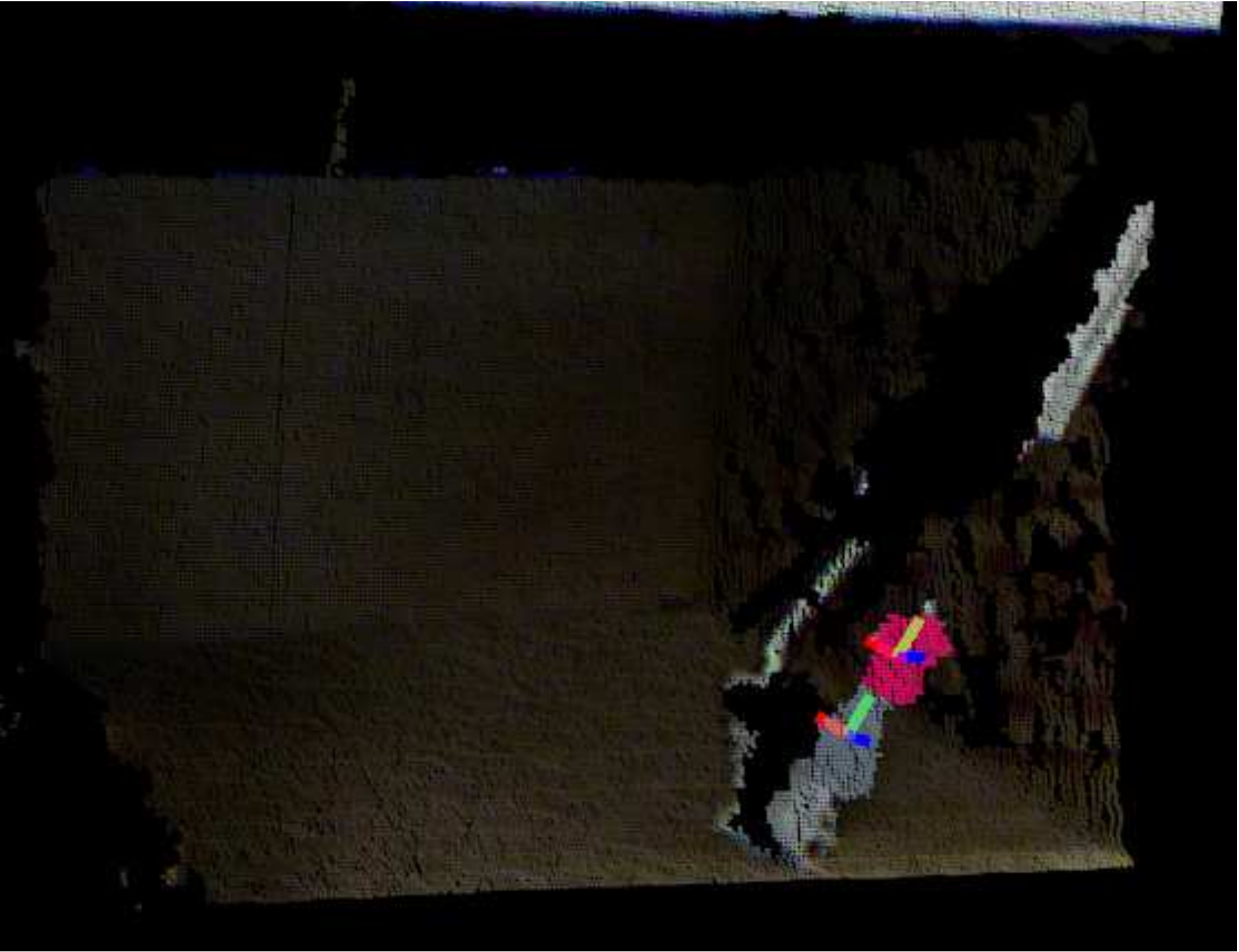} \\
\scriptsize{(d) Sprout Brush} &     \\
\includegraphics[width=0.43\linewidth,height=0.35\linewidth]{pic/handle/single/cup.eps} &
\includegraphics[width=0.43\linewidth,height=0.35\linewidth]{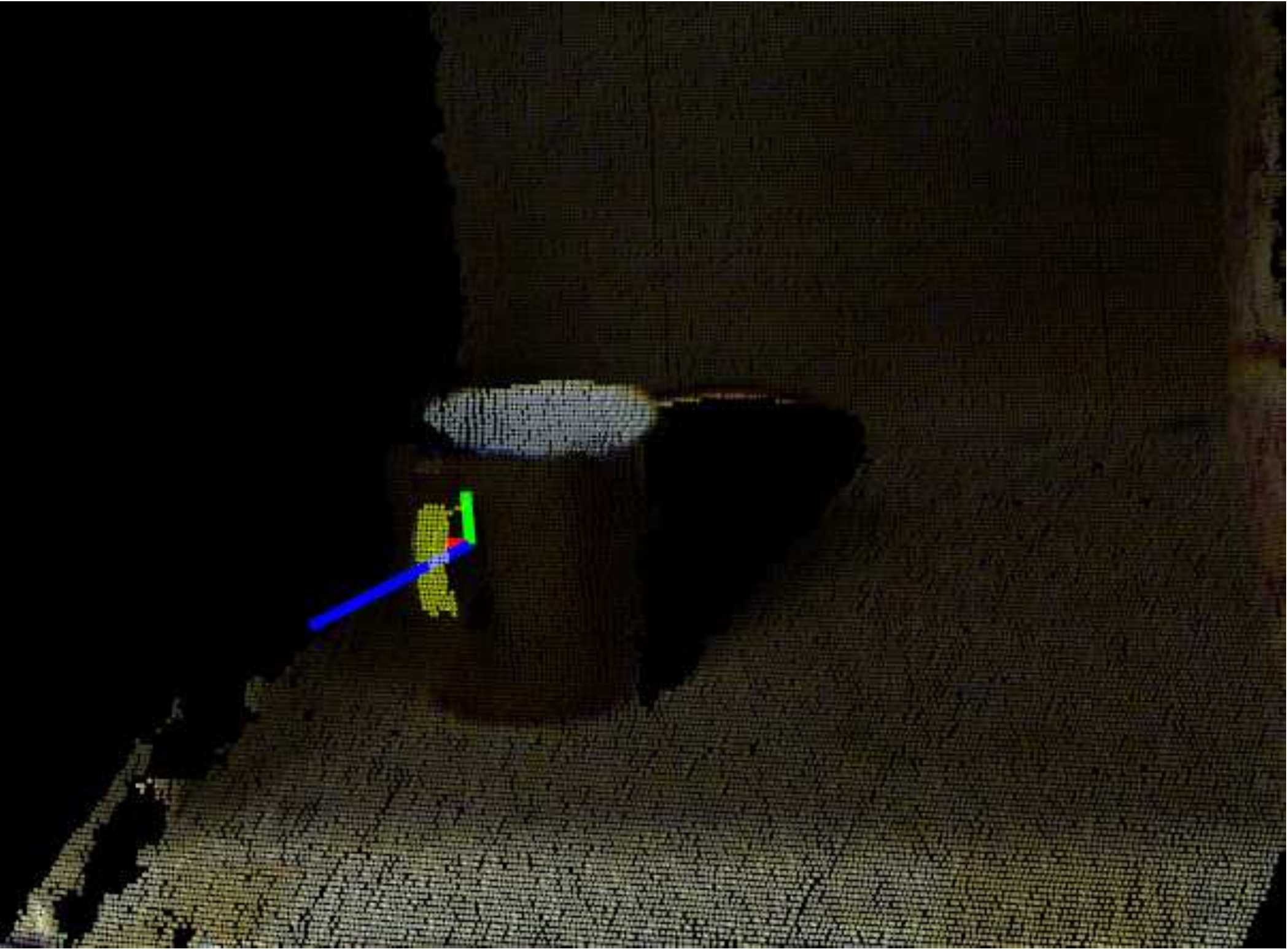} \\
\scriptsize{(e) Cup} &     \\
\includegraphics[width=0.43\linewidth,height=0.35\linewidth]{pic/handle/single/cleaning_brush.eps} &
\includegraphics[width=0.43\linewidth,height=0.35\linewidth]{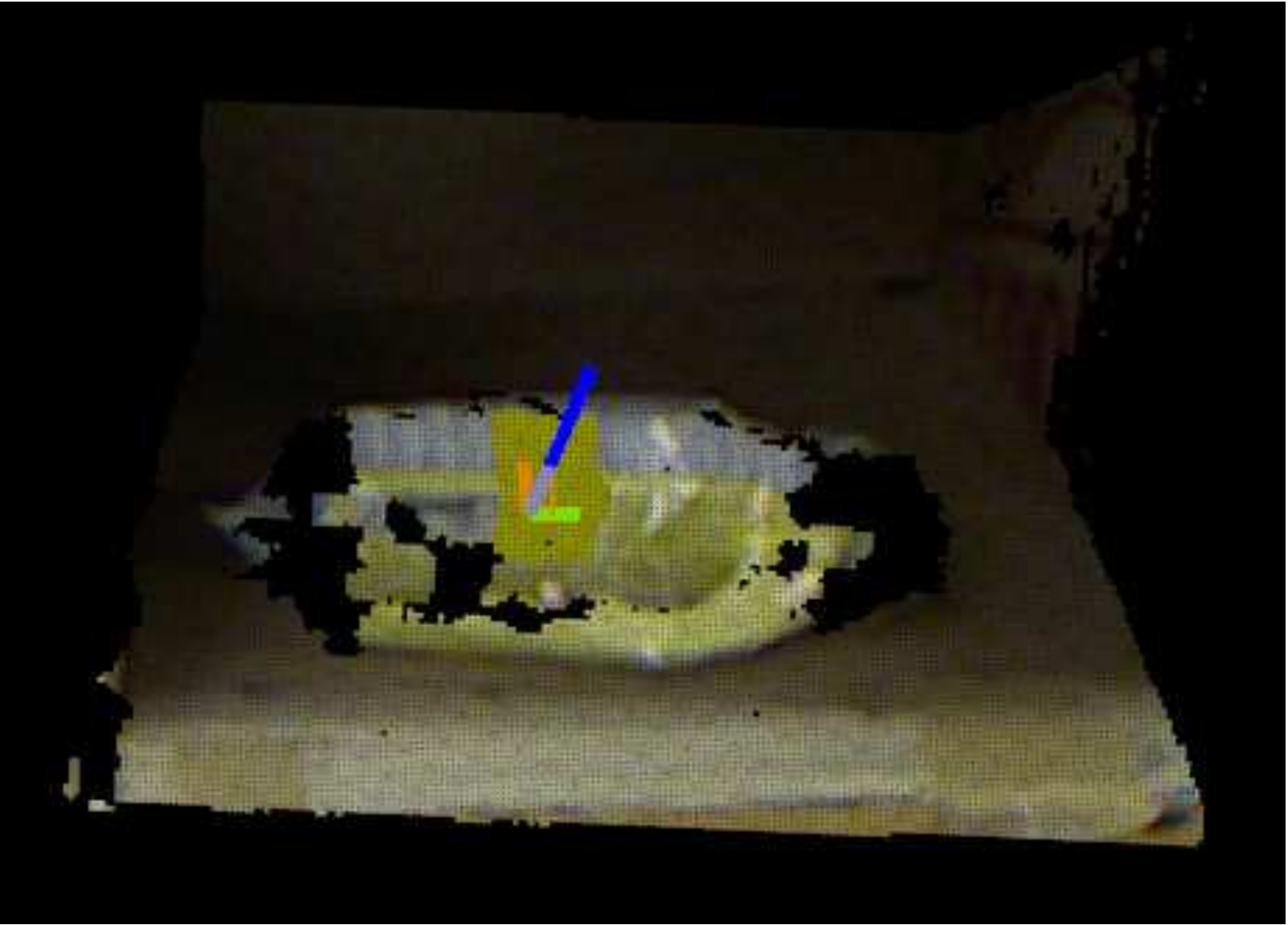} \\
\scriptsize{(f) Cleaning Brush} &     \\
\end{tabular}
\caption{ Finding graspable affordances for few objects inside a rack.
The objects in (d), (e) and (f) show few cases where it is difficult
to find graspable affordances. }
\label{fig:indrack}
\end{figure}

\subsection{Grasping of Individual Objects} 

First, we demonstrate the performance of the proposed algorithm in
picking individual objects. Table \ref{tab:indobjcomp} shows the
performance of the proposed algorithm on TCS dataset 1. This dataset
has 382 frames along with the corresponding 3D point cloud data and
annotations for ground truth. A snapshot of objects present in these
dataset is  shown in Figure \ref{fig:snapshot}. The performance of our
proposed algorithm on this dataset is compared with Platt's algorithm
reported in \cite{Pas2013LocalizingGA} \cite{ten2016localizing}.  As
one can see in Table \ref{tab:indobjcomp}, the proposed algorithm is
able to find graspable affordances for objects in more number of
frames and hence it is more robust compared to the previous approach.
On an average, our algorithm is able to detect handles in 94\% of the
frames compared to Platt's approach which can detect handles only for
51\% of frames. This could be attributed to the fact that Platt's
algorithm primarily relies on surface curvature to find handles and
hence, can not deal with rectangular objects with flat surfaces. They
try to overcome this limitation in \cite{pas2015using} by training a
SVM classifier to detect valid grasp out of a number of hypotheses
created using HoG features. Compared to this approach, our proposed
method is much simpler to implement as it does not require any
training and can be implemented in real-time. It also does not depend
on image features which are more susceptible to various photometric
effects. Some of the handles detected by our algorithm for individual
objects are shown in Figure \ref{fig:indrack}. Examples (a)-(c) shows
few instances of simple objects where it is easier to find affordances
while the (d)-(f) shows few difficult objects for which finding a
suitable handle is challenging.

\begin{table}[!t]
\centering
\caption{Performance Comparison for TCS Grasping Dataset 1 - Individual Objects}
\label{tab:indobjcomp}
\scriptsize
\begin{tabular}{|c|>{\centering\arraybackslash}m{1.3cm}|>{\centering\arraybackslash}m{1.5cm}|>{\centering\arraybackslash}m{1.3cm}|}
  \hline
  Object     & Total Number of frames & \multicolumn{2}{>{\centering\arraybackslash}m{2.5cm}|}{\% of frames where a valid handle is detected} \\ \hline
& &  Platt's Method \cite{ten2016localizing} & Proposed Method \\ \hline
Toothpaste    & 40      & 38         & 90         \\ \hline
Cup           & 50      & 70         & 96         \\ \hline
Dove Soap     & 40      & 25         & 100         \\ \hline
Fevicol       & 40      & 75         & 92        \\ \hline
Battery       & 50      & 36         & 98        \\ \hline
Clips         & 21      & 45         & 90          \\ \hline
CleaningBrush & 40      & 30         & 90         \\ \hline
SproutBrush   & 21      & 63         & 95         \\ \hline
Devi Coffee   & 40      & 76         & 93         \\ \hline
Tissue Paper  & 40      & 40         & 96         \\ \hline
Total         & 382     & 51           & \textbf{94}     \\ \hline
\end{tabular}
\end{table}

\subsection{Grasping Objects in a Clutter}

\begin{figure}[!t]
  \centering
  \begin{tabular}{cc}
\includegraphics[width=0.45\linewidth,height=0.4\linewidth]{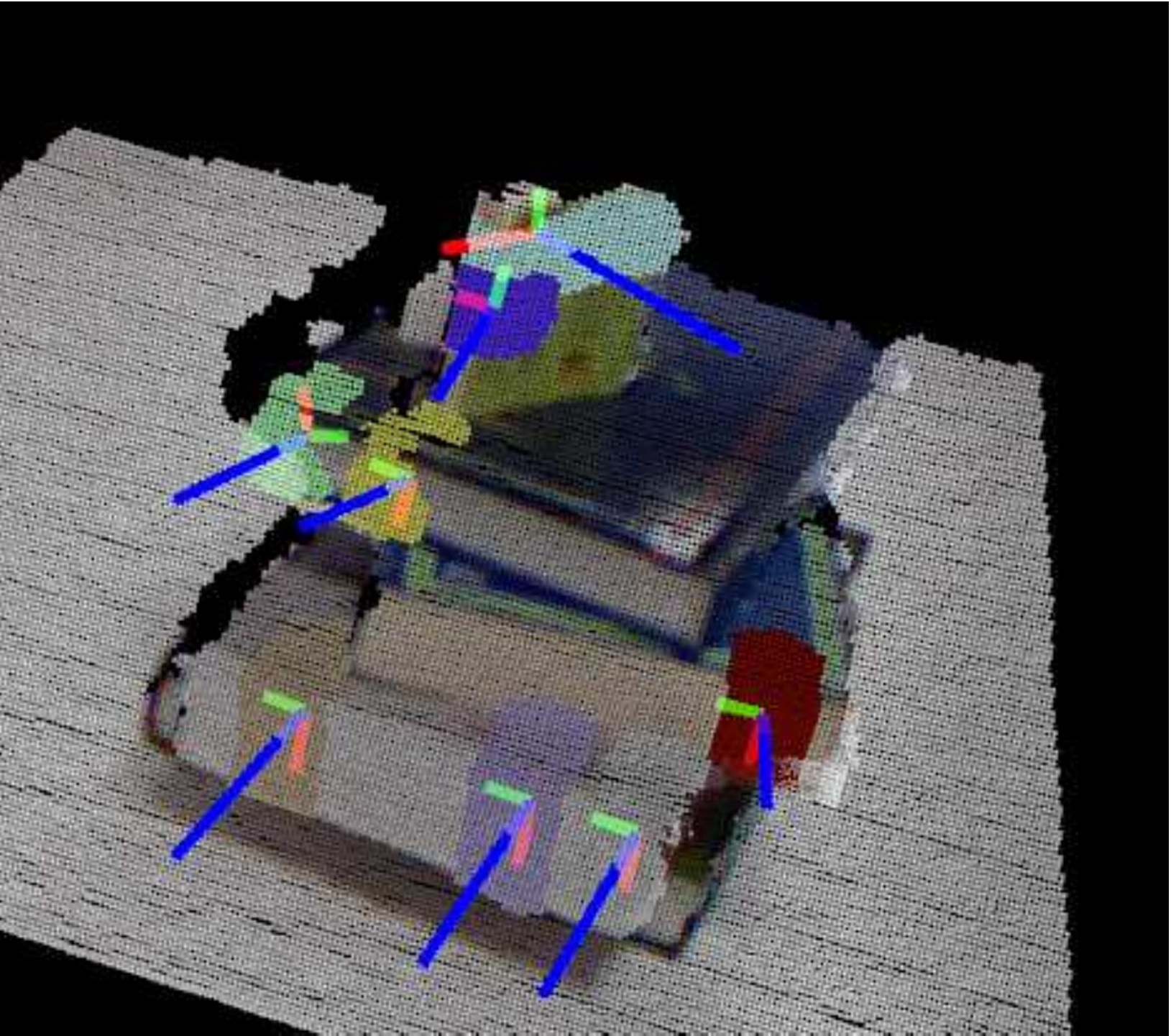} & 
\includegraphics[width=0.45\linewidth,height=0.4\linewidth]{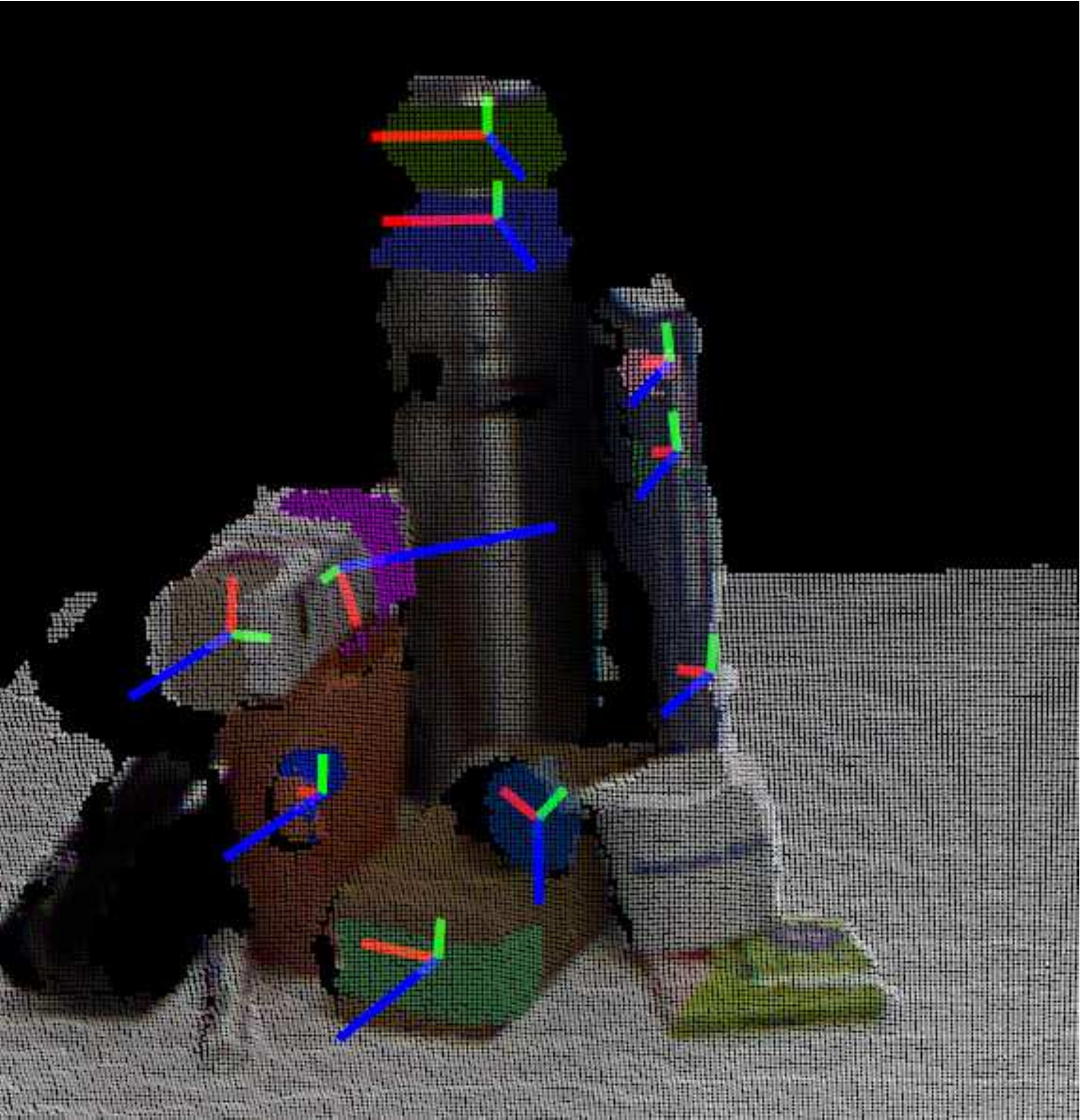}  \\
\scriptsize{(a)} & \scriptsize{(b)} \\
\includegraphics[width=0.45\linewidth,height=0.4\linewidth]{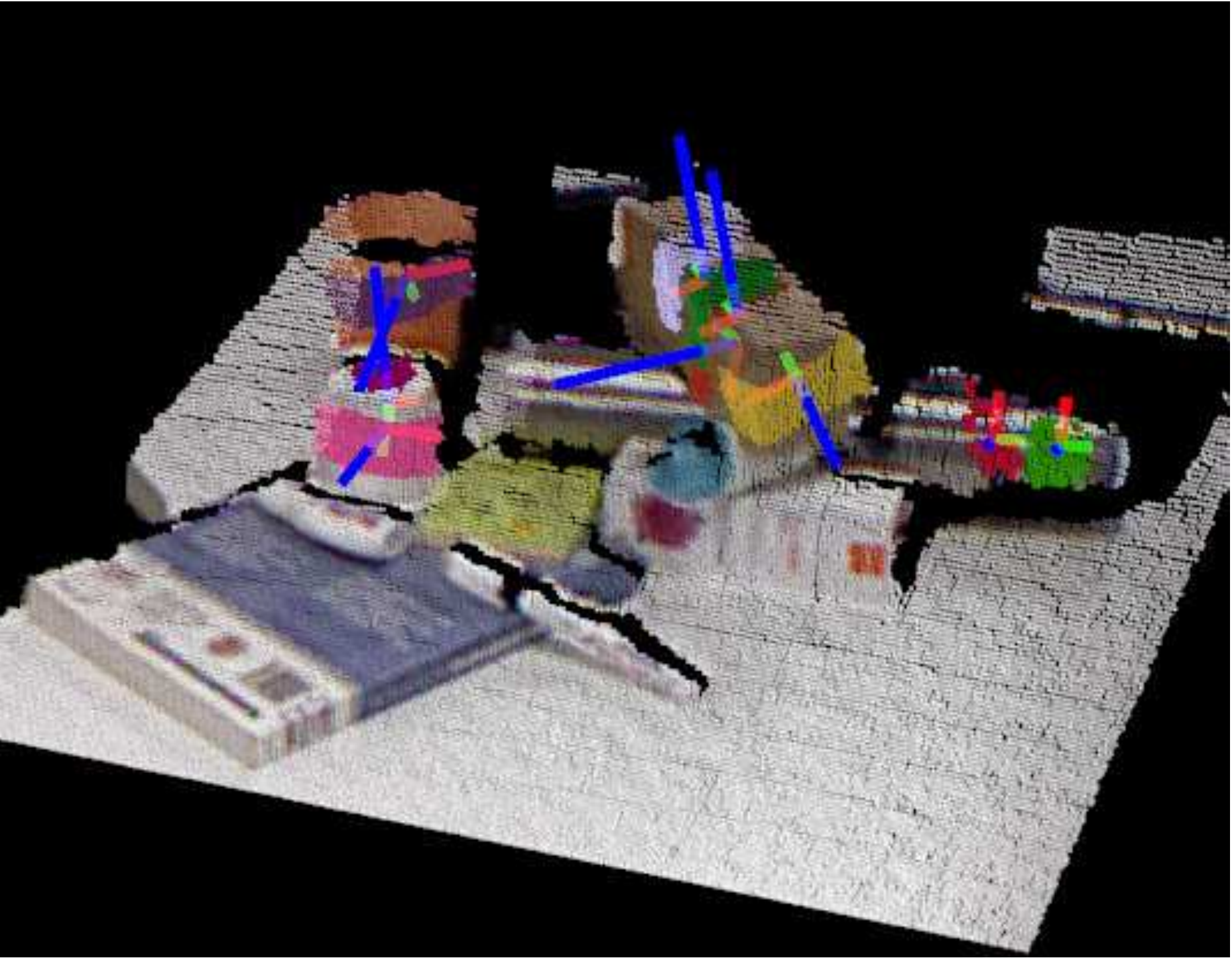} & 
\includegraphics[width=0.45\linewidth,height=0.4\linewidth]{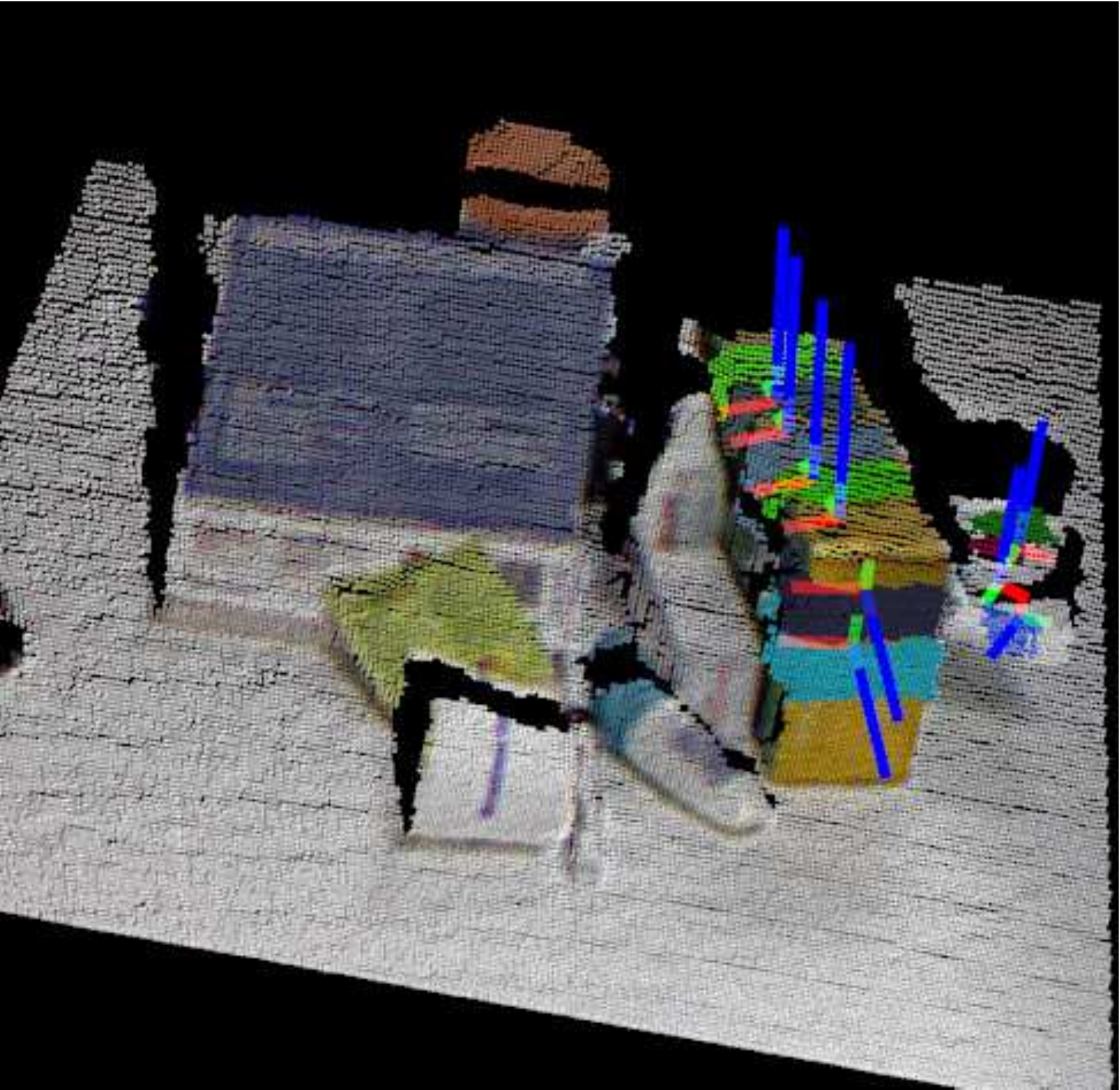} \\
\scriptsize{(c)} & \scriptsize{(d)} \\
\includegraphics[width=0.45\linewidth,height=0.4\linewidth]{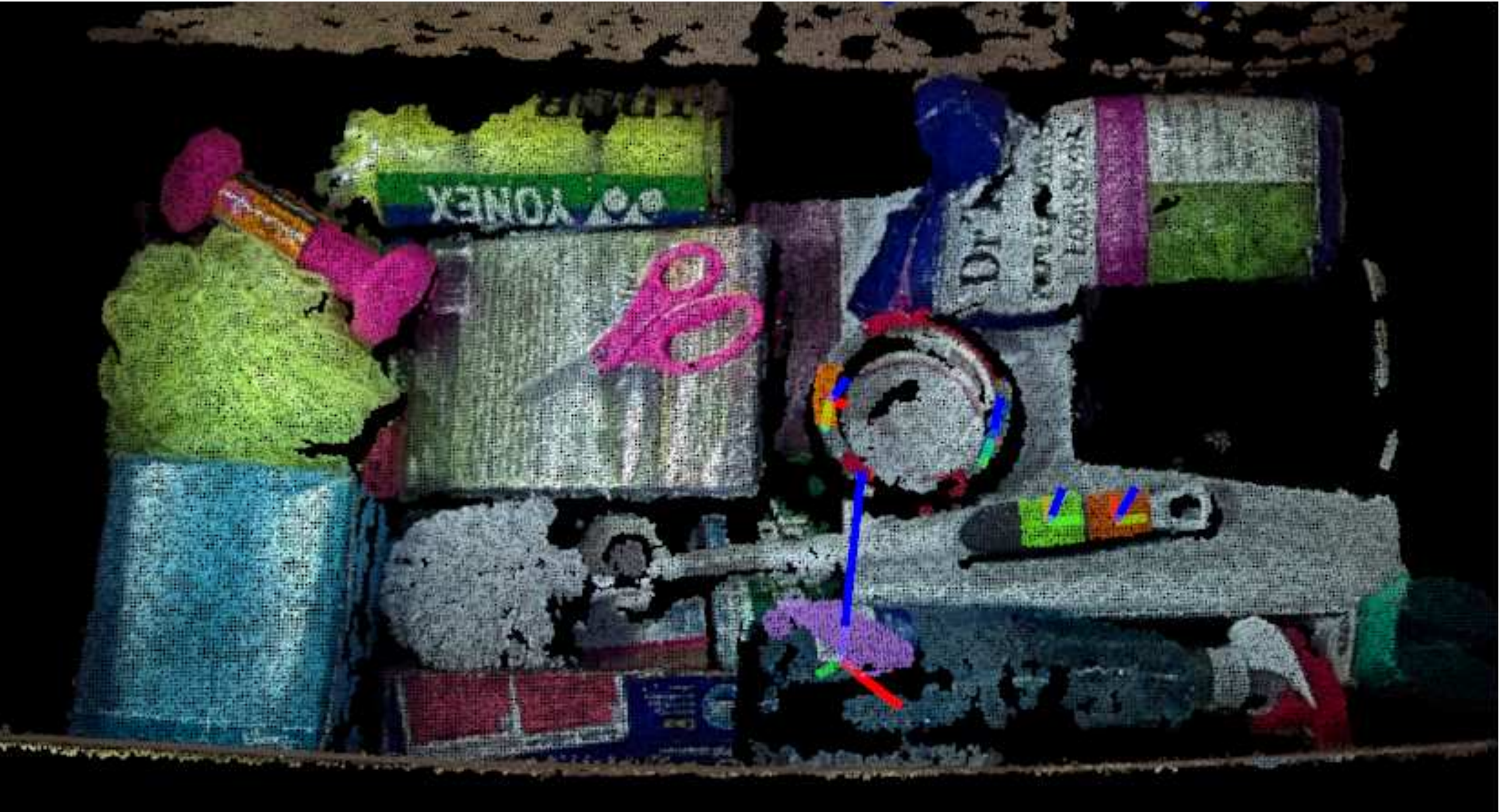} & 
\includegraphics[width=0.45\linewidth,height=0.4\linewidth]{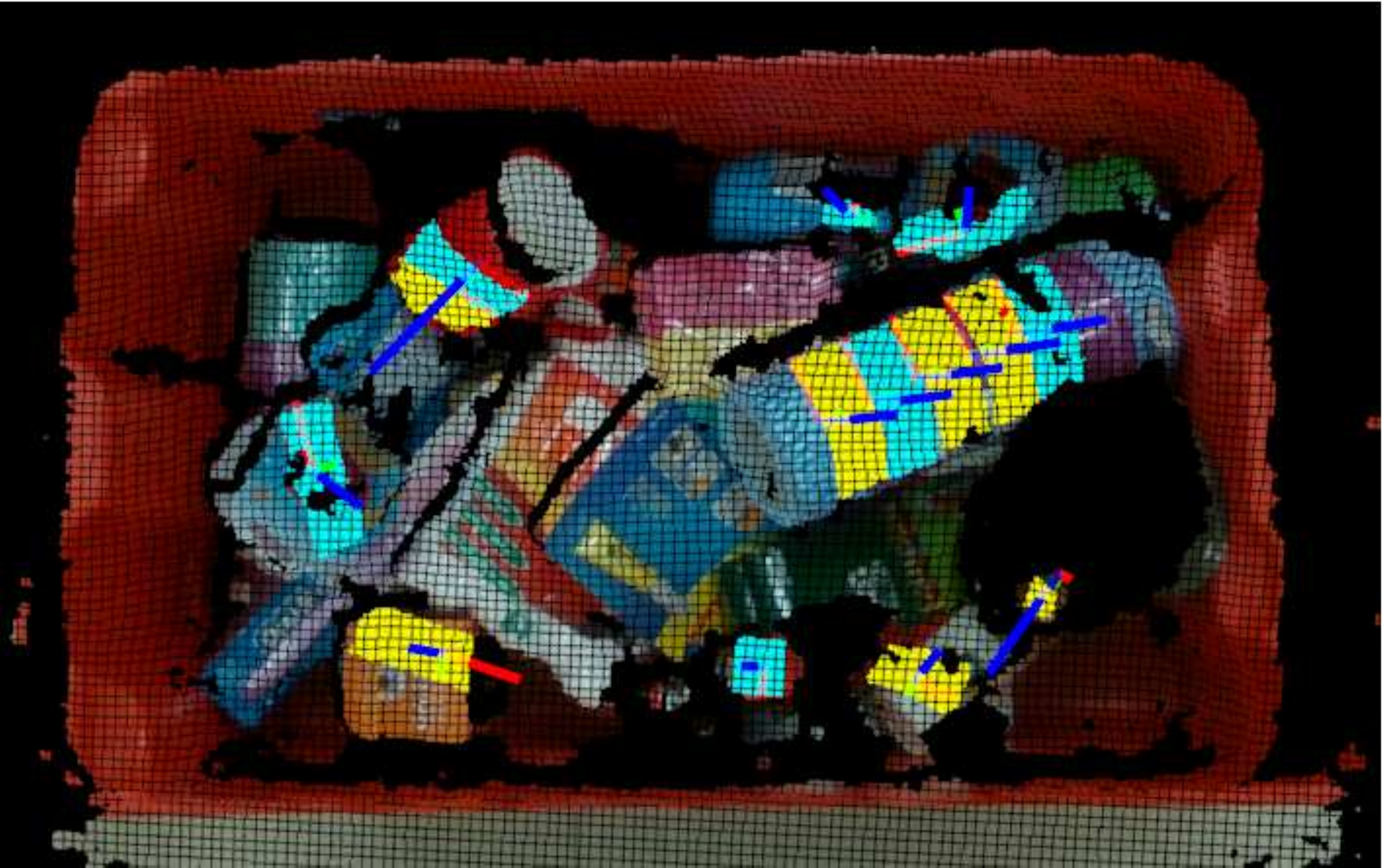} \\
\scriptsize{(e)} & \scriptsize{(f)} 
\end{tabular}
\caption{Finding graspable affordances in extreme clutter. The
proposed algorithm is capable of finding graspable affordances for
rectangular objects as well as objects with curved surface. The
maximum hand aperture ($d$) considered here is 8 cm. }
\label{fig:clutter}
\end{figure}

\begin{figure}[!t]
  \centering
  \begin{tabular}{cc}
\includegraphics[width=0.45\linewidth,height=0.4\linewidth]{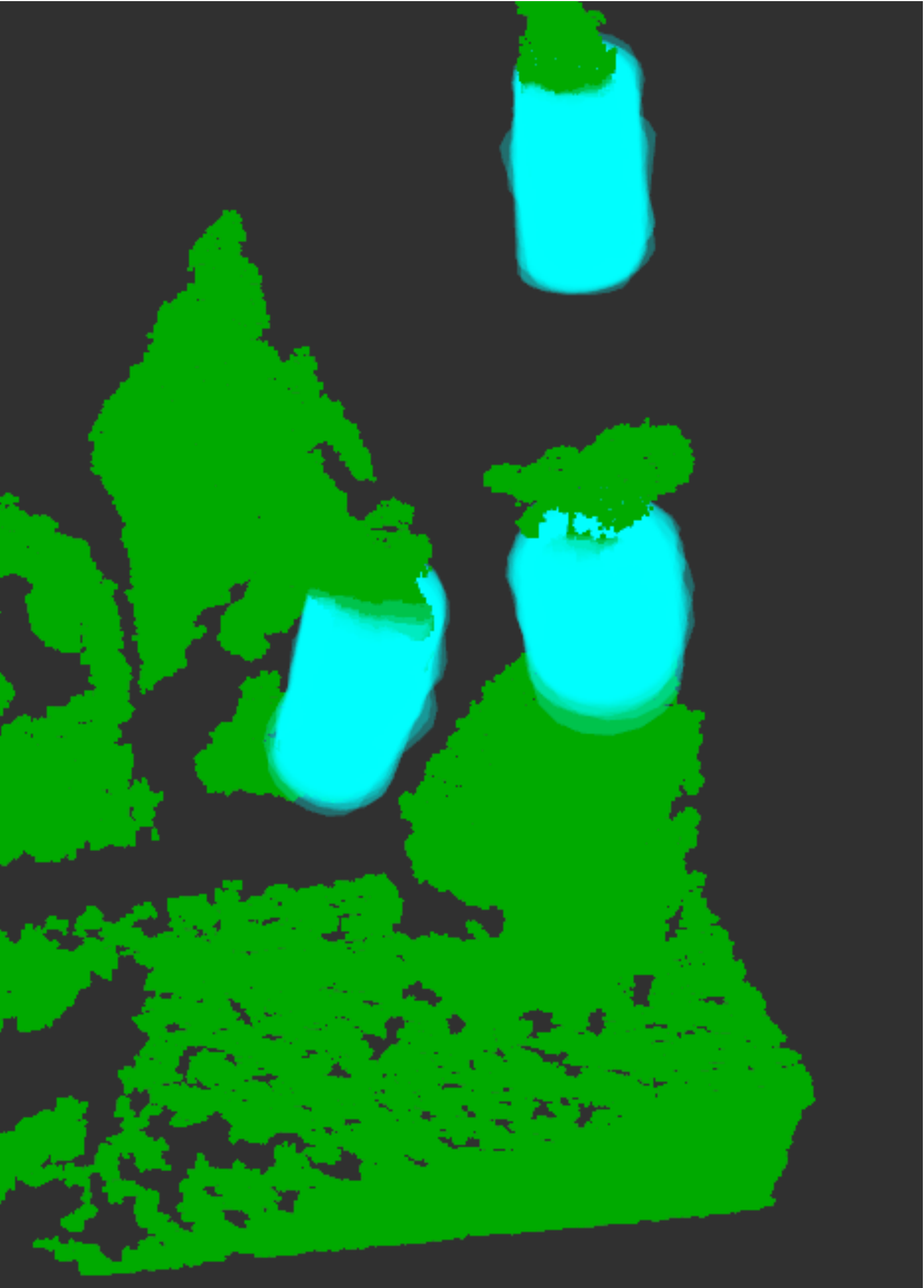} & 
\includegraphics[width=0.45\linewidth,height=0.4\linewidth]{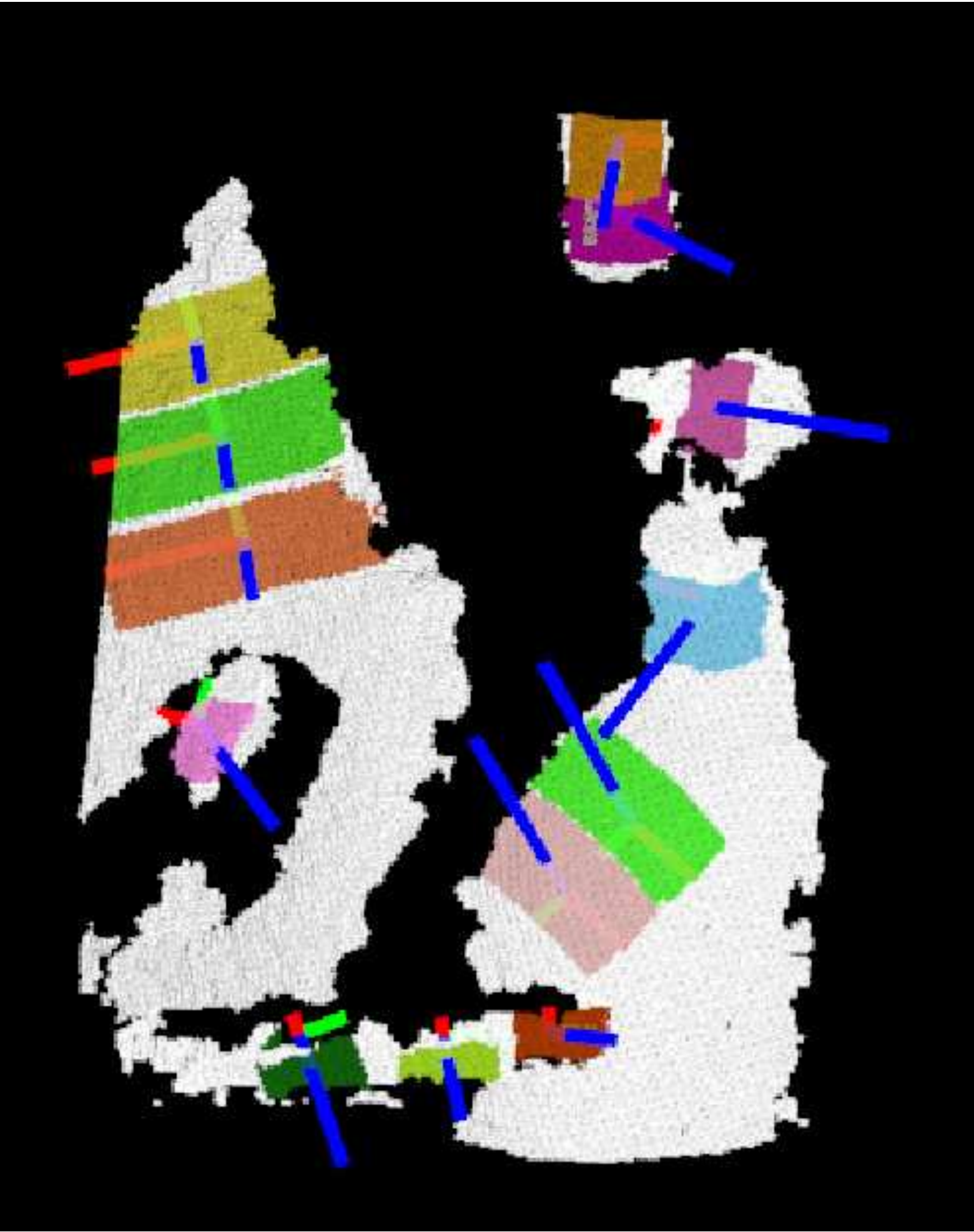} \\
\scriptsize{(a)} & \scriptsize{(b)} \\
\includegraphics[width=0.45\linewidth,height=0.4\linewidth]{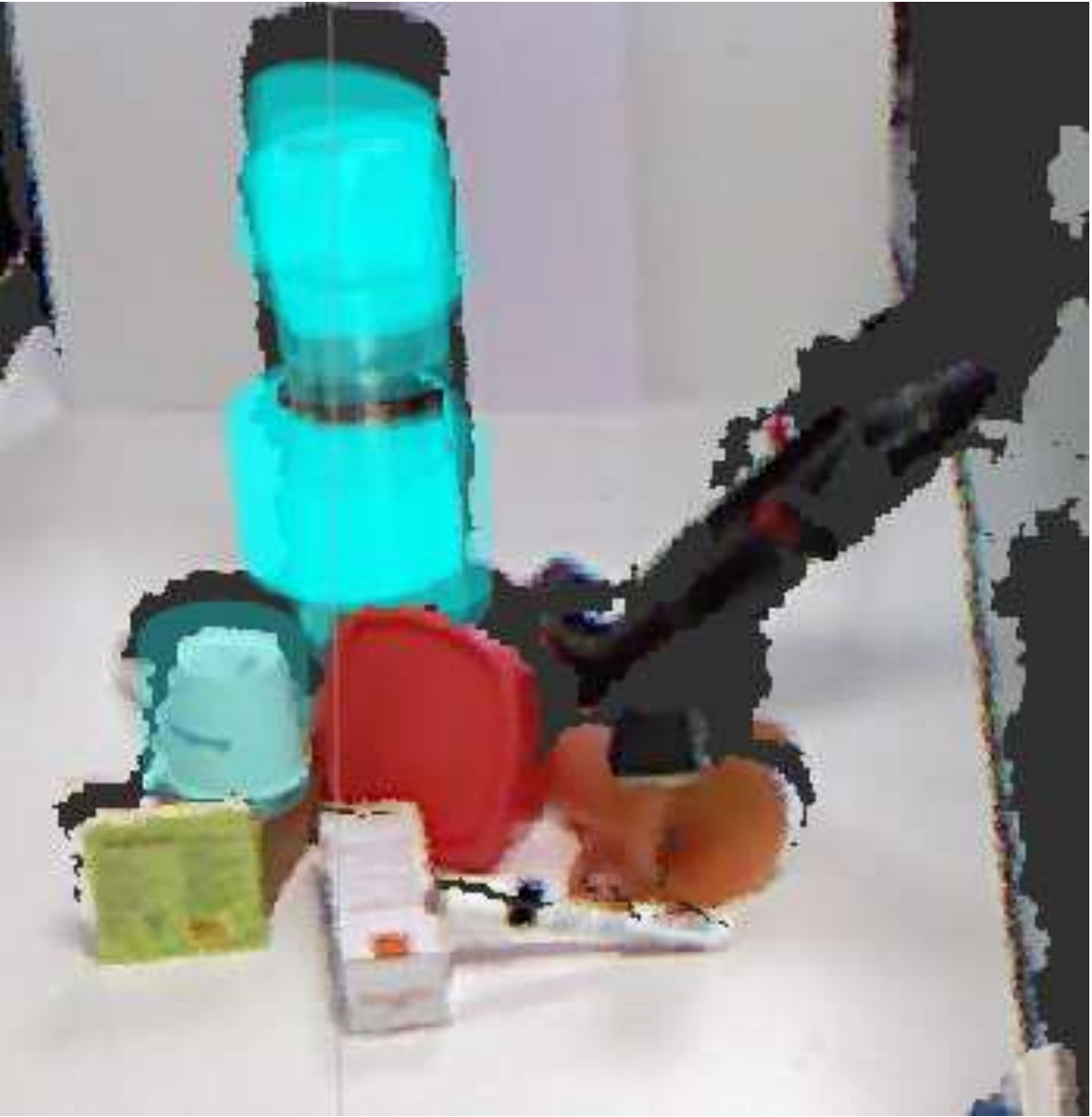} &
\includegraphics[width=0.45\linewidth,height=0.4\linewidth]{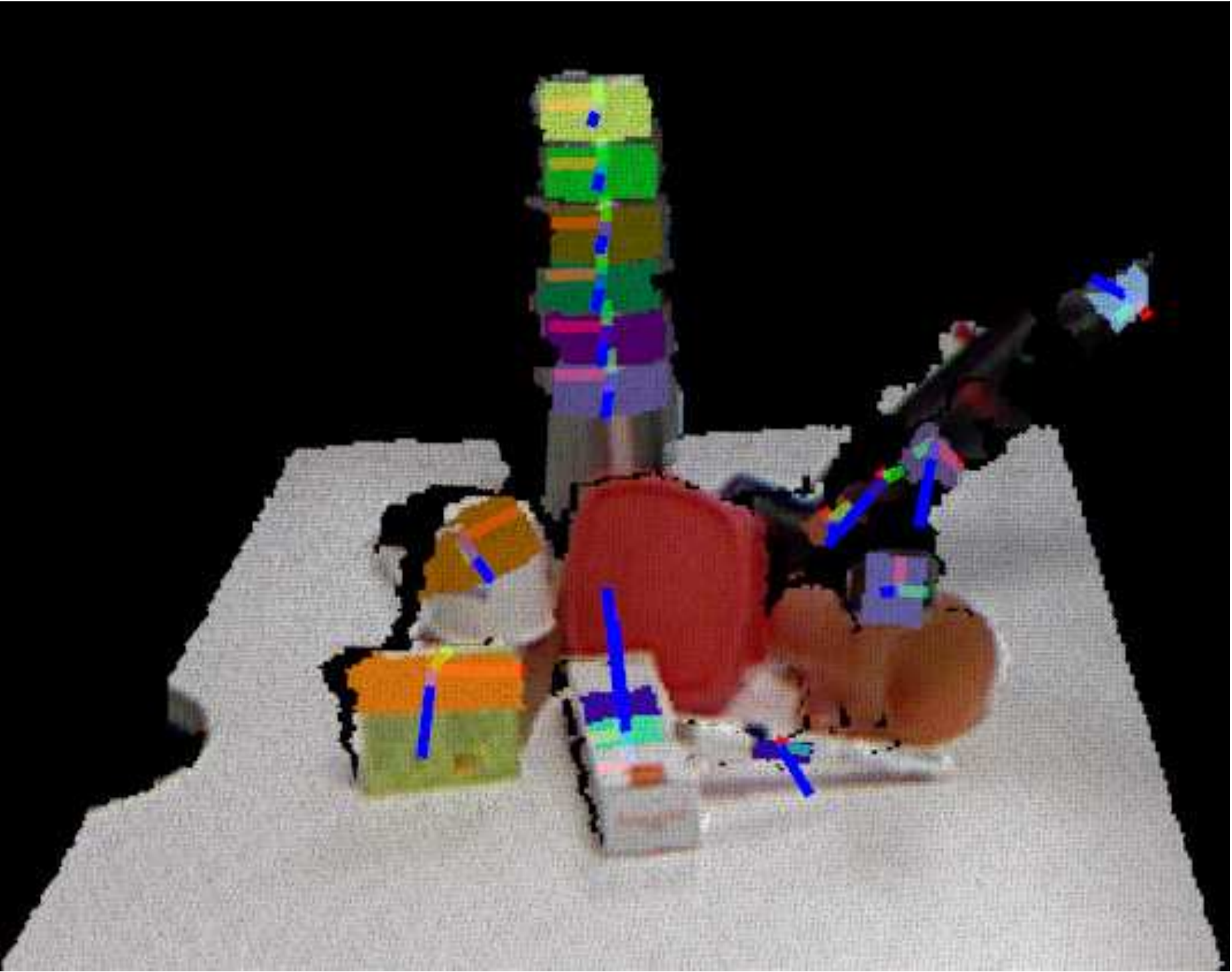} \\
\scriptsize{(c)} & \scriptsize{(d)} \\
\includegraphics[width=0.45\linewidth,height=0.4\linewidth]{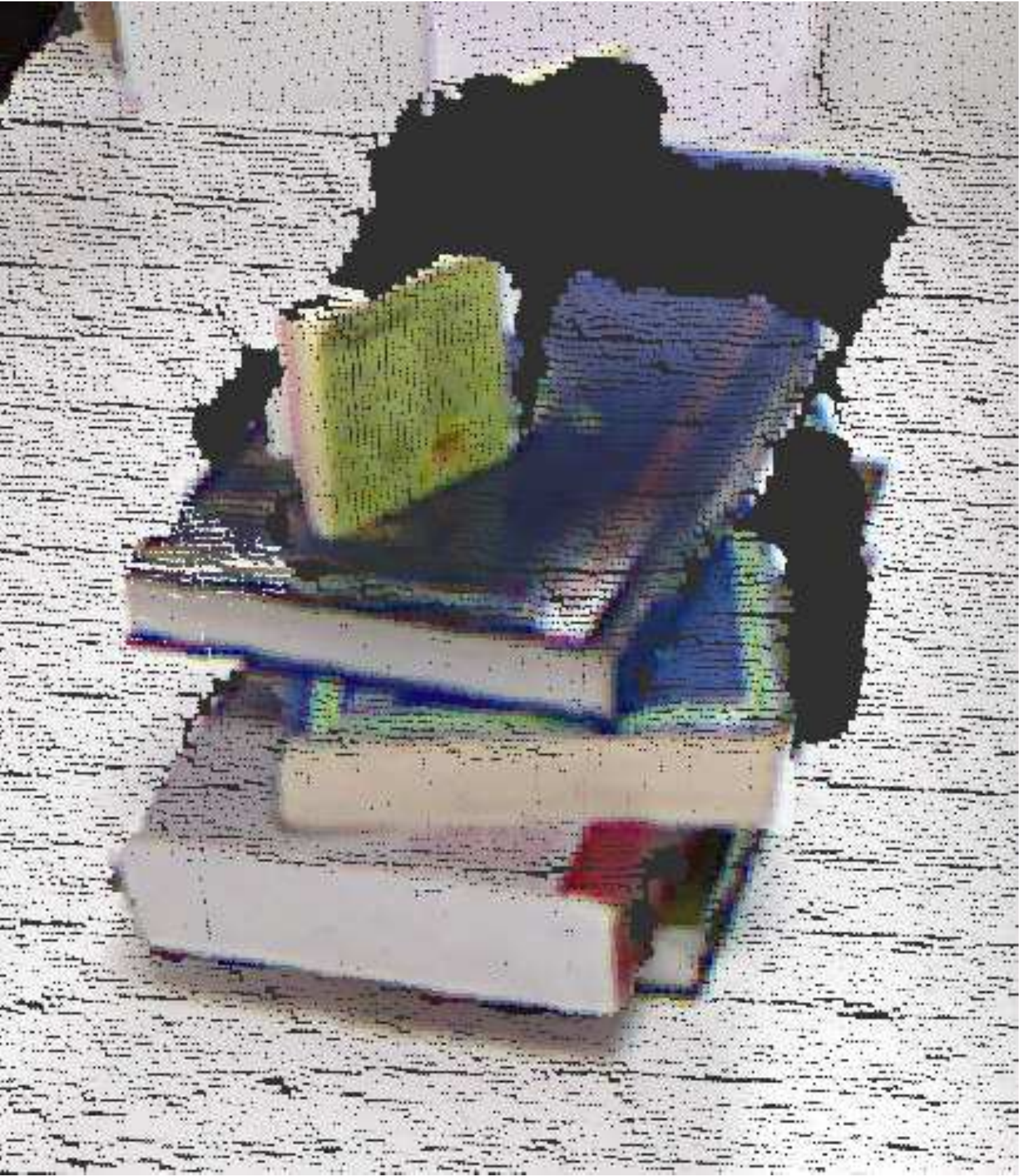} & 
\includegraphics[width=0.45\linewidth,height=0.4\linewidth]{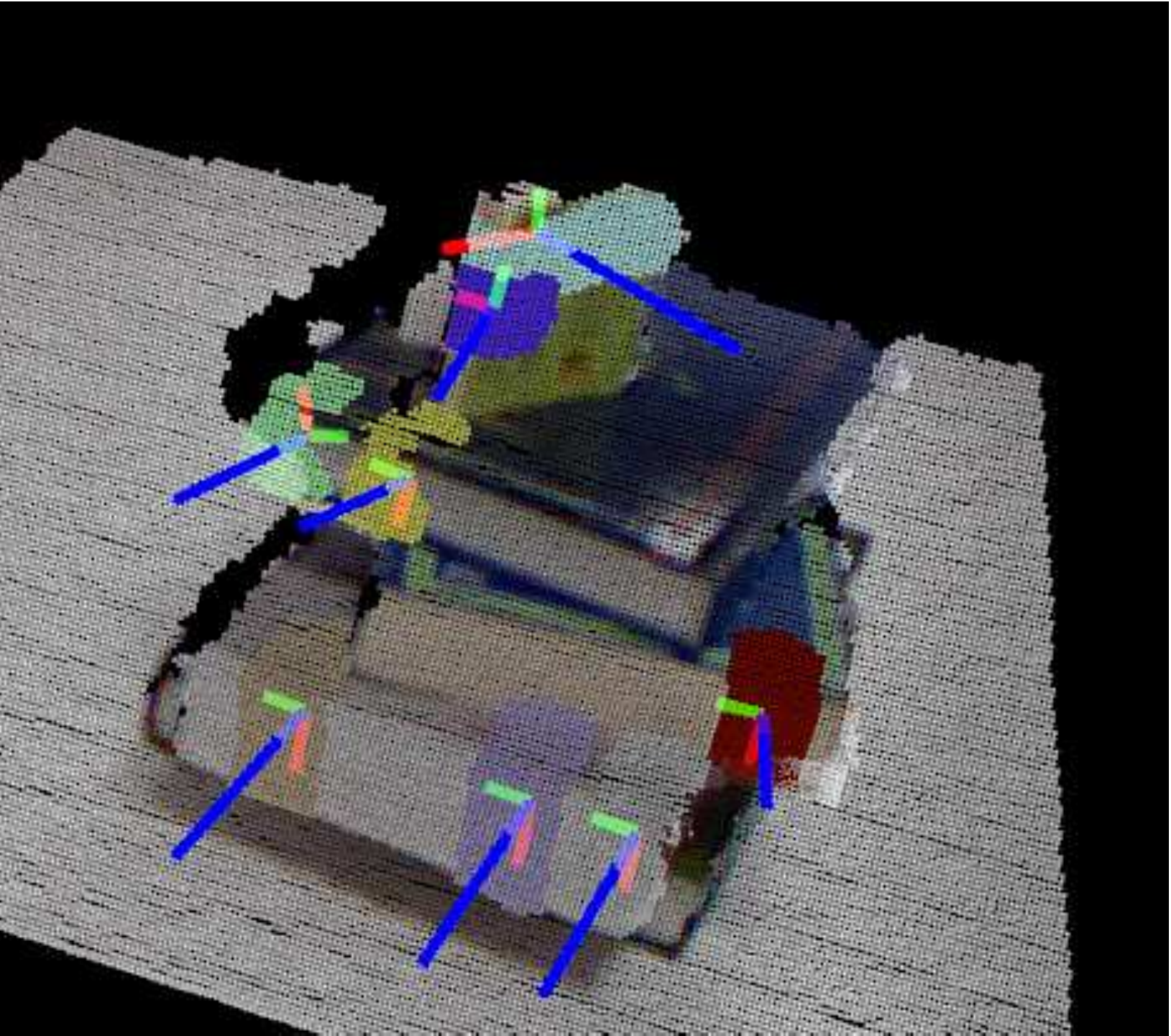} \\
\scriptsize{(e)} & \scriptsize{(f)} 
\end{tabular}
\caption{Visual comparison of the performance of the proposed
  algorithm with Platt's algorithm  \cite{Pas2013LocalizingGA} on TCS
Dataset 2. The Cyan coloured patches on left hand side figures are the
handles detected using Platt's algorithm. The patches on right side
figures along with gripper pose show affordances obtained using the
propose algorithm. }
\label{fig:cluttercomp}
\end{figure}

In this section, we demonstrate the performance of our proposed
algorithm in a cluttered environment. A new dataset is created for
this purpose. It is called `TCS Grasp Dataset 2' and it contains 40
frames each one showing multiple objects in extreme clutter situation.
The objects in the clutter have different shapes and sizes and, may
exhibit partial or full occlusion. The performance of our algorithm on
some of these frames are shown in Figure \ref{fig:clutter}.  The
performance comparison with Platt's algorithm \cite{ten2016localizing}
\cite{Pas2013LocalizingGA} is shown in Figure \ref{fig:cluttercomp}.
As one can see in Figure \ref{fig:clutter}, the proposed algorithm is
successful in finding graspable affordances for rectangular objects
with flat surfaces such as books in addition to objects with curved
surfaces. It also shows multiple handles detected for some of the
objects. All those handles which do not satisfy the geometric
constraints of the gripper are rejected and hence not shown in this
figure. The maximum hand aperture considered for finding these
affordances is 8 cm. In contrast, Platt's algorithm
\cite{ten2016localizing} \cite{Pas2013LocalizingGA} fails to detect
any handles for flat rectangular objects as shown in Figure
\ref{fig:cluttercomp}. The Table \ref{tab:cluttercomp} provides a more
quantitative comparison between these two algorithms. It shows that
the proposed algorithm is able to detect at least 86\% of unique
handles in the dataset compared to 36\% recall achieved with Platt's
algorithm. The performance of these two algorithms on various publicly
available datasets is summarized in Table \ref{tab:padcomp}.
Cornell Grasping Dataset \cite{lenz2015deepgrasp}
contains single object per frame and grasping rectangle as ground
truth. Their best result (93.7\%) reported is in terms of accuracy
whereas recall from our method is 96\% at 100\% precision. The Bird
Bird dataset \cite{singh2014bigbird} consists of segmented individual
objects and yields a maximum recall of 99\%. This high level of
performance is due to the fact that the object point cloud is
segmented and processed for noise removal. This dataset, as such, does
not include clutter and has been included in this section for the sake
of completeness. The ECCV dataset \cite{Aldoma2012}, Kinect Dataset
\cite{SegIROS11} and the Willowgarage dataset \cite{willow} have multiple
objects in one frame and may exhibit low level of clutter. All of
these dataset are created for either segmentation or pose estimation
purposes, therefore ground truth for grasping is not provided. We have
evaluated the performance (as reported in Table \ref{tab:padcomp}) using
manual annotation. The extent of clutter in these datasets is not
comparable to what one will encounter in a real world scenario. This
is one of the reasons why we had to create our own dataset. As one can
see in Figure 7 (b), the TCS grasp dataset 2 exhibits extreme clutter
scenario.  As one can observe in Table \ref{tab:cluttercomp}, the
proposed algorithm provides better grasping performance compared to
the current state-of-the-art reported in literature.

\begin{table}[!t]
\centering
\caption{Performance Comparison for TCS Dataset 2 - Multiple objects
in a Cluttered Environment}
\label{tab:cluttercomp}
\scriptsize
\begin{tabular}{|>{\centering\arraybackslash}m{0.6cm}|>{\centering\arraybackslash}m{1cm}|>{\centering\arraybackslash}m{0.9cm}|>{\centering\arraybackslash}m{0.7cm}|>{\centering\arraybackslash}m{0.9cm}|>{\centering\arraybackslash}m{0.7cm}|} \hline
. &  & \multicolumn{2}{c|}{Platt's Method \cite{Pas2013LocalizingGA} \cite{ten2016localizing}} & \multicolumn{2}{c|}{Proposed Method} \\ \hline
Frame No. & No. of graspable objects in the frame       & max no. of handles detected    & \% Recall      & max no. of handles detected     & \% Recall                     \\ \hline
\#1    & 8      & 2        & 25        & 6     & 75                                                    \\ \hline
\#3    & 8      & 3	     & 38        & 6     & 75                                                   \\ \hline
\#5    & 6      & 3        & 50        & 6     & 100                                                     \\ \hline
\#7         & 7      & 2        & 28        & 7     & 100                                                     \\ \hline
\#10         & 6      & 3        & 50        & 5     & 83                                                    \\ \hline
\#12         & 7      & 2        & 28        & 7     & 100                                                     \\ \hline
\#13         & 7      & 2        & 28        & 7     & 100                                                     \\ \hline
\#16         & 8      & 1        & 13        & 6     & 75                                                     \\ \hline
\#20         & 8      & 2        & 25        & 6     & 75                                                     \\ \hline
\#23         & 9      & 2        & 22        & 8     & 89                                                    \\ \hline
\#24         & 6      & 3        & 50        & 5     & 83                                                     \\ \hline
\#26         & 5      & 3        & 60        & 3     & 60                                                    \\ \hline
\#28         & 5      & 2        & 40        & 5     & 100                                                     \\ \hline
\#30         & 6      & 2        & 33        & 6     & 100                                                     \\ \hline
\#32         & 6      & 2        & 33        & 5     & 83                                                 \\ \hline
\#37         & 5      & 1        & 20        & 5     & 100                                                     \\ \hline
\#38         & 2      & 2        & 100       & 2     & 100                                                    \\ \hline
\#39         & 4      & 2        & 50        & 3     & 75                                                    \\ \hline
\#36         & 5      & 1        & 20        & 4     & 80                                                    \\ \hline
Total    & 118    & 40       & 33        & 102   & 86 \\ \hline
\end{tabular}
\end{table}

\begin{table}[!t]
\centering
\caption{Performance Comparison on Various Publicly Available Datasets}
\label{tab:padcomp}
\small
\begin{tabular}{|>{\centering\arraybackslash}m{0.3cm}|c|>{\centering\arraybackslash}m{1.5cm}|>{\centering\arraybackslash}m{1.5cm}|} \hline
  &    & \multicolumn{2}{c|}{\% Recall} \\ \hline
S. No. & \textbf{Dataset} & Proposed Method & Platt's Algorithm \cite{Pas2013LocalizingGA} \cite{ten2016localizing}   \\ \hline
1 & Big Bird \cite{singh2014bigbird}       & 99\%         & 85\% \cite{plattgrasppose2016}   \\ \hline
2 & Cornell Dataset \cite{lenz2015deepgrasp}   & 95.7\%         & 93.7\% \cite{lenz2015deepgrasp}     \\ \hline
3 & ECCV \cite{Aldoma2012}           & 93\%    & 53\%     \\ \hline
4 & Kinect Dataset  \cite{SegIROS11} & 91\%    &52\%      \\ \hline
5 & Willow Garage \cite{willow}   & 98\%         & 60\%     \\ \hline
6 & TCS Dataset-1 \cite{tcsgraspdata2018}   & 94\%         & 48\%   \\ \hline
7 & TCS Dataset-2 \cite{tcsgraspdata2018}   & 85\%         & 34\%   \\ \hline
\end{tabular}
\end{table}

\subsection{Computation Time}  

The computational performance of the algorithm can be assessed by
analyzing the  Table \ref{tab:comptime}.  This table shows the average
computation time per frame for two TCS datasets. As one can observe, the bulk
of the time is taken by the region growing algorithm which is the
first step of our proposed method. This time is proportional to the
size of the point cloud data. The second stage of our algorithm
detects valid handles by applying geometric constraints on the surface
segments found in the first step. This step is considerably faster
compared to the first step. Many of the segments created in the first
step are rejected in the second step to identify valid grasping
handles as can be see in the 4th and 5th columns in this table. The
computation time for each valid handle for the two datasets is 4 and 5
ms respectively. 

The total processing time for a complete frame with around 40K data
point is approximately 800 ms to 1 second. This is quite reasonable in
the sense that the robot can process around 60 frames per second which
is very good for most of the industrial applications. This time can be
further reduced by detecting a particular ROI within the image thereby
reducing the number of points to be processed in the frame. The
computation time per frame can also be reduced significantly
by downsampling the point cloud.  There is a limit to the extent of
downsampling allowed as it is directly linked to the quality and
quantity of handles detected. For high speed applications, one may use
FPGA or GPU based embedded computing platform.

\begin{table}[!t]
  \centering
  \scriptsize
  \caption{Average computation time per frame. All values are reported per frame basis and are averaged
  over all frames.}
  \label{tab:comptime}
  \begin{tabular}{|>{\centering\arraybackslash}m{0.8cm}|
    >{\centering\arraybackslash}m{0.8cm}|
    >{\centering\arraybackslash}m{1.0cm}|
    >{\centering\arraybackslash}m{0.9cm}|
    >{\centering\arraybackslash}m{0.9cm}|
    >{\centering\arraybackslash}m{0.9cm}|} \hline 
    Dataset & \# data in point cloud & Time for
    Region Growing algorithm (sec) & \# segments detected & \# valid handles detected  &
    Handle detection time (sec) \\ \hline
    TCS Dataset 1 & 37050 & 0.729 & 77 & 10  & 0.055  \\ \hline
    TCS Dataset 2 & 42461 & 0.82 &  182 & 43   & 0.171  \\ \hline
  \end{tabular}
\end{table}

\section{Conclusion}\label{sec:conc}

This paper looks into the problem of finding graspable affordances (or
suitable grasp poses) needed for picking various household objects
using two finger parallel-jaw gripper in extreme clutter environment.
These affordances are to be extracted from a single view 3D point
cloud obtained from a RGBD or a range sensor without any apriori
knowledge of object geometry. The problem is solved by first creating
surface segments using a modified version region growing algorithm
based on surface smoothness condition. This modified version of region
growing algorithm makes use of a pair of user-defined thresholds and a
concept called \emph{edge point} to discard false boundaries arising
out of sensor noise. The problem of real-time pose detection is
simplified by transforming the 6D search problem to a 1D search
problem through scalar projection and exploiting the geometry of the
two-finger gripper. Through experiments on several datasets, it is
demonstrated that the proposed algorithm outperforms the existing
state-of-the-art methods in this field. In the process, we have also
contributed a new dataset to demonstrate its working in extreme
clutter environment and is being made available online for use by
the research community.

%%%%%%%%%%%%%%%%%%%%%%%%%%%%%%%%%%%%%%%%%%%%%55
\section*{References}
\bibliography{root_sk}
\bibliographystyle{elsarticle-num}

\end{document}